\documentclass[accepted]{uai2026} % for initial submission
\usepackage[american]{babel}
\usepackage{natbib} % has a nice set of citation styles and commands
\usepackage{mathtools} % amsmath with fixes and additions
\usepackage{booktabs} % commands to create good-looking tables
\usepackage{tikz} % nice language for creating drawings and diagrams
\usepackage{graphicx} % figures
\usepackage{caption}
\usepackage{subfigure}
\usepackage{multirow}
\usepackage{bm}
\usepackage{amsfonts}
\usepackage{amssymb}

\newtheorem{theorem}{Theorem}[section]
\newtheorem{proposition}[theorem]{Proposition}
\newtheorem{lemma}[theorem]{Lemma}
\newtheorem{corollary}[theorem]{Corollary}
\newtheorem{definition}[theorem]{Definition}

\newenvironment{proof}{{\noindent\it Proof.}}{\hfill $\square$\par}

\title{Learning Expressive Random Feature Models via Parametrized Activations}

\author[1]{Zailin~Ma\thanks{Email: mazailin@stu.pku.edu.cn}}
\author[1]{Jiansheng~Yang}
\author[2]{Yaodong~Yang\thanks{Corresponding author: yaodong.yang@pku.edu.cn}}
\affil[1]{%
    School of Mathematical Sciences, Peking University
}
\affil[2]{%
    Institute for Artificial Intelligence, Peking University
}

\begin{document}
\maketitle
\makeatletter
\global\let\@thanks\@empty
\makeatother

\begin{abstract}
  The random feature (RF) method is a powerful kernel approximation technique, but it typically uses fixed activation functions, limiting its adaptability across diverse tasks. To overcome this limitation, we introduce the Random Feature Model with Learnable Activation Functions (RFLAF), a novel statistical model that parameterizes activation functions as weighted sums of basis functions within the random feature framework. Examples of basis functions include radial basis functions (RBFs), spline functions, polynomials, and so forth. For theoretical results, we consider RBFs as representative basis functions. We start with a single RBF as the activation, and then extend the results to multiple RBFs, demonstrating that RF models with a learnable activation component substantially expand the represented function space. We provide estimates on the required number of samples and random features to achieve low excess risk. In our experiments, we test RFLAF with three types of bases: radial basis functions, spline functions and polynomials. Experimental results show that RFLAFs with RBFs and splines consistently outperform other RF models, where RBFs are three times more computationally efficient than splines. We then unfreeze the first-layer parameters and retrain the models, validating the expressivity advantage of learnable activation components on regular two-layer neural networks. Our work provides a deeper understanding of learnable activation components within modern neural network architectures.
\end{abstract}

\section{INTRODUCTION}

Kernel methods are powerful tools for solving nonlinear learning problems by leveraging kernel functions to implicitly map data into high-dimensional spaces. However, they can be computationally intensive and lack scalability with large datasets. The random feature (RF) model, introduced in \citep{rahimi2008weighted}, offers a solution by approximating kernel functions with a finite number of random features, allowing the application of linear algorithms for large-scale computation \citep{li2021towards,liu2021random}.

Despite their advantages, random feature models typically use a fixed activation function, limiting their adaptability during data fitting. This rigidity prevents the model from automatically searching for activation functions for optimal performance across various tasks. Meanwhile, recent work such as Kolmogorov-Arnold Networks (KANs) \citep{liu2024kan} has demonstrated the effectiveness of learnable activation modules. Therefore, it is natural to study random feature models that incorporate learnable activation functions. 

% Current methods involve parametrizing learnable activation functions with splines \citep{liu2024kan,fakhoury2022exsplinet,bohra2020learning,aziznejad2019deep}. However, the piecewise definition of splines complicates the theoretical analysis, resulting in difficulties such as the infeasibility of deriving the analytic form of the kernel even in the case of a single spline function. Moreover, models equipped with splines may face difficulty with efficient convergence. For instance, KANs uses LBFGS with line search instead of common gradient descent method to boost convergence.

In this paper, we introduce the Random Feature models with Learnable Activation Functions (RFLAF), a novel statistical model that parametrizes learnable activation functions using weighted sums of basis functions within the random feature framework. Examples of basis functions that have universal approximation ability include radial basis functions (RBFs), B-spline functions \citep{fakhoury2022exsplinet}, polynomials \citep{goyal2019learning}, and so forth. Among them, we consider RBFs as representative basis functions for theoretical analyses. We study the analytic properties of the kernel induced by a single RBF, and then provide approximation and generalization bounds for the multiple-RBF case. For experimental validation, we test RFLAF with RBF, spline functions and polynomials, and compare them with standard RF models across various tasks. Experimental results show that RFLAFs with RBFs or B-splines consistently exhibit greater expressivity and adaptability than standard RF models. More strikingly, RBFs are three times more computationally efficient than B-splines, achieving the best overall performance. We further unfreeze the first-layer parameters of RFLAFs and retrain the models, validating the expressivity advantage of learnable activation components on regular two-layer neural networks. This paper offers a comprehensive analysis of the model, covering both the theoretical foundations and the empirical validation. Our contributions are summarized as follows.

\(\bullet\) We identify an unexplored kernel in the case of a single RBF as the activation. We provide the first result on the analytic form of this kernel, and investigate its representation and approximation properties (Section \ref{sec3}).

\(\bullet\) For the general RFLAF, we provide rigorous analyses on the approximation and generalization properties. Our theories guarantee that our model improves its representational ability at a minimal cost of less than twice the number of parameters (Section \ref{sec4}), and the number of random features only needs to scale with the square root of the sample size (Section \ref{sec5}).

\(\bullet\) We conduct extensive experiments to demonstrate the advantages of RFLAF (Section \ref{sec6}). We show that RFLAFs with RBFs or splines generally outperform other baseline RF models, with RBFs being three times more efficient than splines. Finally, we validate the effectiveness of the learnable activation function on regular two-layer networks.

The rest of the paper is organized as follows. Section \ref{sec2} outlines the basics of random feature models and formally introduces our model. Theoretical analyses for models with the single and combined RBF activations are provided in Sections \ref{sec3} and \ref{sec4} respectively. Section \ref{sec5} discusses guarantees on sample complexity, and Section \ref{sec6} presents experimental results to evaluate our models.

\subsection{Related Work}

\paragraph{Random Feature (RF) Models}
Random feature models \citep{rahimi2008weighted,rahimi2007random} are initially motivated by the fact that randomization is computationally cheaper than optimization \citep{amit1997shape,moosmann2006randomized}. More recently, by virtue of the relation between a kernel and its Fourier spectral density, random features act as a technique to scale up kernel methods \citep{lopez2014randomized,sun2018but,jacot2018neural,arora2019exact,zandieh2021scaling,du2019graph,zambon2020graph,fu2024can,shen2019random}. For instance, researchers apply the random feature technique in Transformers to approximate the softmax or Gaussian kernel inside the attention block, successfully reducing the order of the time and space complexity with respect to token length $L$ \citep{choromanski2020rethinking,peng2021random}. From a theoretical perspective, when viewed as a class of two-layer neural networks with fixed weights in the first layer, random feature models provide deep insights for partly understanding deep neural networks \citep{cao2019generalization,arora2019fine,mei2022generalization,chizat2020implicit}. Research effort has also been devoted to deriving approximation and generalization bounds with respect to the number of random features and sample size \citep{sutherland2015error,rudi2017generalization,avron2017random,bach2017equivalence,mei2022generalization}. \citet{li2021towards} contributes to a unified analysis of random Fourier features.

\paragraph{Learnable Activation Functions}
Previous work studies the learnable activation functions and attempts to incorporate them into neural network architectures. Activation functions are parametrized in a continuous way as splines \citep{liu2024kan,liu2024kan2,fakhoury2022exsplinet,bohra2020learning,aziznejad2019deep}, polynomials \citep{goyal2019learning}, sigmoid linear units \citep{ramachandran2017searching} and neural networks \citep{zhang2022neural}, or in a discrete way \citep{bingham2022discovering}. 
The related notion of RBF networks introduced in \citep{lowe1988multivariable} is fundamentally distinct from our model. In our work, RBFs are used for universal approximation, whereas the RBF network applies them for functional interpolation. Optimal activation functions studied in \citep{wang2023optimal,demir2024optimal} directly relate to our work because RFLAF has the potential to learn optimal activations directly from data.

\section{PRELIMINARIES}
\label{sec2}

\subsection{Basics on Random Feature Models}

In this section, we provide some foundations of random feature models \citep{rahimi2007random,rahimi2008uniform} related to our work. 

Given a function \(\sigma(x; w) : \mathcal{X} \times \mathcal{W} \to \mathbb{R}\). Let \(\mu\) be a probability measure on \(\mathcal{W}\). The class of infinite-width random feature model is defined as
\[
\mathcal{F} = \left\{ f : f(x) = \int_\mathcal{W}  \sigma(x; w) v(w)\mu(dw) , v \in \mathcal{H}_\mathcal{W} \right\},
\]
where \(\mathcal{H}_\mathcal{W} = \left\{ v(w) : \int_\mathcal{W} v(w)^2\mu(dw) < \infty \right\}\) is a Hilbert space with norm \(\|v\|_{\mathcal{H}_\mathcal{W}}^2 = \int_\mathcal{W} v(w)^2\mu(dw)\) and inner product \(\langle v, u \rangle_{\mathcal{H}_\mathcal{W}} = \int_\mathcal{W} v(w) u(w) \mu(dw)\). Furthermore, \(\mathcal{F}\) is endowed with a norm \(\|\cdot\|_{\mathcal{F}}\) and the inner product \(\langle \cdot, \cdot \rangle_{\mathcal{F}}\):
\(
\|f\|_{\mathcal{F}} = \inf_{f=\langle v, \sigma(\cdot) \rangle_{\mathcal{H}_\mathcal{W}}} \|v\|_{\mathcal{H}_\mathcal{W}},\) and
\(\langle f, g \rangle_{\mathcal{F}} = \frac{\|f + g\|_{\mathcal{F}}^2 - \|f - g\|_{\mathcal{F}}^2}{4}.
\)

Besides, we define the corresponding reproducing kernel \(K : \mathcal{X} \times \mathcal{X} \to \mathbb{R}\) as
\[
K(x, y) = \int_\mathcal{W} \sigma(x; w)\sigma(y; w) \mu(dw).
\]

Define the RKHS induced by this kernel as \(\mathcal{H}_K\) with corresponding norm \(\|\cdot\|_{\mathcal{H}_K}\) and the inner product \(\langle \cdot, \cdot \rangle_{\mathcal{H}_K}\). Generally \citep{bai2019beyond}, for any feature map \(\phi : \mathcal{X} \to \mathcal{H}\) (where \(\mathcal{H}\) is a Hilbert space) that induces the kernel \(K\), i.e., \(K(x, y) = \langle \phi(x), \phi(y) \rangle_{\mathcal{H}}\), we have that for any function \(f\),
\[
\|f\|_{\mathcal{H}_K} = \inf_{f=\langle \mathbf{u}, \phi(\cdot) \rangle_{\mathcal{H}}} \|\mathbf{u}\|_{\mathcal{H}},
\]
which indicates the equivalence among different feature maps that generate the same kernel. 

Finally, we have the following proposition according to \citep{minh2006mercer}.
\begin{proposition}
	\label{equalnorm}
	Given the above definition of \(\mathcal{F}\) and \(\mathcal{H}_K\), we have that \((\mathcal{F}, \|\cdot\|_{\mathcal{F}}) = (\mathcal{H}_K, \|\cdot\|_{\mathcal{H}_K})\).
\end{proposition}

\subsection{Parametrization of Activation Functions and Finite-width Approximation}
\label{sec22}
Standard random feature models consider the case where the activation function $\sigma$ is a fixed univariate function such as ReLU, and $\sigma(x;w)=\sigma(w^\top x)$. In this work, we broaden the target function class where $\sigma$ can be any function in $C_c(\mathbb{R})$, namely the continuous functions with compact support.

Let $x\in \mathbb{R}^d$, and $w\sim\mathcal{N}(0,I_d)$. For technical convenience, we assume $\sigma:\mathbb{R}\rightarrow\mathbb{R}$ and $v:\mathbb{R}^d\rightarrow\mathbb{R}$ to be Lipschitz continous. Suppose that $\sigma$ is supported on a bounded closed interval $\mathcal{K}\subseteq \mathbb{R}$. We define the target function class as
\begin{equation}
	\label{tgfc}
    \begin{aligned}
	\mathcal{F}_{\mathcal{K}}:=\{& f : f(x) = \mathbb{E}_{w\sim \mathcal{N}(0,I_d)}\left[\sigma(w^\top x)v(w)\right],\\
  & \sigma\in C_c(\mathcal{K}), \|\sigma\|_{\mathrm{Lip}}\leq L, \|v\|_{\mathrm{Lip}}\leq L_v \},
    \end{aligned}
\end{equation}
where $\|\cdot\|_{\mathrm{Lip}}$ denotes the Lipschitz constant of a function.

Suppose the target function \(f=\mathbb{E}\left[\sigma(w^\top x)v(w)\right]\in \mathcal{F}_{\mathcal{K}}\). The motivations of our model are twofold. In the first step, we consider using an array of basis functions \(\left\{B_i(x)\right\}_{i\in[N]}\) to approximate any potentially targeted activation functions, i.e.,
\(
    \tilde{\sigma}(x):=\sum_{i=1}^N a_i B_i(x),
\)
where $a_i$ are learnable parameters.
The corresponding function is then 
\(
    \tilde{f}(x) := \mathbb{E}_{w\sim \mathcal{N}(0,I_d)}\left[\tilde{\sigma}(w^\top x)v(w)  \right]\approx f(x).
\)
In the second step, we approximate \(\tilde{f}(x)\) with the finite-width random feature model \(\sum_{m=1}^{M}\tilde{\sigma}(w_m^\top x)v(w_m)/M\approx\mathbb{E}_{w\sim \mathcal{N}(0,I_d)}[\tilde{\sigma}(w^\top x)v(w)]\), where \(\{w_m\}_{m=1}^{M}\overset{i.i.d}{\sim} \mathcal{N}(0,I_d)\) are sampled independently and identically. Consequently, we formulate the Random Feature model with Learnable Activation Functions (RFLAF) as
\begin{equation}
	\hat{f}(x;\boldsymbol{a},\boldsymbol{v}):=\frac{1}{M}\sum_{m=1}^M\sum_{i=1}^N a_i B_i(w_m^\top x)v_m,
\end{equation}
where $\boldsymbol{a}=(a_1,...,a_N)\in \mathbb{R}^{N}, \boldsymbol{v}=(v_1,...,v_M)\in\mathbb{R}^M$ are learnable parameters. 

For the theoretical analysis hereafter, we consider the radial basis functions as representatives, namely, we choose the array of basis functions to be \[\left\{B_i(x)=\exp\left(-\frac{(x-c_i)^2}{2h_i^2}\right)\right\}_{i\in[N]}\] with centers $c_i$ and widths $h_i$ set in prior. The basis $\{B_i(x)\}$ in RFLAF can be replaced by any function class that has universal approximation properties \citep{nestoridis2007universal} (e.g., B-splines in \citep{fakhoury2022exsplinet}, polynomials in \citep{goyal2019learning}).

\section{RANDOM FEATURE MODELS WITH A SINGLE RADIAL BASIS FUNCTION}
\label{sec3}
We first study the random feature model with a single radial basis function, which is a special case of RFLAF when $N=1$. Theorem \ref{pro31} provides the explicit expression of the corresponding kernel, which extends the result of the centered Gaussian activations in \citep{han2022fast} to general non-centered Gaussian activations. Our result is compatible with the former results but provides a more case-specific analysis.

The target function of interest admits representation 
\begin{equation}
	\label{expform}
	\varphi(x) = \mathbb{E}_{w\sim \mathcal{N}(0,I_d)}\left[B(w^\top x)v(w)\right],
\end{equation}
where the activation function $B(x) = \exp{\left(-{{(x-c)^2} / (2h^2)}\right)}$ is a radial basis function with center $c$ and width $h$.
The corresponding reproducing kernel is
\begin{equation}
	K(x,x^\prime):=\mathbb{E}_{w\sim\mathcal{N}(0,I_d)}\left[B(w^\top x)B(w^\top x^\prime)\right].
\end{equation}

The first result presents the explicit expression of the kernel. 
\begin{theorem}
	\label{pro31}
	For any $x,x^\prime\in\mathbb{R}^d$, we have that
	\begin{equation}
		\label{kernel0}
		\begin{aligned}
			&K(x,x^\prime)={h^2\over \sqrt{(h^2+\|x\|^2)(h^2+\|x^\prime\|^2)-\langle x, x^\prime\rangle^2}}\cdot\\
            &\exp \Bigg(-\frac{c^2}{2}\cdot \frac{(h^2+\|x\|^2)+(h^2+\|x^\prime\|^2)-2\langle x, x^\prime\rangle}{(h^2+\|x\|^2)(h^2+\|x^\prime\|^2)-\langle x, x^\prime\rangle^2}  \Bigg).
		\end{aligned}
	\end{equation}
\end{theorem}

Consider normalized inputs $\|x\|_2=\|x^\prime\|_2=1$, then $r=\langle x, x^\prime\rangle \in[-1,1]$. The kernel degenerates to a rotation-invariant kernel \citep{liu2021random}. We slightly abuse the notation and define the univariate function $K(r)$ to be the rotation-invariant form of the kernel (\ref{kernel0}).
\begin{equation}
	\label{kernel}
	K(r):={h^2\over \sqrt{(1+h^2)^2-r^2}}\exp \Bigg(-\frac{c^2}{1+h^2+r}  \Bigg).
\end{equation}

We present the explicit expression of kernel (\ref{kernel}).
\begin{theorem}
	\label{pro32}
	The rotation-invariant kernel $K(r)$ has Taylor expansion as
	\begin{equation}
		\label{taylor}
		K(r)=e^{-p}\frac{h^2}{1+h^2}\sum_{n=0}^{\infty}\frac{R_n(p)}{n!(1+h^2)^n}r^n,
	\end{equation}
	where $p=\frac{c^2}{1+h^2}\in [0,+\infty)$, and the polynomials are
	\(
		R_n(x)=
		\begin{cases}
			P_k^2(x), & n=2k, \\
			xQ^2_k(x), &  n=2k+1,
		\end{cases}
	\) 
	\[
		P_k(x)=\sum_{i=0}^{k} (-1)^{k-i}\frac{(2k-1)!!}{(2i-1)!!}\cdot\binom{k}{i} x^{i},\]
  \[      Q_k(x)=\sum_{i=0}^{k} (-1)^{k-i}\frac{(2k+1)!!}{(2i+1)!!}\cdot\binom{k}{i} x^{i}.
	\]
	Therefore, the feature mapping with respect to the kernel 
	(\ref{kernel}) is 
	\[
		\phi(x)=\left(\frac{he^{-{p\over 2}}R_n^{1\over 2}(p)}{\sqrt{n!(1+h^2)^{n+1}}}x^{\otimes n}\right)_{n=0}^{\infty}.
	\]
\end{theorem}
% The proof is contained in Appendix \ref{proofpro32}.
Define the represented function class as
\[
	\mathcal{H}_{c,h}=\{\varphi : \varphi(x)=\sum_{n=0}^{\infty}\langle F_n, x^{\otimes n}\rangle, D_{c,h}(\varphi)<\infty\},
\]
where $F_n\in \mathbb{R}^{d^n}$ and 
\[
	D_{c,h}(\varphi):=\frac{e^p}{h^2}\sum_{n=0}^{\infty}\frac{n!(1+h^2)^{n+1}}{R_n(c^2/(1+h^2))}\|F_n\|_{\mathrm{Fr}}^2.
\]
Then we have the following representation theorem.
\begin{corollary}
	\label{pro33}
	For any $f\in\mathcal{H}_{c,h}$, there exists $v:\mathbb{R}^d\rightarrow \mathbb{R}$ such that $f(x) = \mathbb{E}_{w}\left[B(w^\top x)v(w)\right]$ and $\mathbb{E}_{w}\left[v(w)^2\right]\leq {D_{c,h}(f)}$, where $w\sim \mathcal{N}(0,I_d)$ and $B(x) = \exp{\left(-{(x-c)^2\over 2h^2}\right)}$.
\end{corollary}
Approximating $\varphi$ with finite-width model
\(\hat{\varphi}(x)=\frac{1}{M}\sum_{m=1}^{M}B(w_m^\top x)v_m\) where $\{v_m\}_{m\in[M]}$ are learnable parameters, the approximation error can be estimated below.
\begin{theorem}
	\label{pro34}
    Let $v(w)$ be $L_v$-Lipschitz and $R=\sqrt{2L_v^2d+2|v(\boldsymbol{0})|^2}$. Suppose that $\{w_m\}_{m=1}^{M}\overset{i.i.d}{\sim} \mathcal{N}(0,I_d)$, then with probability of at least $1-\delta$, there exists $\{v_m\}_{m=1}^{M}$ such that
    \[\mathbb{E}_{x}\left|\hat{\varphi}(x) - \varphi(x)\right|\leq 18R\sqrt{\log{(4/\delta)}/M},\]
	and
	\(
		\sum_{m=1}^{M}v_m^2\leq 49MR^2{\log(2/\delta)},
	\)
	where we assume $\delta<1/2$. Note that the inequalities hold for whatever distribution $x$ are sampled from.
\end{theorem}
% The proof is contained in Appendix \ref{proofpro34}. 
Proofs of all the above statements are provided in Appendix \ref{proofsec3}. The proof of Theorem \ref{pro34} is not trivial, because the concentration property of $\left|\hat{\varphi}(x) - \varphi(x)\right|$ may not be uniform over $x$. We use some techniques to circumvent this problem. 

Implied by Theorem \ref{pro32} and Corollary \ref{pro33}, the represented function $f$ corresponds to a fixed feature mapping with fast decaying coefficients $F_n$, indicating a narrow function class similar to other standard RF models. Hence, using a single RBF as the activation function does not necessarily lead to a leap in the expressivity of the RF model. The key step is to combine the RBFs with learnable weights. The mechanism of learnable activation functions results in evidently enhanced expressivity of RF models, as we will demonstrate in the next section.

\section{RANDOM FEATURE MODELS WITH LEARNABLE ACTIVATION FUNCTIONS}
\label{sec4}

This section provides the result on the approximation error between the RFLAF 
\[
\hat{f}(x)=\frac{1}{M}\sum_{m=1}^M\sum_{i=1}^N a_i B_i(w_m^\top x)v_m
\]
of multiple RBFs and the target function $f^*\in\mathcal{F}_{\mathcal{K}}$ defined in Section \ref{sec22}. In the following, we denote $B_i(x)=\exp\left(-{(x-c_i)^2}/({2h_i^2})\right)$. We recall the Gaussian universal approximation theorem in \citep{bacharoglou2010approximation,nestoridis2007universal}.

\paragraph{Gaussian Universal Approximation Theorem (Gaussian UAT)}
    Suppose the target function $\sigma(x)$ is a continuous function with compact support $\mathcal{K}$. For any $\epsilon>0$, there exists $N>0$ and $\{h_i,c_i,a_i\}_{i=1}^N$ such that
    \(
		\|\sigma(x)-\sum_{i=1}^{N}a_i B_i(x)\|_{\infty} < \epsilon.
	\)

Inspired by the theorem, to bridge the gap between $\hat{f}$ and $f^*$, we consider an intermediate function \[\tilde{f}(x) := \mathbb{E}_{w}\left[\sum_{i=1}^N a_i B_i(w^\top x)v(w)\right],\] where $\{a_i\}_{i\in[N]}$ are learnable and $\{c_i,h_i\}_{i\in[N]}$ are set in prior. We expect that $\sum_{i=1}^{N}a_iB_i(x)$ in $\tilde{f}$ can approximate the ground truth $\sigma(x)$ to an arbitrarily low error. To describe $c_i$ and $h_i$ precisely, we partition the support set $\mathcal{K}$ of $\sigma$.

Let the grid number be $N$. We define the grid points as $y_0=\min_{x\in\mathcal{K}}x$, $y_N=\max_{x\in\mathcal{K}}x$ and $y_i=y_0+\frac{i}{N}(y_N-y_0)$ for $1\leq i\leq N-1$. The diameter of the support is $|\mathcal{K}|:=y_N-y_0$. The grid size then is $|\mathcal{K}|/N$. Because $\sigma$ is continuous over the compact set $\mathcal{K}$, it is also bounded. Hence, $\|\sigma\|_{\infty}<\infty$. Our first result measures the approximation error between $f^*$ and $\tilde{f}$ with respect to the choice of $h_i$ and grid size. 
\begin{proposition}
	\label{pro43}
	Suppose $f^*\in \mathcal{F}_{\mathcal{K}}$ with activation function $\sigma$. For any $\epsilon>0$, by setting
    \[
    h\leq \frac{\epsilon}{4\sqrt{2}LR\sqrt{\log \frac{16\|\sigma\|_{\infty}R}{\epsilon}}},\]
    \[\frac{|\mathcal{K}|}{N}\leq\frac{\epsilon h\sqrt{{\pi e}}}{16\sqrt{2}\|\sigma\|_{\infty}R\log\left({\frac{8\|\sigma\|_{\infty}|\mathcal{K}|R}{\sqrt{2\pi}\epsilon h^2}}\right)}\land \frac{\epsilon}{4LR},
    \]
    and $h_i=h$, $c_i\in[y_{i-1},y_i]$, there exists $\{a_i\}_{i=1}^N$ such that
	\[
		\label{inftyapprox}
		\|\tilde{f}(x)-f^*(x)\|_{\infty}\leq \epsilon,\quad\quad\sum_{i=1}^{N}|a_i|^2\leq  \frac{\|\sigma\|_{\infty}^2|\mathcal{K}|^2}{2\pi h^2N}.
	\]
\end{proposition}
We use the notation $a \land b := \min\{a,b\}$ for brevity. The proof of Proposition \ref{pro43} is contained in Appendix \ref{proofpro42}. Now we are ready to measure the approximation error between $f^*$ and $\hat{f}$ in the sense of $L_1$ norm.

\begin{theorem}
	\label{pro44}
	Suppose $f^*\in \mathcal{F}_{\mathcal{K}}$ with activation function $\sigma$. For all $\epsilon>0$, under the parameter settings of Proposition \ref{pro43}, let $\{w_m\}_{m=1}^{M}\overset{i.i.d}{\sim} \mathcal{N}(0,1)$, then with probability of at least $1-\delta$, there exists $\{a_i\}_{i=1}^N$ and $\{v_m\}_{m=1}^{M}$ such that
    \[
        \mathbb{E}_{x}|\hat{f}(x)-f^*(x)|\leq \frac{18(\|\sigma\|_{\infty}R+\epsilon)\sqrt{\log(4/\delta)}}{\sqrt{M}}+\epsilon,
    \]
	and
	\[
		\sum_{i=1}^{N}a_i^2\leq  \frac{\|\sigma\|_{\infty}^2|\mathcal{K}|^2}{2\pi h^2N},\quad\quad \sum_{m=1}^{M}v_m^2\leq 49MR^2{\log\left({2\over\delta}\right)},
	\]
	where we assume $\delta<1/2$.
\end{theorem}
The proof of Theorem \ref{pro44} is contained in Appendix \ref{proofpro44}. Theorem \ref{pro44} indicates that to obtain $O(\epsilon)$ approximation error, the model requires $M=\Theta(1/\epsilon^2)$. Moreover, Proposition \ref{pro43} indicates that $1/h=\widetilde{\Theta}(1/\epsilon)$\footnote{$\widetilde{\Theta}(\cdot)$ stands for $\Theta(\cdot)$ but hides the logarithmic terms} and $N=\widetilde{\Theta}\left({1}/{\epsilon h}\right)$. Hence, $N=\widetilde{\Theta}\left({1}/{\epsilon^2}\right)=\widetilde{\Theta}(M)$. The number of grid points $N$ should scale with approximately the same order of $M$.

In practice, however, we find that a very humble number of basis functions $N$ are sufficient to improve the expressivity of the model to a large extent. For instance, RFLAF with $N=16$ already outperforms standard RF models across various tasks (see section \ref{sec6}). To summarize, RFLAF gains enhanced expressivity with very minor increase in parameter number.

\section{GENERALIZATION BOUNDS AND SAMPLE COMPLEXITY OF LEARNING}
\label{sec5}

To complete the theoretical analysis of the model, we provide the worst-case analysis regarding the generalization bounds of learning in this section.

Suppose the data distribution is $\mathtt{P}$ and the samples are $S=\{(x_i,y_i)\}_{i=1}^{n}\overset{i.i.d.}{\sim}\mathtt{P}$. Suppose the loss function $\ell(\hat{y},y)$ is $\rho$-Lipschitz in $\hat{y}$ and $|\ell(0,y)|\leq \rho$ for any $y$ (a common setting as in \citep{li2021towards}). The population risk and the empirical risk are defined respectively as 
\[
	L_{D}(f):=\mathbb{E}_{x,y\sim \mathtt{P}}[\ell(f(x),y)],~
    L_{S}(f):=\frac{1}{n}\sum_{i=1}^{n}\ell(f(x_i),y_i).
\]
The minimizer of the population risk is
\begin{equation}
	\label{eq51}
	f^*:=\mathop{\mathrm{argmin}}_{f\in \mathcal{F}_{\mathcal{K}}}L_{D}(f).
\end{equation}
Under the setting of Theorem \ref{pro44}, suppose $\{w_m\}_{m=1}^{M}\overset{i.i.d}{\sim} \mathcal{N}(0,I_d)$ are sampled, and $h$, $N$ and $\{c_i\}_{i\in[N]}$ are fixed. We aim at learning the parameters $V=(\boldsymbol{a},\boldsymbol{v})=(a_1,...,a_N,v_1,...,v_M)$ in ${f}_V(x):=\frac{1}{M}\sum_{m=1}^M\sum_{i=1}^N a_i B_i(w_m^\top x)v_m$. Guided by Theorem \ref{pro44}, the constrained set is set to be
\begin{equation}
	\label{constrainedset}
    \begin{aligned}
	\mathcal{V}:=\Bigg\{&V=(\boldsymbol{a},\boldsymbol{v}) \in \mathbb{R}^{N}\times \mathbb{R}^{M}:\|\boldsymbol{a}\|_2\leq\frac{\|\sigma\|_{\infty}|\mathcal{K}|}{h\sqrt{2\pi N}},\\
  &\quad\quad\quad\|\boldsymbol{v}\|_2\leq 7R\sqrt{M{\log\left({2\over\delta}\right)}}\Bigg\}.
    \end{aligned}
\end{equation}
% Let $K_1:=\sqrt{\frac{C_2\epsilon}{R}}$, $K_2:=\sqrt{\frac{2MR^2}{\delta}}$.
Denote $f_{\mathcal{V}}=\{f_V\}_{ V\in \mathcal{V}}$. The minimizer of the empirial risk is 
\begin{equation}
	\label{eq52}
	f_S:=\mathop{\mathrm{argmin}}_{f_V \in f_\mathcal{V}}L_{S}(f_V).
\end{equation}

\begin{table*}[tbp]
  \renewcommand{\arraystretch}{1.3}
  \caption{\centering{Test Losses of Random Feature Models. $N = 16$ for RFLAFs.\\Results are reported as mean $\pm$ std. Best in bold. Second best in italics.}}
	\label{losscomp}
  \begin{center}
	\begin{small}
	\begin{sc}
	\begin{tabular}{l|ccc|cccc}
	\noalign{\vskip 0.5pt}      
  \noalign{\hrule height 0.75pt}
  \noalign{\vskip 0.5pt}
  \multirow{2}{*}{Data set}& \multicolumn{3}{c|}{rflaf} & \multicolumn{4}{c}{rfmlp} \\ \cline{2-8}
%  & rbf & bs & pl & relu & cos & tanh & sigmoid \\
    & rbf & bs & pl & relu & cos & tanh & sigmoid \\
  \hline
  MNIST       	& $\it{0.126_{\pm 0.011}}$ & $0.165_{\pm 0.021}$ & $\bf{0.124_{\pm 0.003}}$ & $0.159_{\pm 0.002}$ & $1.390_{\pm 0.055}$ & $0.277_{\pm 0.003}$ & $0.498_{\pm 0.017}$ \\
  CIFAR-10    	& $\bf{1.450_{\pm 0.011}}$ & $1.609_{\pm 0.025}$ & $1.482_{\pm 0.016}$ & $\it{1.466_{\pm 0.004}}$ & $2.641_{\pm 0.125}$ & $1.769_{\pm 0.003}$ & $1.930_{\pm 0.028}$ \\
  adult       	& $\bf{0.309_{\pm 0.002}}$ & $0.324_{\pm 0.005}$ & - & $\it{0.311_{\pm 0.002}}$ & $0.363_{\pm 0.020}$ & $0.324_{\pm 0.003}$ & $0.327_{\pm 0.004}$ \\
  \hline
  protein     	& $\it{0.204_{\pm 0.003}}$ & $\bf{0.194_{\pm 0.002}}$ & - & $0.241_{\pm 0.001}$ & $0.371_{\pm 0.022}$ & $0.650_{\pm 0.009}$ & $0.280_{\pm 0.025}$ \\
  ct          	& $\bf{0.212_{\pm 0.016}}$ & $\it{0.302_{\pm 0.038}}$ & - & $0.356_{\pm 0.096}$ & $0.589_{\pm 0.060}$ & $1.241_{\pm 0.133}$ & $0.692_{\pm 0.020}$ \\
  workloads   	& $\bf{0.465_{\pm 0.035}}$ & $\it{0.546_{\pm 0.014}}$ & - & $2.771_{\pm 0.039}$ & $2.634_{\pm 0.015}$ & $24.997_{\pm 0.265}$ & $1.707_{\pm 0.015}$ \\
  millionsongs	& $\bf{0.102_{\pm 0.001}}$ & $0.120_{\pm 0.002}$ & - & $0.951_{\pm 0.009}$ & $0.280_{\pm 0.094}$ & $8.434_{\pm 0.091}$ & $\it{0.118_{\pm 0.007}}$ \\
  \noalign{\vskip 0.5pt}      
  \noalign{\hrule height 0.75pt}
  \noalign{\vskip 0.5pt}
  \end{tabular}
  \end{sc}
  \end{small}
  \end{center}
\end{table*}

\begin{theorem}
	\label{pro51}
	Under the conditions and parameter settings of $h,N,\{c_i\}_{i=1}^{N}$ in Theorem \ref{pro44}, let $f^*$ and $f_S$ be the minimizers of the population risk and the empirical risk in Eq. (\ref{eq51}) and (\ref{eq52}) respectively. For all $\epsilon>0$, with probability of at least $1-\delta$ over $\{w_m\}_{m=1}^{M}\overset{i.i.d}{\sim} \mathcal{N}(0,I_d)$ and $\{(x_i,y_i)\}_{i=1}^{n}\overset{i.i.d.}{\sim}\mathtt{P}$, the excess risk is bounded by
	\[
		L_{D}(f_S)-L_D(f^*)\leq \frac{\rho C{\log(16/\delta)}}{h\sqrt{n}}+\frac{\rho C\sqrt{\log(8/\delta)}}{\sqrt{M}}+\rho\epsilon,
	\]
	where $C=\max\{14\left(1+7\|\sigma\|_{\infty}|\mathcal{K}|R\right), 18(\|\sigma\|_{\infty}R+\epsilon)\}$, and we assume $\delta\leq 1/2$, $h\leq 1$.
\end{theorem}
The proof is contained in Appendix \ref{proofpro51}, which is mainly reduced to estimating the Rademacher complexity of the function class induced by the constrained set (\ref{constrainedset}).

Theorem \ref{pro51} implies that to achieve $O(\epsilon)$ excess risk, it suffices to have the sample size $n$, the random feature number $M$ and the grid number $N$ to scale as
% \begin{equation}
   \[n=\widetilde{\Theta}\left({1}/{\epsilon^2h^2}\right), \quad M={\Theta}\left({1}/{\epsilon^2}\right),\quad N=\widetilde{\Theta}\left({1}/{\epsilon h}\right).\]
% \end{equation}
Indicated by Proposition \ref{pro43}, we set $h$ such that $1/h=\widetilde{\Theta}(1/\epsilon)$. Hence, even in the worst case, only $M= \widetilde{\Theta}(\sqrt{n})$ number of random features are required, matching the sharpest results on the number of features presented in \citep{li2021towards, rudi2017generalization} for standard random feature models.
% We will substantiate these findings through experimental results in the subsequent section.

\section{NUMERICAL EXPERIMENTS}
\label{sec6}

We test the models on seven real-world datasets, including three classification tasks (MNIST, CIFAR-10 and \texttt{adult}) and four large-scale UCI regression datasets (\texttt{protein}, \texttt{ct}, \texttt{workloads} and \texttt{millionsongs}), evaluated with squared error loss and cross-entropy loss respectively. For all random feature models, experiments are repeated for 10 different seeds and quantities including lossses and time are averaged to provide statistical confidence.
% We run the experiments on GeForce RTX 3090 GPU.

For RFLAF, we set the grid range $\mathcal{K}=[-2,2]$, consider a list of grid number $N=8, 16, 32, 64, 128$ and set $h=4/N$. Proposition \ref{pro43} indicates that \(c_i\) can be arbitrarily chosen within each interval \([y_{i-1},y_i]\), so we choose \(c_i\) to be on the grid points. The model width is $M=1000$ for MNIST and $M=3000$ for the other datasets. 

\subsection{Optimization Setup}
\label{E2}
% \paragraph{Optimization details.} 
We formulate the learning problem (\ref{eq52}) in the case of MSE loss for regression tasks as an unconstrained optimization problem:
\begin{equation}
   \min_{\boldsymbol{a},\boldsymbol{v}} \frac{1}{n}\sum_{i=1}^n (\boldsymbol{a}^\top \mathbf{B}(x_i)\boldsymbol{v}-y_i)^2 + \lambda_1 (\|\boldsymbol{a}\|_2^2-\|\boldsymbol{v}\|_2^2)^2+\lambda_2 \|\boldsymbol{a}\|_1,
\end{equation}
where $(\mathbf{B}(x_i))_{k,m}={B}_k(w_m^\top x_i)$. 

The problem without the \(L_1\) term is recognized as the matrix sensing problem, a canonical optimization problem in low-rank matrix
factorization \citep{chi2019nonconvex,li2018algorithmic,tu2016low}. The first regularizer $\mathcal{R}_1:=(\|\boldsymbol{a}\|_2^2-\|\boldsymbol{v}\|_2^2)^2$ is necessary to guarantee convergence by balancing the length of the two vectors, otherwise a longer \(\boldsymbol{a}\) leads to a shorter \(\boldsymbol{v}\) and the number of solutions can be infinite. The second regularizer is the common $L_1$ regularizer and is not strictly necessary. It aims to obtain sparse components for $\boldsymbol{a}$. For classification tasks, we substitute the MSE loss function with cross-entropy loss.

\subsection{Baseline Comparisons}

We consider two types of RF models. The first type is RFLAFs with RBFs, B-splines of degree two (\texttt{BS}) and Taylor polynomials (\texttt{PL}). In both cases, $N$ represents the number of the basis functions. The second type is the plain random feature models (RFMLP) with fixed activation functions (\texttt{RELU},\texttt{COS},\texttt{TANH},\texttt{SIGMOID}). All models are compared within \emph{the same width} $M$.

\begin{table}[htbp]
  \renewcommand{\arraystretch}{1.3}
  \caption{\centering{Time Comparison between RBF and BS.\\ $N = 16$. The values of train time and test time of RFMLP with ReLU are set to be 1 respectively.}}
	\label{timecomp}
	\begin{center}
	\begin{small}
	\begin{sc}
	\begin{tabular}{l|cc|cc}
	\noalign{\vskip 0.5pt}      
  \noalign{\hrule height 0.75pt}
  \noalign{\vskip 0.5pt}
  \multirow{2}{*}{Data set}& \multicolumn{2}{c|}{train time} & \multicolumn{2}{c}{test time}\\ \cline{2-5}
	& rbf & bs & rbf & bs \\
	\hline
	MNIST       	&1.1	&1.5	&1.1	&1.4	\\
  CIFAR-10    	&1.1	&1.4	&1.0	&1.4	\\
  adult       	&1.7	&4.2	&2.1	&6.2	\\
  protein     	&1.8	&4.7	&2.0	&6.3	\\
  ct          	&1.6	&3.9	&1.6	&4.4	\\
  workloads   	&1.3	&3.4	&1.4	&4.4	\\
  millionsongs	&2.0	&5.1	&2.4	&7.3	\\
	\noalign{\vskip 0.5pt}      
  \noalign{\hrule height 0.75pt}
  \noalign{\vskip 0.5pt}
  \end{tabular}
  \end{sc}
  \end{small}
  \end{center}
\end{table}

% \begin{table*}[tbp]
%     \caption{Time Consumption. $N = 16$ for RFLAFs.}
% 	\label{timecomp}
% 	\begin{center}
% 	\begin{small}
% 	\begin{sc}
% 	\begin{tabular}{l|cccc|cccc}
% 	\toprule
%     \multirow{2}{*}{Data set}& \multicolumn{4}{c|}{train time} & \multicolumn{4}{c}{test time}\\
%     % \cmidrule(r){2-9}
% 	 & relu & rbf & bs & pl & relu & rbf & bs & pl \\
%     % Data set & rbf & bs & pl & relu & cos & tanh & sigmoid \\
% 	\midrule
% 	MNIST       	&1	&1.13	&1.54	&1.10	&1	&1.09	&1.53	&1.07	\\
%     CIFAR-10    	&1	&1.11	&1.38	&1.07	&1	&1.10	&1.38	&1.06	\\
%     adult       	&1	&1.76	&4.09	&1.48	&1	&2.01	&5.80	&1.49	\\
%     protein     	&1	&1.96	&5.26	&1.44	&1	&2.35	&7.91	&1.62	\\
%     ct          	&1	&1.79	&4.25	&1.50	&1	&1.74	&4.68	&1.39	\\
%     workloads   	&1	&1.11	&3.02	&0.89	&1	&1.21	&3.86	&0.82	\\
%     millionsongs	&1	&1.02	&2.69	&0.80	&1	&1.09	&3.30	&0.73	\\
% 	\bottomrule
%     \end{tabular}
%     \end{sc}
%     \end{small}
%     \end{center}
% \end{table*}

Table \ref{losscomp} showcases the test losses among all RF models. For brevity, we will refer to various RF models as their activation functions (e.g., RFLAF with RBFs are referred to as \texttt{RBF}). Several observations can be made:

(1) RFLAFs consistently outperform RFMLPs in all tasks. The results of accuracies for classification tasks in Appendix \ref{E3} are similar. Because RFMLP represents fixed feature mappings, it inevitably performs well in tasks that match its feature mapping but poorly in other tasks. In contrast, the learnable activation module in RFLAF allows the model to adaptively fits the data.

(2) Specifically, \texttt{RBF} and \texttt{BS} achieves the best performances among all RF models. Furthermore, \texttt{RBF} shows $2\sim 4$ times faster computational efficiency than \texttt{RBF} (Table \ref{timecomp}). As a result, RFLAF with \texttt{RBF} presents the best performance comprehensively. For \texttt{PL}, the models suffer from training instability due to the exploding magnitude of $x^n$ when $n$ is large, so they fail to converge in the last five tasks. To further substantiate the results, we also provide comparative results on low-degree polynomials that converge successfully, and results on RFLAF of $N=8,32,64,128$. In all cases, RFLAF presents consistent superiority, especially with \texttt{RBF} (see Appendix \ref{E3}).

(3) Finally, we highlight that the performance improvement of \texttt{RBF} is significant, but the cost of extra time consumption and extra parameter number compared to those of RFMLPs are actually minimal. For instance, in \texttt{workloads}, \texttt{RBF} improves the test losses compared to the best of RFMLP (\texttt{SIGMOID}) by around $70\% $, with training and testing time increased by no more than $30\%$. RFLAF with RBFs enhances standard RF models with minimal cost, showcasing the potential of RBF-based learnable activation module in enhancing modern neural network structures.

\subsection{Model Performance with respect to Grid Number}

In this part, we study how the performance of RFLAF (\texttt{RBF}) evolves as the grid number $N$ increases.
% Theorem \ref{pro44} indicates that $N$ should scale linearly with respect to $M$. However, in real scenarios, a small number like $N=8,16,32$ is sufficient for the model to fit the real data well.

\begin{figure}[htbp]
  \centering
  \includegraphics[width=\linewidth]{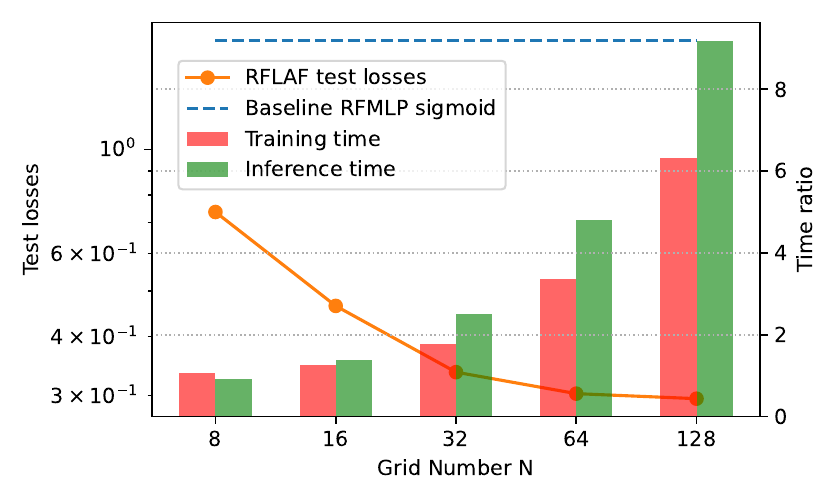}
  \caption{Test Losses and Time Consumption with respect to $N$ on Dataset \texttt{workloads}.}
  \label{Nfigure}
\end{figure}

% \begin{minipage}[htbp]{0.48\textwidth}
%     \centering
%     \includegraphics[width=\linewidth]{Nfigure.pdf}
%     \captionof{figure}{Test losses and time consumption with respect to $N$ on dataset \texttt{workloads}.}
%     \label{Nfigure}
% \end{minipage}
% \hfill 
% \begin{minipage}[htbp]{0.48\textwidth}
%     \centering
%     \captionof{table}{Epoch number of convergence.} 
%     \begin{small}
%     \begin{sc}
%     \begin{tabular}{l|ccccc}
%     \toprule
%     % Data set & 8 & 16 & 32 & 64 & 128 & relu & total \\
%     \multirow{2}{*}{Data set} & \multicolumn{5}{c}{rflaf (rbf) with $N=$} \\ %\cline{2-6}
%         & 8 & 16 & 32 & 64 & 128 \\
%     \midrule
%     MNIST	& 10	& {9}	&         6	& 4	& 3	\\
%     CIFAR-10	& {7}	& 4	&         2	& 2	& 2	\\
%     adult	& 13	& 8	&         {3}	& 1	& 1	\\
%     \bottomrule
%     \end{tabular}
%     \end{sc}
%     \end{small}
%     \label{cvcomp} 
% \end{minipage}

Figure \ref{Nfigure} shows that as $N$ grows, the test error declines until limited by the representative function class. This fact substantiates Proposition \ref{pro43} that finer grid size results in a lower approximation error. Figure \ref{Nfigure} also shows that the running time grows linearly with $N$ as the time complexity of the model is $O(MN)$. However, for $N=8,16,32$, the model performance significantly improves as $N$ increases, whilst the time consumption is acceptable (almost the same as RFMLP when $N=8,16$, around 2 times when $N=32$). Results on other datasets are similar (see Appendix \ref{E3}).

\begin{table}[htbp]
  \renewcommand{\arraystretch}{1.3}
  \centering
  \caption{Epoch Number of Convergence.} 
  \begin{small}
  \begin{sc}
  \begin{tabular}{l|ccccc}
  \noalign{\vskip 0.5pt}
  \noalign{\hrule height 0.75pt}
  \noalign{\vskip 0.5pt}
  \multirow{2}{*}{Data set} & \multicolumn{5}{c}{rflaf (rbf) with $N=$} \\ 
      & 8 & 16 & 32 & 64 & 128 \\
  \hline
  MNIST       	& 10 	& 9 	& 4 	& 3 	& 3 	\\
  CIFAR-10    	& 8 	& 3 	& 2 	& 2 	& 2 	\\
  adult       	& 14 	& 4 	& 2 	& 1 	& 1 	\\
  \noalign{\vskip 0.5pt}
  \noalign{\hrule height 0.75pt}
  \noalign{\vskip 0.5pt}
  \end{tabular}
  \end{sc}
  \end{small}
  \label{cvcomp} 
\end{table}

Moreover, larger $N$ equips the model with faster kernel learning ability. In Table \ref{cvcomp}, we consider three tasks where overfitting occurs, and record the epoch number when the model reaches the lowest test error. Table \ref{cvcomp} shows that the epoch number of convergence declines as $N$ increases. This indicates that the model with larger $N$ represents a broader class of function, and hence the model is able to converge to the local minimizer very fast in the training phase. 

To summarize, a small grid number such as $N=16,32,64$ is sufficient for the model to fit the real data well, and is probably a good trade-off between model performance and time efficiency.

\subsection{Ability to Approximate the Optimal Activation Function}

% \begin{table*}[tbp]
%     \vskip -5pt
%     \caption{Test Losses and Time Consumption for Regular Networks. $N = 16$ for LAN and KAN.}
% 	\label{losscomp4}
% 	\begin{center}
% 	\begin{small}
% 	\begin{sc}
% 	\begin{tabular}{l|ccc|c|c|ccc|c|c}
% 	\toprule
%     \multirow{3}{*}{Data set} & \multicolumn{5}{c|}{loss} & \multicolumn{5}{c}{test time} \\
%     \cmidrule(){2-6} \cmidrule(){7-11}
%      & \multicolumn{3}{c|}{lan} &  {mlp} & \multirow{2}{*}{kan} & \multicolumn{3}{c|}{lan} &  {mlp} & \multirow{2}{*}{kan} \\
%     %  \cmidrule(){2-4} \cmidrule(){7-9}
%      & rbf & bs & pl & relu & & rbf & bs & pl & relu & \\
% 	\midrule
%     MNIST       	& 0.180	& \bf{0.102} & 0.176	& 0.171	& \underline{0.162}	& 1.08	& 1.35	& 1.06	& 1.00	& 1.09	\\
%     CIFAR-10    	& \bf{1.343}	& \underline{1.374}	& 1.401	& 1.456	& 1.476	& 1.07	& 1.31	& 1.02	& 1.00	& 1.12	\\
%     adult       	& \bf{0.301}	& 0.308	& 63.521	& \underline{0.305}	& \underline{0.305}	& 1.73	& 5.03	& 1.22	& 1.00	& 1.45	\\
%     \midrule
%     protein     	& \underline{0.218}	& \bf{0.197}	& -	& 0.235	& 0.372	& 1.54	& 4.70	& 0.99	& 1.00	& 1.50	\\
%     ct          	& \underline{0.026}	& \bf{0.023}	& -	& 0.031	& 4.938	& 1.88	& 4.85	& 1.36	& 1.00	& 1.66	\\
%     workloads   	& \underline{0.298}	& \bf{0.283}	& 0.670	& 2.092	& 7.207	& 1.65	& 5.06	& 1.09	& 1.00	& 1.36	\\
%     millionsongs	& \bf{0.076} & \underline{0.080}	& -	& 0.546	& 0.367	& 1.75	& 5.12	& 1.49	& 1.00	& 1.66	\\
% 	\bottomrule
%     \end{tabular}
%     \end{sc}
%     \end{small}
%     \end{center}
%     \vskip -10pt
% \end{table*}

\begin{table*}[tbp]
  \renewcommand{\arraystretch}{1.3}
  \caption{Test Losses of Regular Two-layer Networks. $N = 16$ for LAN and KAN.}
	\label{losscomp4}
	\begin{center}
	\begin{small}
	\begin{sc}
	\begin{tabular}{l|ccc|c|c}
  \noalign{\vskip 0.5pt}      
  \noalign{\hrule height 0.75pt}
  \noalign{\vskip 0.5pt}
  \multirow{3}{*}{Data set} & \multicolumn{5}{c}{loss} \\
  \cline{2-6}
    & \multicolumn{3}{c|}{lan} &  {mlp} & \multirow{2}{*}{kan}  \\ \cline{2-5}
    & rbf & bs & pl & relu &  \\
	\hline
  MNIST       	& $\it{0.124_{\pm 0.040}}$ & $\bf{0.105_{\pm 0.003}}$ & $0.181_{\pm 0.003}$ & $0.166_{\pm 0.003}$ & $0.161_{\pm 0.003}$ \\
  CIFAR-10    	& $\bf{1.349_{\pm 0.010}}$ & $\it{1.375_{\pm 0.007}}$ & $1.484_{\pm 0.115}$ & $1.449_{\pm 0.005}$ & $1.475_{\pm 0.003}$ \\
  adult       	& $\bf{0.302_{\pm 0.001}}$ & $0.308_{\pm 0.002}$ & - & $\it{0.305_{\pm 0.000}}$ & $0.305_{\pm 0.000}$ \\
  \hline
  protein     	& $\it{0.220_{\pm 0.002}}$ & $\bf{0.195_{\pm 0.002}}$ & - & $0.234_{\pm 0.001}$ & $0.372_{\pm 0.000}$ \\
  ct          	& $\it{0.026_{\pm 0.000}}$ & $\bf{0.023_{\pm 0.002}}$ & - & $0.031_{\pm 0.001}$ & $4.938_{\pm 0.000}$ \\
  workloads   	& $\it{0.285_{\pm 0.012}}$ & $\bf{0.266_{\pm 0.021}}$ & $0.956_{\pm 0.276}$ & $2.090_{\pm 0.025}$ & $7.207_{\pm 0.000}$ \\
  millionsongs	& $\bf{0.077_{\pm 0.002}}$ & $\it{0.082_{\pm 0.001}}$ & - & $0.550_{\pm 0.007}$ & $0.367_{\pm 0.000}$ \\
  \noalign{\vskip 0.5pt}      
  \noalign{\hrule height 0.75pt}
  \noalign{\vskip 0.5pt}
  \end{tabular}
  \end{sc}
  \end{small}
  \end{center}
\end{table*}

\begin{figure}[htbp]
    \centering
    \includegraphics[width=0.75\linewidth]{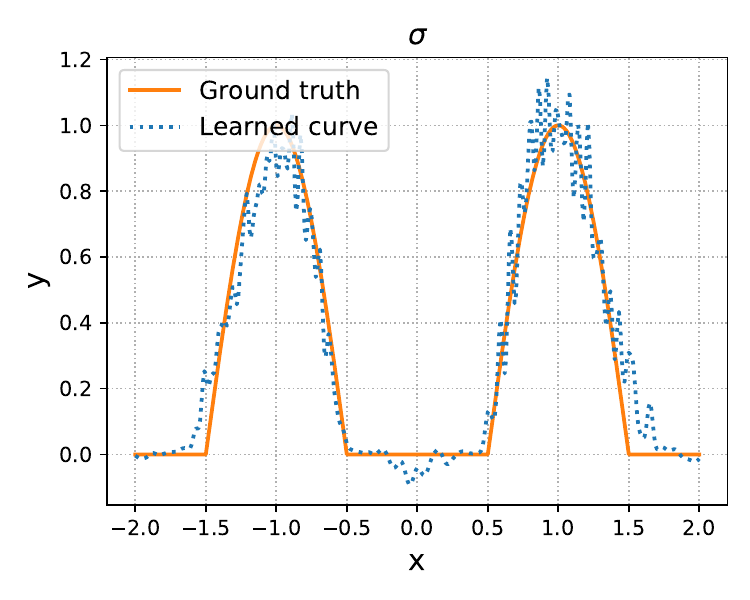}
    \caption{The Activation Function Learned in RFLAF.}
    \label{synactf}
\end{figure}

To further examine the kernel learning ability of RFLAF with \texttt{RBF}, we design synthetic functions to check if RFLAF successfully recovers the ground-truth activation function from data. We choose target functions to be of the form $f(x)=\mathbb{E}_{w\sim \mathcal{N}(0,I_d)}\left[\sigma(w^\top x)v(w)\right],$ where $x\in\mathbb{R}^d$ and $d=2$. For RFLAF with \texttt{RBF}, we set $M=1000$ and $N=400$, $h=0.005$ for the precise learning of the optimal activation. We show the result of 
\[
\begin{split}
\sigma(x)=&-\sin(\pi(x+0.5))\bm{1}_{[-1.5,-0.5]} \\
&+  \sin(\pi (x-0.5))\bm{1}_{[0.5,1.5]}
\end{split}
\]
in Figure \ref{synactf} as an example. 

The blue dotted line represents the learned activation functions and the orange solid line represents the ground-truth function. RFLAF successfully recovers the ground truth activation functions \(\sigma\) through learning \(f\), demonstrating its ability to directly learn from data the optimal activation function and skip the step for grid searching the optimal parameters of the activation function as proposed in \citep{demir2024optimal}. We repeat the experiment with another two synthetic functions. The results are similar and are provided in Appendix \ref{E1}.

\subsection{Comparisons among the Regular Two-layer Networks}

% \begin{table*}[tbp]
%   \renewcommand{\arraystretch}{1.3}
%   \caption{Test Losses and Time Consumption for Regular Networks. $N = 16$ for LAN and KAN.}
% 	\label{timecomp4}
% 	\begin{center}
% 	\begin{small}
% 	\begin{sc}
% 	\begin{tabular}{l|ccc|c|c}
% 	\noalign{\vskip 0.5pt}      
%   \noalign{\hrule height 0.75pt}
%   \noalign{\vskip 0.5pt}
%   \multirow{3}{*}{Data set} & \multicolumn{5}{c}{test time} \\
%   \cline{2-6}
%   & \multicolumn{3}{c|}{lan} &  {mlp} & \multirow{2}{*}{kan} \\
%   & rbf & bs & pl & relu & \\
% 	\hline
%   MNIST       	& $1.1$ & $1.4$ & $1.0$ & $1.0$ & $1.1$ \\
%   CIFAR-10    	& $1.1$ & $1.3$ & $1.0$ & $1.0$ & $1.1$ \\
%   adult       	& $2.2$ & $6.0$ & - & $1.0$ & $1.9$ \\
%   \hline
%   protein     	& $1.7$ & $5.1$ & - & $1.0$ & $1.8$ \\
%   ct          	& $1.8$ & $4.6$ & - & $1.0$ & $1.7$ \\
%   workloads   	& $2.0$ & $5.5$ & $1.4$ & $1.0$ & $1.7$ \\
%   millionsongs	& $1.9$ & $5.2$ & - & $1.0$ & $1.7$ \\
% 	\noalign{\vskip 0.5pt}      
%   \noalign{\hrule height 0.75pt}
%   \noalign{\vskip 0.5pt}
%   \end{tabular}
%   \end{sc}
%   \end{small}
%   \end{center}
% \end{table*}

In the last part, we unfreeze the first-layer parameters of all RF models, where RFLAF becomes Learnable Activation Networks (LAN), and test them over all datasets. We include the novel neural network KAN for comparison. The width of KAN is set $M/N$ to ensure the same parameter number among models. 

Table \ref{losscomp4} shows that LAN with \texttt{RBF} or \texttt{BS} generally achieves the best performance, and Table \ref{timecomp4} shows that \texttt{RBF} runs about $3$ times faster than \texttt{BS}, consistent with the results of the random feature cases. We highlight that LAN, as an intermediate form between MLP and KAN, showcases its unique characteristics. KAN has superior interpretability in small-scale and science-related tasks, however, it suffers from scaling problem that LAN handles more easily. On the other side, LAN enhances the capability of MLP with economic increase in time and parameter number. This result further substantiates the potential of learnable activation in modern neural network structure. More discussions are provided in Appendix \ref{E4}.

\begin{table}[htbp]
  \renewcommand{\arraystretch}{1.3}
  \caption{\centering{Test Time Comparison among\\ Regular Networks. $N = 16$ for LAN and KAN.}}
	\label{timecomp4}
	\begin{center}
	\begin{small}
	\begin{sc}
	\begin{tabular}{l|cc|c|c}
	\noalign{\vskip 0.5pt}      
  \noalign{\hrule height 0.75pt}
  \noalign{\vskip 0.5pt}
  \multirow{3}{*}{Data set} & \multicolumn{4}{c}{test time} \\
  \cline{2-5}
  & \multicolumn{2}{c|}{lan} &  {mlp} & \multirow{2}{*}{kan} \\ \cline{2-4}
  & rbf & bs & relu & \\
	\hline
  MNIST       	& $1.1$ & $1.4$ & $1.0$ & $1.1$ \\
  CIFAR-10    	& $1.1$ & $1.3$ & $1.0$ & $1.1$ \\
  adult       	& $2.2$ & $6.0$ & $1.0$ & $1.9$ \\
  \hline
  protein     	& $1.7$ & $5.1$ & $1.0$ & $1.8$ \\
  ct          	& $1.8$ & $4.6$ & $1.0$ & $1.7$ \\
  workloads   	& $2.0$ & $5.5$ & $1.0$ & $1.7$ \\
  millionsongs	& $1.9$ & $5.2$ & $1.0$ & $1.7$ \\
	\noalign{\vskip 0.5pt}      
  \noalign{\hrule height 0.75pt}
  \noalign{\vskip 0.5pt}
  \end{tabular}
  \end{sc}
  \end{small}
  \end{center}
\end{table}

% \vspace{-12pt}
\section{CONCLUSION}
\label{sec7}
% In this work, we propose the random feature model with learnable activation functions. We provide theoretical guarantees and showcase its superior performance in practice. Our work deepens the comprehension of the module of learnable activation, and initiates an array of open problems for future work, including the derivation of tighter bounds and the model inductive bias through the eigenstructure of the kernel.
In this work, we propose the Random Feature Model with Learnable Activation Functions (RFLAF), which parameterizes activation functions as weighted sums of basis functions. Theoretically, we provide the first analytic form of the kernel induced by a single RBF as the activation and establish approximation and generalization bounds for RFLAF. Experimentally, RFLAF with RBFs or B-splines consistently outperforms standard RF models, with RBFs being three times more efficient. Our work deepens the understanding of learnable activation functions in neural architectures and opens future directions for random feature models.

\begin{acknowledgements} % will be removed in pdf for initial submission,
						 % (without ‘accepted’ option in \documentclass)
                         % so you can already fill it to test with the
                         % ‘accepted’ class option
    This work is sponsored by the National Natural Science Foundation of China (62376013, 623B2003, 624B100026). Any opinions, findings, conclusions, or recommendations expressed in this material are those of the author(s) and do not necessarily reflect the views of the funding agencies.
\end{acknowledgements}

% References
\bibliography{uai2026-template}

\newpage

\onecolumn

\title{Learning Expressive Random Feature Models via Parametrized Activations\\(Supplementary Material)}
\begingroup
\makeatletter
\let\thanks\@gobble
\maketitle
\global\let\@thanks\@empty
\makeatother
\endgroup

\newpage

\appendix

\section{TECHNICAL TOOLS}
\subsection{Basics on Sub-gaussian and Sub-exponential Random Variables}

\begin{definition}
	A random variable $Y$ is a sub-gaussian random variable if there exists $K>0$ such that $\mathbb{E}\exp\left(Y^2/K^2\right)\leq 2$. Define the sub-gaussian norm as $\|Y\|_{\psi_2}:=\inf\{K>0:\mathbb{E}\exp\left(Y^2/K^2\right)\leq 2\}$.
\end{definition}
\begin{definition}
	A random variable $Y$ is a sub-exponential random variable if there exists $K>0$ such that $\mathbb{E}\exp\left(|Y|/K\right)\leq 2$. Define the sub-exponential norm as $\|Y\|_{\psi_1}:=\inf\{K>0:\mathbb{E}\exp\left(|Y|/K\right)\leq 2\}$.
\end{definition}
\begin{lemma}
	\label{normeq}
	If $Y$ is a sub-gaussian random variable, then $\|Y^2\|_{\psi_1}=\|Y\|_{\psi_2}^2$.
\end{lemma}

The following properties of sub-gaussian random variable are stated in Proposition 2.5.2 in \citep{vershynin2018high}. For this paper to be self-contained, we also state them here with explicit constants.
\begin{lemma}
	\label{subg0}
	Suppose $Y$ is a random variable.
	\begin{enumerate}
		\item If $Y$ is a sub-gaussian random variable, then $P(|Y|\geq \epsilon)\leq 2\exp\left(-\epsilon^2/\|Y\|_{\psi_2}^2\right)$.
		\item If $P(|Y|\geq \epsilon)\leq 2\exp\left(-\epsilon^2/K^2\right)$, then $\|Y\|_{\psi_2}\leq \sqrt{2}K$.
	\end{enumerate}
\end{lemma}

\begin{lemma}
	\label{subg}
	Suppose $Y$ is a random variable.
	\begin{enumerate}
		\item If there exist $K_0>0$ such that $\mathbb{E}e^{\lambda^2Y^2}\leq e^{K_0^2\lambda^2 }$ for all $\lambda^2\leq 1/K_0^2$, then $Y$ is a sub-gaussian random variable with sub-gaussian norm $\|Y\|_{\psi_2}\leq {K_0}/{\sqrt{\log 2}}\leq \sqrt{2}K_0$.
		\item If $Y$ is a sub-gaussian random variable, then $K_0=2\|Y\|_{\psi_2}$ such that $\mathbb{E}e^{\lambda^2Y^2}\leq e^{K_0^2\lambda^2 }$ for all $\lambda^2\leq 1/K_0^2$.
	\end{enumerate}
\end{lemma} 

\begin{lemma}
	\label{subg2}
	Suppose $Y$ is a random variable and $\mathbb{E}Y=0$.
	\begin{enumerate}
		\item If $\mathbb{E}e^{\lambda^2Y^2}\leq e^{K_0^2\lambda^2 }$ for all $\lambda^2\leq 1/K_0^2$, then $\mathbb{E}e^{\lambda Y}\leq e^{K_0^2\lambda^2 }$ for all $\lambda\in\mathbb{R}$.
		\item If $\mathbb{E}e^{\lambda Y}\leq e^{K_0^2\lambda^2 }$ for all $\lambda\in\mathbb{R}$, then $\mathbb{E}e^{\lambda^2Y^2}\leq e^{16 K_0^2\lambda^2 }$ for all $\lambda^2\leq 1/16K_0^2$.
	\end{enumerate}
\end{lemma}

For sums of independent sub-gaussian random variables, the Proposition 2.6.1 in \citep{vershynin2018high} states that
\begin{lemma}
	\label{sumsubg}
	Let $X_1,...,X_M$ be independent copies of a sub-gaussian random variable $X$ and $\mathbb{E}X=0$. Then
	\[
		\left\|\sum_{m=1}^M X_m\right\|_{\psi_2} \leq 4\sqrt{M}\left\| X\right\|_{\psi_2}.
	\]
\end{lemma}

We also state a concentration inequality for sums of independent sub-exponential random variables.
\begin{lemma}[Bernstein's inequality (e.g., Theorem 2.8.1 in \citep{vershynin2018high})]
	\label{bern}
	Let $X_1,...,X_M$ be independent copies of a sub-exponential random variable $X$ and $\mathbb{E}X=0$. Then, for every $t>0$, we have
	\[
		P\left(\frac{1}{M}\sum_{m=1}^{M}X_m>t\right)\leq \exp\left(-\min\left\{\frac{Mt^2}{16\|X\|_{\psi_1}^2},\frac{Mt}{4\|X\|_{\psi_1}}\right\}\right).
	\]
\end{lemma}

\newpage
\section{DEFERRED PROOFS IN SECTION 3}
% eferred proof in section 3}
\label{proofsec3}

\subsection{Proof of Theorem \ref{pro31}}
\label{proofpro31}
\begin{proof}
	For the simplicity of calculation, we denote
	\[
		s:=\|x\|_2,\quad s^\prime:=\|x^\prime\|_2,\quad \rho=\frac{\langle x, x^\prime \rangle}{\|x\|_2\|x^\prime\|_2}.
	\]
	The statistical properties of Gaussian distribution indicate that
	\begin{equation*}
		\mathbb{E}_{w\sim\mathcal{N}(0,I_d)}\left[B(w^\top x)B(w^\top x^\prime)\right]=\mathbb{E}_{x\sim \mathcal{N}(0,1),z\sim \mathcal{N}(0,1-\rho^2)}\left[B(sx)B(s^\prime(\rho x+z))\right],
	\end{equation*}
	where $x,z$ are two independent Gaussian random variables. Then we do the calculation based on the latter expression. The calculation is quite complicated, so we illustrate it here in a detailed way.
	\begin{equation*}
		\begin{aligned}
			&\mathbb{E}_{x\sim \mathcal{N}(0,1),z\sim \mathcal{N}(0,1-\rho^2)}\left[B(sx)B(s^\prime(\rho x+z))\right]\\
			=&{1\over \sqrt{2\pi}}{1\over \sqrt{2\pi(1-\rho^2)}}\iint_{\mathbb{R}^2}\exp{\left(-{(sx-c)^2\over 2h^2}\right)}\exp{\left(-{(s^\prime(\rho x+z)-c)^2\over 2h^2}\right)}\\
			&\exp{\left(-{x^2\over 2}\right)}\exp{\left(-{z^2\over 2(1-\rho^2)}\right)}dxdz\\
			=&{1\over \sqrt{2\pi}}{1\over \sqrt{2\pi(1-\rho^2)}}\iint_{\mathbb{R}^2}\exp \Bigg(-{1\over 2h^2(1-\rho^2)} \bigg[(1-\rho^2)(sx-c)^2+(1-\rho^2)(s^\prime\rho x-c+s^\prime z)^2\\
			& +(1-\rho^2) h^2 x^2+h^2 z^2\bigg] \Bigg)dxdz\\
			=&{1\over \sqrt{2\pi}}{1\over \sqrt{2\pi(1-\rho^2)}}\iint_{\mathbb{R}^2}\exp \Bigg(-{1\over 2h^2(1-\rho^2)} \bigg[(1-\rho^2)(sx-c)^2+(1-\rho^2)(s^\prime\rho x-c)^2+(1-\rho^2)(s^\prime z)^2\\
			& +2(1-\rho^2)(s^\prime\rho x-c)s^\prime z+(1-\rho^2) h^2 x^2+h^2 z^2\bigg] \Bigg)dxdz\\
			=&{1\over \sqrt{2\pi}}{1\over \sqrt{2\pi(1-\rho^2)}}\iint_{\mathbb{R}^2}\exp \Bigg(-{1\over 2h^2(1-\rho^2)} \bigg[(1-\rho^2)(sx-c)^2+(1-\rho^2)(s^\prime\rho x-c)^2+(1-\rho^2) h^2 x^2\\
			& +[(1-\rho^2)(s^\prime )^2+h^2] z^2 + 2(1-\rho^2)(s^\prime\rho x-c)s^\prime z\bigg] \Bigg)dxdz\\
			=&{1\over \sqrt{2\pi}}{1\over \sqrt{2\pi(1-\rho^2)}}\iint_{\mathbb{R}^2}\exp \Bigg(-{1\over 2h^2(1-\rho^2)} \bigg[(1-\rho^2)(sx-c)^2+(1-\rho^2)(s^\prime\rho x-c)^2+(1-\rho^2) h^2 x^2\\
			& +[(1-\rho^2)(s^\prime )^2+h^2] \bigg[z + {(1-\rho^2)(s^\prime\rho x-c)s^\prime\over (1-\rho^2)(s^\prime )^2+h^2} \bigg]^2-{(1-\rho^2)^2(s^\prime\rho x-c)^2(s^\prime)^2\over (1-\rho^2)(s^\prime )^2+h^2}\bigg] \Bigg)dxdz\\
			=&{1\over \sqrt{2\pi}}{1\over \sqrt{2\pi(1-\rho^2)}}\\
			&\int_{\mathbb{R}}\exp \Bigg(-{(1-\rho^2)(sx-c)^2+(1-\rho^2)(s^\prime\rho x-c)^2+(1-\rho^2) h^2 x^2-{(1-\rho^2)^2(s^\prime\rho x-c)^2(s^\prime)^2\over (1-\rho^2)(s^\prime )^2+h^2}\over 2h^2(1-\rho^2)}  \Bigg)dx\\
			& \int_{\mathbb{R}}\exp \Bigg(-{\bigg[z + {(1-\rho^2)(s^\prime\rho x-c)s^\prime\over (1-\rho^2)(s^\prime )^2+h^2} \bigg]^2\over 2\frac{h^2(1-\rho^2)}{(1-\rho^2)(s^\prime )^2+h^2}}  \Bigg)dz\\
			=&{1\over \sqrt{2\pi}}{\sqrt{\frac{h^2(1-\rho^2)}{(1-\rho^2)(s^\prime )^2+h^2}}\over \sqrt{(1-\rho^2)}}\int_{\mathbb{R}}\exp \Bigg(-{(1-\rho^2)(sx-c)^2+(1-\rho^2)(s^\prime\rho x-c)^2+(1-\rho^2) h^2 x^2-{(1-\rho^2)^2(s^\prime\rho x-c)^2(s^\prime)^2\over (1-\rho^2)(s^\prime )^2+h^2}\over 2h^2(1-\rho^2)}  \Bigg)dx.\\
		\end{aligned}
	\end{equation*}
	
	To continue, we have
	\begin{equation*}
		\begin{aligned}
			&\mathbb{E}_{x\sim \mathcal{N}(0,1),z\sim \mathcal{N}(0,1-\rho^2)}\left[B(sx)B(s^\prime(\rho x+z))\right]\\
			=&{1\over \sqrt{2\pi}}\frac{h}{\sqrt{(1-\rho^2)(s^\prime )^2+h^2}}\\
			&\int_{\mathbb{R}}\exp \Bigg(-{(1-\rho^2)(sx-c)^2+(1-\rho^2)(s^\prime\rho x-c)^2+(1-\rho^2) h^2 x^2-{(1-\rho^2)^2(s^\prime\rho x-c)^2(s^\prime)^2\over (1-\rho^2)(s^\prime )^2+h^2}\over 2h^2(1-\rho^2)}  \Bigg)dx\\
			=& {1\over \sqrt{2\pi}}\frac{h}{\sqrt{(1-\rho^2)(s^\prime )^2+h^2}}\\
			&\int_{\mathbb{R}}\exp \Bigg(-{(sx-c)^2+(s^\prime\rho x-c)^2+ h^2 x^2-{(1-\rho^2)(s^\prime)^2(s^\prime\rho x-c)^2\over (1-\rho^2)(s^\prime )^2+h^2}\over 2h^2}  \Bigg)dx\\
			=& {1\over \sqrt{2\pi}}\frac{h}{\sqrt{(1-\rho^2)(s^\prime )^2+h^2}}\\
			&\int_{\mathbb{R}}\exp \Bigg(-{[(1-\rho^2)(s^\prime )^2+h^2][(sx-c)^2+h^2x^2]+h^2(s^\prime\rho x-c)^2\over 2h^2[(1-\rho^2)(s^\prime )^2+h^2]}  \Bigg)dx\\
			=& {1\over \sqrt{2\pi}}\frac{h}{\sqrt{(1-\rho^2)(s^\prime )^2+h^2}}\\
			&\int_{\mathbb{R}}\exp \Bigg(-{[(1-\rho^2)(s^\prime )^2+h^2][(s^2+h^2)x^2-2scx+c^2]+h^2[(s^\prime)^2\rho^2 x^2-2s^\prime\rho c x +c^2]  \over 2h^2[(1-\rho^2)(s^\prime )^2+h^2]}  \Bigg)dx\\
			=& {1\over \sqrt{2\pi}}\frac{h}{\sqrt{(1-\rho^2)(s^\prime )^2+h^2}}\\
			&\int_{\mathbb{R}}\exp \Bigg(-{[h^4+(1-\rho^2)s^2(s^\prime )^2+h^2(s^2+(s^\prime)^2)]x^2 \over 2h^2[(1-\rho^2)(s^\prime )^2+h^2]}-\\
			&{-2[(1-\rho^2)(s^\prime )^2s+h^2s+h^2s^\prime\rho] c x +[(1-\rho^2)(s^\prime )^2+2h^2]c^2 \over 2h^2[(1-\rho^2)(s^\prime )^2+h^2]}  \Bigg)dx\\
			=& {1\over \sqrt{2\pi}}\frac{h}{\sqrt{(1-\rho^2)(s^\prime )^2+h^2}}\\
			&\int_{\mathbb{R}}\exp \Bigg(-{[h^4+(1-\rho^2)s^2(s^\prime )^2+h^2(s^2+(s^\prime)^2)]\left(x-\frac{[(1-\rho^2)(s^\prime )^2s+h^2s+h^2s^\prime\rho] c}{h^4+(1-\rho^2)s^2(s^\prime )^2+h^2(s^2+(s^\prime)^2)}\right)^2 \over 2h^2[(1-\rho^2)(s^\prime )^2+h^2]}-\\
			&{-\frac{[(1-\rho^2)(s^\prime )^2s+h^2s+h^2s^\prime\rho]^2 c^2}{h^4+(1-\rho^2)s^2(s^\prime )^2+h^2(s^2+(s^\prime)^2)} +[(1-\rho^2)(s^\prime )^2+2h^2]c^2 \over 2h^2[(1-\rho^2)(s^\prime )^2+h^2]}  \Bigg)dx\\
			=& {h\sqrt{(1-\rho^2)(s^\prime )^2+h^2}\over \sqrt{h^4+(1-\rho^2)s^2(s^\prime )^2+h^2(s^2+(s^\prime)^2)}}\frac{h}{\sqrt{(1-\rho^2)(s^\prime )^2+h^2}}\\
			&\exp \Bigg({\frac{[(1-\rho^2)(s^\prime )^2s+h^2s+h^2s^\prime\rho]^2 c^2}{h^4+(1-\rho^2)s^2(s^\prime )^2+h^2(s^2+(s^\prime)^2)} -[(1-\rho^2)(s^\prime )^2+2h^2]c^2 \over 2h^2[(1-\rho^2)(s^\prime )^2+h^2]}  \Bigg)\\
			=& {h^2\over \sqrt{(h^2+s^2)(h^2+(s^\prime)^2)-\rho^2s^2(s^\prime )^2}}\exp \Bigg({\frac{[(1-\rho^2)(s^\prime )^2s+h^2s+h^2s^\prime\rho]^2 c^2}{h^4+(1-\rho^2)s^2(s^\prime )^2+h^2(s^2+(s^\prime)^2)} -[(1-\rho^2)(s^\prime )^2+2h^2]c^2 \over 2h^2[(1-\rho^2)(s^\prime )^2+h^2]}  \Bigg).\\
		\end{aligned}
	\end{equation*}
	
	For the exponential term, we calculate as follows.

	\begin{equation*}
		\begin{aligned}
			&{\frac{[(1-\rho^2)(s^\prime )^2s+h^2s+h^2s^\prime\rho]^2 c^2}{h^4+(1-\rho^2)s^2(s^\prime )^2+h^2(s^2+(s^\prime)^2)} -[(1-\rho^2)(s^\prime )^2+h^2+h^2]c^2 \over 2h^2[(1-\rho^2)(s^\prime )^2+h^2]}\\
			=& -\frac{c^2}{2h^2}\left(1-\frac{[(1-\rho^2)(s^\prime )^2s+h^2s+h^2s^\prime\rho]^2-h^2[h^4+(1-\rho^2)s^2(s^\prime )^2+h^2(s^2+(s^\prime)^2)]}{[(1-\rho^2)(s^\prime )^2+h^2][h^4+(1-\rho^2)s^2(s^\prime )^2+h^2(s^2+(s^\prime)^2)]}\right)\\
			=& -\frac{c^2}{2h^2}\left(1-\frac{[((1-\rho^2)(s^\prime )^2+h^2)s+h^2s^\prime\rho]^2-h^2[[(1-\rho^2)(s^\prime)^2+h^2]s^2+h^2(h^2+(s^\prime)^2)]}{[(1-\rho^2)(s^\prime )^2+h^2][(h^2+s^2)(h^2+(s^\prime)^2)-\rho^2s^2(s^\prime )^2]}\right)\\
			=&-\frac{c^2}{2h^2}\Bigg(1-\frac{[((1-\rho^2)(s^\prime )^2+h^2)^2s^2+h^4(s^\prime)^2\rho^2+2((1-\rho^2)(s^\prime )^2+h^2)h^2ss^\prime\rho]}{[(1-\rho^2)(s^\prime )^2+h^2][(h^2+s^2)(h^2+(s^\prime)^2)-\rho^2s^2(s^\prime )^2]}\\
			&\quad\quad\quad\quad~ -\frac{-h^2[(1-\rho^2)(s^\prime)^2+h^2]s^2-h^4(h^2+(s^\prime)^2)}{[(1-\rho^2)(s^\prime )^2+h^2][(h^2+s^2)(h^2+(s^\prime)^2)-\rho^2s^2(s^\prime )^2]}\Bigg)\\
			=&-\frac{c^2}{2h^2}\left(1-\frac{[(1-\rho^2)(s^\prime )^2+h^2]\{[(1-\rho^2)(s^\prime )^2+h^2]s^2+2h^2ss^\prime\rho-h^2s^2\}-h^4[(1-\rho^2)(s^\prime)^2+h^2]}{[(1-\rho^2)(s^\prime )^2+h^2][(h^2+s^2)(h^2+(s^\prime)^2)-\rho^2s^2(s^\prime )^2]}\right)\\
			=&-\frac{c^2}{2h^2}\left(1-\frac{[(1-\rho^2)(s^\prime )^2+h^2]\{s^2(s^\prime )^2-\rho^2s^2(s^\prime )^2+2h^2\rho ss^\prime-h^4\}}{[(1-\rho^2)(s^\prime )^2+h^2][(h^2+s^2)(h^2+(s^\prime)^2)-\rho^2s^2(s^\prime )^2]}\right)\\
			=&-\frac{c^2}{2h^2}\left(1-\frac{s^2(s^\prime )^2-(h^2-\rho ss^\prime)^2}{(h^2+s^2)(h^2+(s^\prime)^2)-\rho^2s^2(s^\prime )^2}\right).\\
		\end{aligned}
	\end{equation*}

	Combining the former results, we obtain
	\begin{equation*}
		\begin{aligned}
			&\mathbb{E}_{x\sim \mathcal{N}(0,1),z\sim \mathcal{N}(0,1-\rho^2)}\left[B(sx)B(s^\prime(\rho x+z))\right]\\
			=& {h^2\over \sqrt{(h^2+s^2)(h^2+(s^\prime)^2)-\rho^2s^2(s^\prime )^2}}\exp \Bigg(-\frac{c^2}{2h^2}\left(1-\frac{s^2(s^\prime)^2-(h^2-\rho s s^\prime)^2}{(h^2+s^2)(h^2+(s^\prime)^2)-\rho^2s^2(s^\prime)^2}\right)  \Bigg)\\
			=& {h^2\over \sqrt{(h^2+s^2)(h^2+(s^\prime)^2)-\rho^2s^2(s^\prime )^2}}\exp \Bigg(-\frac{c^2}{2}\cdot \frac{(h^2+s^2)+(h^2+(s^\prime)^2)-2\rho s (s^\prime)}{(h^2+s^2)(h^2+(s^\prime)^2)-\rho^2s^2(s^\prime)^2}  \Bigg).\\
		\end{aligned}
	\end{equation*}

	Then using the relations
	\[
		s=\|x\|_2,\quad s^\prime=\|x^\prime\|_2,\quad \rho s s^\prime=\langle x,x^\prime\rangle,
	\]
	we obtain
	\begin{equation*}
		\begin{aligned}
			&\mathbb{E}_{x\sim \mathcal{N}(0,1),z\sim \mathcal{N}(0,1-\rho^2)}\left[B(sx)B(s^\prime(\rho x+z))\right]\\
			=& {h^2\over \sqrt{(h^2+\|x\|^2)(h^2+\|x^\prime\|^2)-\langle x, x^\prime\rangle^2}}\exp \Bigg(-\frac{c^2}{2}\cdot \frac{(h^2+\|x\|^2)+(h^2+\|x^\prime\|^2)-2\langle x, x^\prime\rangle}{(h^2+\|x\|^2)(h^2+\|x^\prime\|^2)-\langle x, x^\prime\rangle^2}  \Bigg).\\
		\end{aligned}
	\end{equation*}
\end{proof}

\subsection{Proof of Theorem \ref{pro32}}
\label{proofpro32}
\begin{proof}

	\textbf{Step 1. Transform $K(r)$.}

	To obtain a uniform expression regardless of $h$, we transform $K(r)$ in the following manner.
	\[
		\begin{split}
			K(r)&={h^2\over \sqrt{(1+h^2)^2-r^2}}\exp \left(-\frac{c^2}{1+h^2+r}  \right)\\
			&=\frac{h^2}{1+h^2}\frac{1}{\sqrt{1-\left(\frac{r}{1+h^2}\right)^2}}\exp \left(-\frac{{c^2\over 1+h^2}}{1+\frac{r}{1+h^2}}  \right).
		\end{split}
	\]
	Let $p={c^2\over 1+h^2}\in [0,+\infty)$, $u=\frac{r}{1+h^2}\in [-\frac{1}{1+h^2},\frac{1}{1+h^2}]\subsetneq (-1,1)$, and
	\[
		f(u):=\frac{1}{\sqrt{1-u^2}}\exp\left(-\frac{p}{1+u}\right).
	\]
	Then
	\[
		K(r)=\frac{h^2}{1+h^2}f(u).
	\]
	Hence we only need to consider the Taylor expansion of $f(u)$ where $u\in(-1,1)$.

	\[
		f(u)=\sum_{n=0}^{\infty}\frac{f^{(n)}(0)}{n!}u^n.
	\]

	\paragraph{Step 2. Deriving the recurrence relation of $f^{(n)}(0)$.}

	Solving the Taylor coefficients of $f(u)$ at $u=0$ is highly technical. For starters, we derive the recurrence formula. For notational convenience, let $y=f(u)$.

	From the definition of $y$, we have the equality
	\[
		y\sqrt{1-u^2}=\exp\left(-\frac{p}{1+u}\right).
	\]

	Taking derivatives on both sides, we have
	\[
		\begin{split}
		y^\prime\sqrt{1-u^2} -\frac{uy}{\sqrt{1-u^2}} &= \frac{p}{(1+u)^2}\exp\left(-\frac{p}{1+u}\right)\\
		&=\frac{p\sqrt{1-u^2}}{(1+u)^2}y.
		\end{split}
	\]

	Multiplying $\frac{(u-1)^2(u+1)^2}{\sqrt{1-u^2}}$ on both sides, we have
	\[
			(u^2-1)^2 y^\prime +u(u^2-1)y=p(u-1)^2y.
	\]
	Eliminating the factor $(u-1)$ and expanding the polynomials lead to
	\[
		(u^3+u^2-u-1)y^\prime +(u^2+u)y=p(u-1)y.
	\]
	Taking $n$-th derivatives on both sides and applying the Leibniz rule, we have
	\[
		\begin{array}{lll}
			y^{(n+1)}(u^3+u^2-u-1) &+y^{(n)}(u^2+u) & \\
			+ny^{(n)}(3u^2+2u-1)  &+ny^{(n-1)}(2u+1) =&y^{(n)}p(u-1)\\
			+\frac{n(n-1)}{2}y^{(n-1)}(6u+2) &+\frac{n(n-1)}{2}y^{(n-2)}\cdot 2 &+ny^{(n-1)}p.\\
			+\frac{n(n-1)(n-2)}{6}y^{(n-2)}\cdot6 & &
		\end{array}
	\]
	Let $u=0$, and let $y^{(n)}=y^{(n)}(0)$ in the statements hereafter, we have that
	\[
		-y^{(n+1)}-ny^{(n)}+n^2y^{(n-1)}+n(n-1)^2y^{(n-2)}=-py^{(n)}+npy^{(n-1)}.
	\]
	Finally, we have the recurrence formula.
	\begin{equation}
		\label{rec}
		y^{(n+1)}=(p-n)y^{(n)}-n(p-n)y^{(n-1)}+n(n-1)^2y^{(n-2)}.
	\end{equation}

	To solve $\{y^{(n)}(0)\}_{n=0}^{\infty}$ from the recurrence relation, we also need to obtain $y(0),y^\prime(0),y^{\prime\prime}(0)$ by hand. A simple calculation shows that
	\[
		\begin{split}
			f(u)=&\frac{1}{\sqrt{1-u^2}}\exp\left(-\frac{p}{1+u}\right)\\
			f^{\prime}(u)=&\left(\frac{u}{\sqrt{(1-u^2)^3}}+\frac{1}{\sqrt{1-u^2}}\cdot\frac{p}{(1+u)^2}\right)\exp\left(-\frac{p}{1+u}\right)\\
			f^{\prime\prime}(u)=&\Bigg(\frac{1}{\sqrt{(1-u^2)^3}}+\frac{3u^2}{\sqrt{(1-u^2)^5}}\\
			&+\frac{u}{\sqrt{(1-u^2)^3}}\cdot\frac{p}{(1+u)^2}+\frac{1}{\sqrt{1-u^2}}\cdot\frac{-2p}{(1+u)^3}+\\
			&\left(\frac{u}{\sqrt{(1-u^2)^3}}+\frac{1}{\sqrt{1-u^2}}\cdot\frac{p}{(1+u)^2}\right)\frac{p}{(1+u)^2}\Bigg)\exp\left(-\frac{p}{1+u}\right).\\
		\end{split}
	\]
	Hence, we obtain
	\[
		\begin{split}
			y^{(0)}=&e^{-p},\\
			y^{(1)}=&pe^{-p},\\
			y^{(2)}=&(p-1)^2e^{-p}.\\
		\end{split}
	\]

	Solving $\{y^{(n)}\}_{n=0}^{\infty}$ remains to be difficult. To simplify the problem, we try to make some observations on the properties of $y^{(n)}$. We supplement $y^{(n)}$ till the first $8$ terms.
	\[
		\begin{split}
			y^{(3)}&=(p^3-6p^2+9p) e^{-p},\\
			y^{(4)}&=(p^4-12p^3+42p^2-36p+9)e^{-p},\\
			y^{(5)}&=(p^5-20p^4+130p^3-300p^2+225p)e^{-p},\\
			y^{(6)}&=(p^6-30p^5+315p^4-1380p^3+2475p^2-1350p+225)e^{-p},\\
			y^{(7)}&=(p^7-42p^6 +651p^5-4620p^4+15435p^3-22050p^2+11025p)e^{-p},\\
			y^{(8)}&=(p^8-56p^7 +1204p^6-12600p^5+67830p^4-182280p^3+220500p^2-88200p+11025)e^{-p}.\\
		\end{split}
	\]
	A further observation shows that
	\[
		\begin{split}
			y^{(3)}
			&=p(p-3)^2 e^{-p},\\
			y^{(4)}
			&=(p^2-6p+3)^2e^{-p},\\
			y^{(5)}
			&=p(p^2-10p+15)^2e^{-p},\\
			y^{(6)}
			&=(p^3-15p^2+45p-15)^2e^{-p},\\
			y^{(7)}
        	&=p(p^3-21p^2+105p-105)^2e^{-p},\\
			y^{(8)}
			&=(p^4-28p^3+210p^2-420p+105)^2e^{-p}.
		\end{split}
	\]
	To conclude, we have the following observations.
	\begin{enumerate}
		\item $y^{(n)}=e^{-p}R_n(p)$, where $R_n(p)$ is a polynomial of degree $n$.
		\item For $n=2k$, $R_n(p)=P_k^2(p)$, where $P_k(p)$ is a polynomial of degree $k$.
		\item For $n=2k+1$, $R_n(p)=p\cdot Q^2_k(p)$, where $Q_k(p)$ is a polynomial of degree $k$.
	\end{enumerate}
	The correctness of the first observation is easily proved by induction. In the next step, we give a formal proof of the correctness of the second and third observations.
	\paragraph{Step 3. Formal proof of the general term formula of the Taylor coefficients.}
	The intuition of the proof is to directly derive the general term formula of $\{P_k\}$ and $\{Q_k\}$ from observations. Note that the observations are non-trivial.
	
	We claim that
	\begin{equation}
		\label{exP}
		\begin{aligned}
			P_k(x)
			%  =& \sum_{i=0}^{k} (-1)^{k-i}\frac{(2k-1)!!}{(2i-1)!!}\cdot\frac{k!}{i!(k-i)!}x^{i}\\
			=&\sum_{i=0}^{k} (-1)^{k-i}\frac{(2k-1)!!}{(2i-1)!!}\cdot\binom{k}{i} x^{i},
		\end{aligned}
	\end{equation}
	and
	\begin{equation}
		\label{exQ}
		\begin{aligned}
			Q_k(x)
			%  =& \sum_{i=0}^{k} (-1)^{k-i}\frac{(2k+1)!!}{(2i+1)!!}\cdot\frac{k!}{i!(k-i)!}x^{i}\\
			=&\sum_{i=0}^{k} (-1)^{k-i}\frac{(2k+1)!!}{(2i+1)!!}\cdot\binom{k}{i} x^{i},
		\end{aligned}
	\end{equation}
	where $(-1)!!:=1$, and $\binom{0}{0}:=1$ in the above expressions, and
	\begin{equation}
		\label{R2k}
		R_{2k}(x)=P_k^2(x),
	\end{equation}
	\begin{equation}
		\label{R2k1}
		R_{2k+1}(x)=xQ_k^2(x).
	\end{equation}

	We aim to prove the above four equalities true for all $k\in\mathbb{N}$ by induction.

	First of all, it is easy to verify that the first three terms conform with the above expressions, where
	\[
		\begin{split}
			P_0(x)&=1,\\
			Q_0(x)&=1,\\
			P_1(x)&=x-1,
		\end{split}
	\]
	and
	\[
		\begin{split}
			y^{(0)}&=P_0^2(p)e^{-p},\\
			y^{(1)}&=pQ_0^2(p)e^{-p},\\
			y^{(2)}&=P_1^2(p)e^{-p}.\\
		\end{split}
	\]

	For $n=2k+1$, where $k\geq 1$, suppose that Eq. (\ref{exP}) and Eq. (\ref{R2k}) hold for all $i\leq k$ and Eq. (\ref{exQ}) and Eq. (\ref{R2k1}) hold for all $i\leq k-1$. We need to prove that Eq. (\ref{exQ}) and Eq. (\ref{R2k1}) also hold for $i = k$. By Eq. (\ref{rec}), we only need to prove
	\begin{equation}
		\label{rec1}
		xQ_k^2=(x-2k)P_k^2-2k(x-2k)xQ_{k-1}^2+2k(2k-1)^2P_{k-1}^2.
	\end{equation}
	
	For $n=2k$, where $k\geq 2$, suppose that Eq. (\ref{exP}) and Eq. (\ref{R2k}) hold for all $i\leq k-1$ and Eq. (\ref{exQ}) and Eq. (\ref{R2k1}) hold for all $i\leq k-1$. We need to prove that Eq. (\ref{exP}) and Eq. (\ref{R2k}) also hold for $i = k$. By Eq. (\ref{rec}), we only need to prove
	\begin{equation}
		\label{rec2}
		P_k^2=(x-(2k-1))xQ_{k-1}^2-(2k-1)(x-(2k-1))P_{k-1}^2+(2k-1)(2k-2)^2xQ_{k-2}^2.
	\end{equation}

	For notational simplicity, we set for $i\in[k]$,
	\[
		\begin{split}
			a_i^k=(-1)^{k-i}\frac{(2k-1)!!}{(2i-1)!!}\cdot\binom{k}{i},\quad b_i^k=(-1)^{k-i}\frac{(2k+1)!!}{(2i+1)!!}\cdot\binom{k}{i}.
		\end{split}
	\]
	For $i \in [2k]$,
	\[
		\begin{split}
			A_{i}^{k}=\sum_{j=0\lor i-k}^{i\land k}a_{j}^{k}a_{i-j}^{k},\quad B_{i}^{k}=\sum_{j=0\lor i-k}^{i\land k}b_{j}^{k}b_{i-j}^{k}.
		\end{split}
	\]
	The polynomials are written as
	\[
		\begin{split}
			P_k(x)=\sum_{i=0}^{k} a_i^k x^i,\quad Q_k(x)=\sum_{i=0}^{k} b_i^k x^i.
		\end{split}
	\]
	\[
		(P_{k}(x))^2=\sum_{i=0}^{2k}A_{i}^{k} x^i,\quad (Q_{k}(x))^2=\sum_{i=0}^{2k} B_{i}^{k} x^i.
	\]
	\paragraph{Proof of Eq. (\ref{rec1}).} Now consider the right-hand side of Eq. (\ref{rec1}).
	\[
		\begin{split}
			\textrm{RHS}=&(x-2k)\sum_{i=0}^{2k}A_{i}^{k} x^i-2kx(x-2k)\sum_{i=0}^{2k-2} B_{i}^{k-1} x^i+2k(2k-1)^2\sum_{i=0}^{2k-2}A_{i}^{k-1} x^i\\
			=&\sum_{i=0}^{2k}A_{i}^{k} x^{i+1}+\sum_{i=0}^{2k}(-2k) A_{i}^{k} x^{i}\\
			&+\sum_{i=0}^{2k-2} (-2k)B_{i}^{k-1} x^{i+2}+\sum_{i=0}^{2k-2}(2k)^2 B_{i}^{k-1} x^{i+1}\\
			&+\sum_{i=0}^{2k-2}2k(2k-1)^2A_{i}^{k-1} x^i\\
			=&\sum_{i=1}^{2k+1}A_{i-1}^{k} x^{i}+\sum_{i=0}^{2k}(-2k) A_{i}^{k} x^{i}\\
			&+\sum_{i=2}^{2k} (-2k)B_{i-2}^{k-1} x^{i}+\sum_{i=1}^{2k-1}(2k)^2 B_{i-1}^{k-1} x^{i}\\
			&+\sum_{i=0}^{2k-2}2k(2k-1)^2A_{i}^{k-1} x^i\\
			=&\sum_{i=2}^{2k-2}(A_{i-1}^{k}-2kA_{i}^{k}-2kB_{i-2}^{k-1}+(2k)^2 B_{i-1}^{k-1}+2k(2k-1)^2A_{i}^{k-1})x^i\\
			&+(A_{2k-2}^{k}-2kA_{2k-1}^{k}-2kB_{2k-3}^{k-1}+(2k)^2 B_{2k-2}^{k-1})x^{2k-1}\\
			&+(A_{2k-1}^{k}-2kA_{2k}^{k}-2kB_{2k-2}^{k-1})x^{2k}+A_{2k}^{k}x^{2k+1}\\
			&+(A_{0}^{k}-2kA_{1}^{k}+(2k)^2B_{0}^{k-1}+2k(2k-1)^2A_{1}^{k-1})x\\
			&+(2k(2k-1)^2A_{0}^{k-1}-2kA_{0}^{k}).
		\end{split}
	\]
	For the constant term, the general term formula is 
	\[
		A_0^k=(a_0^k)^2=((2k-1)!!)^2.
	\]
	Hence,
	\[
		2k(2k-1)^2A_{0}^{k-1}-2kA_{0}^{k}=2k[(2k-1)^2\cdot ((2k-3)!!)^2-((2k-1)!!)^2]=0.
	\]
	Plug the result into the right-hand side, we obatin
	\[
		\begin{split}
			\textrm{RHS}=&x\Bigg\{\sum_{i=2}^{2k-2}(A_{i-1}^{k}-2kA_{i}^{k}-2kB_{i-2}^{k-1}+(2k)^2 B_{i-1}^{k-1}+2k(2k-1)^2A_{i}^{k-1})x^{i-1}\\
			&+(A_{2k-2}^{k}-2kA_{2k-1}^{k}-2kB_{2k-3}^{k-1}+(2k)^2 B_{2k-2}^{k-1})x^{2k-2}\\
			&+(A_{2k-1}^{k}-2kA_{2k}^{k}-2kB_{2k-2}^{k-1})x^{2k-1}+A_{2k}^{k}x^{2k}\\
			&+(A_{0}^{k}-2kA_{1}^{k}+(2k)^2B_{0}^{k-1}+2k(2k-1)^2A_{1}^{k-1})\Bigg\}\\
			=&x\Bigg\{\sum_{i=1}^{2k-3}(A_{i}^{k}-2kA_{i+1}^{k}-2kB_{i-1}^{k-1}+(2k)^2 B_{i}^{k-1}+2k(2k-1)^2A_{i+1}^{k-1})x^{i}\\
			&+(A_{2k-2}^{k}-2kA_{2k-1}^{k}-2kB_{2k-3}^{k-1}+(2k)^2 B_{2k-2}^{k-1})x^{2k-2}\\
			&+(A_{2k-1}^{k}-2kA_{2k}^{k}-2kB_{2k-2}^{k-1})x^{2k-1}+A_{2k}^{k}x^{2k}\\
			&+(A_{0}^{k}-2kA_{1}^{k}+(2k)^2B_{0}^{k-1}+2k(2k-1)^2A_{1}^{k-1})\Bigg\}.\\
		\end{split}
	\]
	We then verify the coefficients are equal to those of $xQ_k^2(x)=x\left(\sum_{i=0}^{2k} B_{i}^{k} x^i\right)$.

	For $i=2k$, \[A_{2k}^k=(a_k^k)^2=1^2=(b_k^k)^2=B_{2k}^k.\]

	For $i=2k-1$,
	\[
		\begin{split}
			&A_{2k-1}^{k}-2kA_{2k}^{k}-2kB_{2k-2}^{k-1}\\
			=&2a_{k-1}^ka_{k}^k-2k(a_{k}^k)^2-2k(b_{k-1}^{k-1})^2\\
			=&-2(2k-1)k-2k-2k\\
			=&-2k(2k+1)=B_{2k-1}^k.
		\end{split}
	\]

	For $i=2k-2$,
	\[
		\begin{split}
			&A_{2k-2}^{k}-2kA_{2k-1}^{k}-2kB_{2k-3}^{k-1}+(2k)^2 B_{2k-2}^{k-1}\\
			=&(a_{k-1}^k)^2+2a_{k-2}^ka_{k}^k-2k\cdot2a_{k-1}^ka_{k}^k-2k\cdot2b_{k-2}^{k-1}b_{k-1}^{k-1}+(2k)^2(b_{k-1}^{k-1})^2\\
			=&((2k-1)\cdot k)^2+2\cdot(2k-1)(2k-3)\frac{k(k-1)}{2}+2k\cdot2(2k-1)k\\
			&+2k\cdot 2 (2k-1)(k-1)+(2k)^2\\
			=&((2k+1)k)^2+2(2k+1)(2k-1)\frac{k(k-1)}{2}=B_{2k-2}^k.
		\end{split}
	\]

	For $i=0$, 
	\[
		\begin{split}
			&A_{0}^{k}-2kA_{1}^{k}+(2k)^2B_{0}^{k-1}+2k(2k-1)^2A_{1}^{k-1}\\
			=&((2k-1)!!)^2-2k(-2(2k-1)!!\cdot(2k-1)!!\cdot k)\\
			&+(2k)^2((2k-1)!!)^2+2k(2k-1)^2(-2(2k-3)!!\cdot(2k-3)!!\cdot (k-1))\\
			=&((2k-1)!!)^2+2(2k)^2((2k-1)!!)^2-(2k)(2k-2)((2k-1)!!)^2\\
			=&(2k+1)^2((2k-1)!!)^2=((2k+1)!!)^2=B_{0}^{k}.
		\end{split}
	\]

	For $1\leq i \leq 2k-3$, we need to show that
	\[
		A_{i}^{k}-2kA_{i+1}^{k}-2kB_{i-1}^{k-1}+(2k)^2 B_{i}^{k-1}+2k(2k-1)^2A_{i+1}^{k-1}=B_{i}^k.
	\]
	
	For starters, we have for the right-hand side that
	% \[
	% 	\begin{split}
	% 		(-1)^i(B_{i}^{k}-A_{i}^{k})=&\sum_{j=0\lor i-k}^{i\land k}\frac{(2k+1)!!}{(2j+1)!!}\binom{k}{j}\frac{(2k+1)!!}{(2(i-j)+1)!!}\binom{k}{i-j}
	% 		\\
	% 		&-\sum_{j=0\lor i-k}^{i\land k}\frac{(2k-1)!!}{(2j-1)!!}\binom{k}{j}\frac{(2k-1)!!}{(2(i-j)-1)!!}\binom{k}{i-j}\\
	% 		=&\sum_{j=0\lor i-k}^{i\land k}((2k+1)^2-(2j+1)(2(i-j)+1))\frac{(2k-1)!!}{(2j+1)!!}\binom{k}{j}\frac{(2k-1)!!}{(2(i-j)+1)!!}\binom{k}{i-j}\\
	% 		=&\sum_{j=0\lor i-k}^{i\land k}((2k)^2+2(2k)-2j2(i-j)-2i)\frac{(2k-1)!!}{(2j+1)!!}\binom{k}{j}\frac{(2k-1)!!}{(2(i-j)+1)!!}\binom{k}{i-j}\\
	% 		=&\sum_{j=0\lor i-k}^{i\land k}((2k)^2+2(2k)-(2j+1)2i+(2j)^2)\frac{(2k-1)!!}{(2j+1)!!}\binom{k}{j}\frac{(2k-1)!!}{(2(i-j)+1)!!}\binom{k}{i-j}
	% 	\end{split}	
	% \]

	\[
		\begin{split}
			(-1)^iB_{i}^{k}=&\sum_{j=0\lor i-k}^{i\land k}\frac{(2k+1)!!}{(2j+1)!!}\binom{k}{j}\frac{(2k+1)!!}{(2(i-j)+1)!!}\binom{k}{i-j}
			\\
			=&(2k+1)^2\sum_{j=0\lor i-k}^{i\land k}\frac{(2k-1)!!}{(2j+1)!!}\binom{k}{j}\frac{(2k-1)!!}{(2(i-j)+1)!!}\binom{k}{i-j}.\\
		\end{split}	
	\]

	% For the other parts, we have
	For the left-hand side, we have
	\[
		\begin{split}
			&(-1)^i(A_{i}^{k}-2kA_{i+1}^{k}-2kB_{i-1}^{k-1}+(2k)^2 B_{i}^{k-1}+2k(2k-1)^2A_{i+1}^{k-1})\\
			=&\sum_{j=0\lor i-k}^{i\land k}\frac{(2k-1)!!}{(2j-1)!!}\binom{k}{j}\frac{(2k-1)!!}{(2(i-j)-1)!!}\binom{k}{i-j}\\
			&+2k\sum_{j=0\lor i+1-k}^{i+1\land k}\frac{(2k-1)!!}{(2j-1)!!}\binom{k}{j}\frac{(2k-1)!!}{(2(i+1-j)-1)!!}\binom{k}{i+1-j}\\
			&+2k\sum_{j=0\lor i-k}^{i-1\land k-1}\frac{(2k-1)!!}{(2j+1)!!}\binom{k-1}{j}\frac{(2k-1)!!}{(2(i-1-j)+1)!!}\binom{k-1}{i-1-j}\\
			&+(2k)^2\sum_{j=0\lor i-k+1}^{i\land k-1}\frac{(2k-1)!!}{(2j+1)!!}\binom{k-1}{j}\frac{(2k-1)!!}{(2(i-j)+1)!!}\binom{k-1}{i-j}\\
			&-2k(2k-1)^2\sum_{j=0\lor i-k+2}^{i+1\land k-1}\frac{(2k-3)!!}{(2j-1)!!}\binom{k-1}{j}\frac{(2k-3)!!}{(2(i+1-j)-1)!!}\binom{k-1}{i+1-j}\\
			=&\sum_{j=0\lor i-k}^{i\land k}\frac{(2k-1)!!}{(2j-1)!!}\binom{k}{j}\frac{(2k-1)!!}{(2(i-j)-1)!!}\binom{k}{i-j}\\
			&+2k\sum_{j=0\lor i+1-k}^{i+1\land k}\frac{(2k-1)!!}{(2j-1)!!}\binom{k}{j}\frac{(2k-1)!!}{(2(i-j)+1)!!}\binom{k}{i+1-j}\\
			&+2k\sum_{j=0\lor i-k}^{i-1\land k-1}\frac{(2k-1)!!}{(2j+1)!!}\binom{k-1}{j}\frac{(2k-1)!!}{(2(i-j)-1)!!}\binom{k-1}{i-1-j}\\
			&+(2k)^2\sum_{j=0\lor i-k+1}^{i\land k-1}\frac{(2k-1)!!}{(2j+1)!!}\binom{k-1}{j}\frac{(2k-1)!!}{(2(i-j)+1)!!}\binom{k-1}{i-j}\\
			&-2k\sum_{j=0\lor i-k+2}^{i+1\land k-1}\frac{(2k-1)!!}{(2j-1)!!}\binom{k-1}{j}\frac{(2k-1)!!}{(2(i-j)+1)!!}\binom{k-1}{i+1-j}\\
			=&\sum_{j=0\lor i-k}^{i\land k}[(2j+1)(2(i-j)+1)]\frac{(2k-1)!!}{(2j+1)!!}\binom{k}{j}\frac{(2k-1)!!}{(2(i-j)+1)!!}\binom{k}{i-j}\\
			&+2k\sum_{j=0\lor i+1-k}^{i+1\land k}\frac{(2k-1)!!}{(2j-1)!!}\binom{k}{j}\frac{(2k-1)!!}{(2(i-j)+1)!!}\binom{k}{i+1-j}\\
			&+2k\sum_{j=0\lor i-k}^{i-1\land k-1}\frac{(2k-1)!!}{(2j+1)!!}\binom{k-1}{j}\frac{(2k-1)!!}{(2(i-j)-1)!!}\binom{k-1}{i-1-j}\\
			&+\sum_{j=0\lor i-k+1}^{i\land k-1}[2(k-j)2(k-(i-j))]\frac{(2k-1)!!}{(2j+1)!!}\binom{k}{j}\frac{(2k-1)!!}{(2(i-j)+1)!!}\binom{k}{i-j}\\
			&-2k\sum_{j=0\lor i-k+2}^{i+1\land k-1}\frac{(2k-1)!!}{(2j-1)!!}\binom{k-1}{j}\frac{(2k-1)!!}{(2(i-j)+1)!!}\binom{k-1}{i+1-j}.\\
		\end{split}
	\]

For the second, third and fifth terms, we have
	\[
		\begin{split}
			&2k\sum_{j=0\lor i+1-k}^{i+1\land k}\frac{(2k-1)!!}{(2j-1)!!}\binom{k}{j}\frac{(2k-1)!!}{(2(i-j)+1)!!}\binom{k}{i+1-j}\\
			&+2k\sum_{j=0\lor i-k}^{i-1\land k-1}\frac{(2k-1)!!}{(2j+1)!!}\binom{k-1}{j}\frac{(2k-1)!!}{(2(i-j)-1)!!}\binom{k-1}{i-1-j}\\
			&-2k\sum_{j=0\lor i-k+2}^{i+1\land k-1}\frac{(2k-1)!!}{(2j-1)!!}\binom{k-1}{j}\frac{(2k-1)!!}{(2(i-j)+1)!!}\binom{k-1}{i+1-j}\\
			=&2k\sum_{j=0\lor i+1-k}^{i+1\land k}\frac{(2k-1)!!}{(2j-1)!!}\binom{k}{j}\frac{(2k-1)!!}{(2(i-j)+1)!!}\binom{k}{i+1-j}\\
			&+2k\sum_{j=0\lor i+1-k}^{i\land k-1}\frac{(2k-1)!!}{(2j+1)!!}\binom{k-1}{j}\frac{(2k-1)!!}{(2(i-j)-1)!!}\binom{k}{i-j}\\
			&-2k\sum_{j=0\lor i+1-k}^{i\land k-1}\frac{(2k-1)!!}{(2j+1)!!}\binom{k-1}{j}\frac{(2k-1)!!}{(2(i-j)-1)!!}\binom{k-1}{i-j}\\
			&-2k\sum_{j=0\lor i-k+2}^{i+1\land k-1}\frac{(2k-1)!!}{(2j-1)!!}\binom{k-1}{j}\frac{(2k-1)!!}{(2(i-j)+1)!!}\binom{k-1}{i+1-j}\\
			=&2k\sum_{j=0\lor i+1-k}^{i+1\land k}\frac{(2k-1)!!}{(2j-1)!!}\frac{k!}{j!(k-j)!}\frac{(2k-1)!!}{(2(i-j)+1)!!}\frac{k!}{(i-j+1)!(k-i+j-1)!}\\
			&+2k\sum_{j=0\lor i+1-k}^{i\land k-1}\frac{(2k-1)!!}{(2j+1)!!}\frac{(k-1)!}{j!(k-j-1)!}\frac{(2k-1)!!}{(2(i-j)-1)!!}\frac{k!}{(i-j)!(k-i+j)!}\\
			&-2k\sum_{j=0\lor i+1-k}^{i\land k-1}\frac{(2k-1)!!}{(2j+1)!!}\frac{(k-1)!}{j!(k-j-1)!}\frac{(2k-1)!!}{(2(i-j)-1)!!}\frac{(k-1)!}{(i-j)!(k-i+j-1)!}\\
			&-2k\sum_{j=0\lor i-k+2}^{i+1\land k-1}\frac{(2k-1)!!}{(2j-1)!!}\frac{(k-1)!}{j!(k-j-1)!}\frac{(2k-1)!!}{(2(i-j)+1)!!}\frac{(k-1)!}{(i-j+1)!(k-i+j-2)!}\\
			=&2k\sum_{j=0\lor i+1-k}^{i+1\land k}\frac{(2k-1)!!}{(2j-1)!!}\frac{k!}{j!(k-j)!}\frac{(2k-1)!!}{(2(i-j)+1)!!}\frac{k!}{(i-j+1)!(k-i+j-1)!}\\
			&+\sum_{j=0\lor i+1-k}^{i\land k-1}[2(k-j)(2(i-j)+1)]\frac{(2k-1)!!}{(2j+1)!!}\frac{k!}{j!(k-j)!}\frac{(2k-1)!!}{(2(i-j)+1)!!}\frac{k!}{(i-j)!(k-i+j)!}\\
			&-2k\sum_{j=0\lor i+1-k}^{i\land k-1}\frac{(2k-1)!!}{(2j+1)!!}\frac{(k-1)!}{j!(k-j-1)!}\frac{(2k-1)!!}{(2(i-j)-1)!!}\frac{(k-1)!}{(i-j)!(k-i+j-1)!}\\
			&-2k\sum_{j=0\lor i-k+2}^{i+1\land k-1}\frac{(2k-1)!!}{(2j-1)!!}\frac{(k-1)!}{j!(k-j-1)!}\frac{(2k-1)!!}{(2(i-j)+1)!!}\frac{(k-1)!}{(i-j+1)!(k-i+j-2)!},\\
		\end{split}
	\]
where in the first equality, we use the relation
\[
	\binom{k}{i-j}-\binom{k-1}{i-j}=\binom{k-1}{i-j-1}.
\]

For the first, third and fourth terms of the former expression, we have
\[
	\begin{split}
		&2k\sum_{j=0\lor i+1-k}^{i+1\land k}\frac{(2k-1)!!}{(2j-1)!!}\frac{k!}{j!(k-j)!}\frac{(2k-1)!!}{(2(i-j)+1)!!}\frac{k!}{(i-j+1)!(k-i+j-1)!}\\
		&-2k\sum_{j=0\lor i+1-k}^{i\land k-1}\frac{(2k-1)!!}{(2j+1)!!}\frac{(k-1)!}{j!(k-j-1)!}\frac{(2k-1)!!}{(2(i-j)-1)!!}\frac{(k-1)!}{(i-j)!(k-i+j-1)!}\\
		&-2k\sum_{j=0\lor i-k+2}^{i+1\land k-1}\frac{(2k-1)!!}{(2j-1)!!}\frac{(k-1)!}{j!(k-j-1)!}\frac{(2k-1)!!}{(2(i-j)+1)!!}\frac{(k-1)!}{(i-j+1)!(k-i+j-2)!}\\
		=&2k\sum_{j=0\lor i+1-k}^{i+1\land k}\frac{(2k-1)!!}{(2j-1)!!}\frac{k!}{j!(k-j)!}\frac{(2k-1)!!}{(2(i-j)+1)!!}\frac{k!}{(i-j+1)!(k-i+j-1)!}\\
		&-2k\sum_{j=0\lor i+1-k}^{i\land k-1}\frac{(2k-1)!!}{(2j+1)!!}\frac{(k-1)!}{j!(k-j-1)!}\frac{(2k-1)!!}{(2(i-j)-1)!!}\frac{(k-1)!}{(i-j)!(k-i+j-1)!}\\
		&-2\sum_{j=0\lor i-k+2}^{i+1\land k-1}(k-j)\frac{(2k-1)!!}{(2j-1)!!}\frac{k!}{j!(k-j)!}\frac{(2k-1)!!}{(2(i-j)+1)!!}\frac{(k-1)!}{(i-j+1)!(k-i+j-2)!}\\
		=&2k\sum_{j=0\lor i+1-k}^{i+1\land k}\frac{(2k-1)!!}{(2j-1)!!}\frac{k!}{j!(k-j)!}\frac{(2k-1)!!}{(2(i-j)+1)!!}\frac{k!}{(i-j+1)!(k-i+j-1)!}\\
		&-2k\sum_{j=0\lor i+1-k}^{i\land k-1}\frac{(2k-1)!!}{(2j+1)!!}\frac{(k-1)!}{j!(k-j-1)!}\frac{(2k-1)!!}{(2(i-j)-1)!!}\frac{(k-1)!}{(i-j)!(k-i+j-1)!}\\
		&-2k\sum_{j=0\lor i-k+2}^{i+1\land k-1}\frac{(2k-1)!!}{(2j-1)!!}\frac{k!}{j!(k-j)!}\frac{(2k-1)!!}{(2(i-j)+1)!!}\frac{(k-1)!}{(i-j+1)!(k-i+j-2)!}\\
		&+\sum_{j=0\lor i-k+2}^{i+1\land k-1}(2j)\frac{(2k-1)!!}{(2j-1)!!}\frac{k!}{j!(k-j)!}\frac{(2k-1)!!}{(2(i-j)+1)!!}\frac{(k-1)!}{(i-j+1)!(k-i+j-2)!}\\
		=&2k\sum_{j=0\lor i-k}^{i\land k}\frac{(2k-1)!!}{(2j-1)!!}\frac{k!}{j!(k-j)!}\frac{(2k-1)!!}{(2(i-j)+1)!!}\frac{(k-1)!}{(i-j)!(k-i+j-1)!}\\
		&-2k\sum_{j=0\lor i+1-k}^{i\land k-1}\frac{(2k-1)!!}{(2j+1)!!}\frac{(k-1)!}{j!(k-j-1)!}\frac{(2k-1)!!}{(2(i-j)-1)!!}\frac{(k-1)!}{(i-j)!(k-i+j-1)!}\\
		&+\sum_{j=0\lor i-k+2}^{i+1\land k-1}(2j)\frac{(2k-1)!!}{(2j-1)!!}\frac{k!}{j!(k-j)!}\frac{(2k-1)!!}{(2(i-j)+1)!!}\frac{(k-1)!}{(i-j+1)!(k-i+j-2)!}\\
		=&\sum_{j=0\lor i-k}^{i\land k}[(2j+1)2(k-i+j)]\frac{(2k-1)!!}{(2j+1)!!}\frac{k!}{j!(k-j)!}\frac{(2k-1)!!}{(2(i-j)+1)!!}\frac{(k-1)!}{(i-j)!(k-i+j-1)!}\\
		&-2k\sum_{j=0\lor i+1-k}^{i\land k-1}\frac{(2k-1)!!}{(2j+1)!!}\frac{(k-1)!}{j!(k-j-1)!}\frac{(2k-1)!!}{(2(i-j)-1)!!}\frac{(k-1)!}{(i-j)!(k-i+j-1)!}\\
		&+\sum_{j=0\lor i-k+2}^{i+1\land k-1}(2j)\frac{(2k-1)!!}{(2j-1)!!}\frac{k!}{j!(k-j)!}\frac{(2k-1)!!}{(2(i-j)+1)!!}\frac{(k-1)!}{(i-j+1)!(k-i+j-2)!},\\
	\end{split}
\]
where in the third equality, we combine the first and third terms using the relation
\[
	\binom{k}{i-j+1}-\binom{k-1}{i-j+1}=\binom{k-1}{i-j}.
\]

For the last two terms, we have
\[
	\begin{split}
		&-2k\sum_{j=0\lor i+1-k}^{i\land k-1}\frac{(2k-1)!!}{(2j+1)!!}\frac{(k-1)!}{j!(k-j-1)!}\frac{(2k-1)!!}{(2(i-j)-1)!!}\frac{(k-1)!}{(i-j)!(k-i+j-1)!}\\
		&+\sum_{j=0\lor i-k+2}^{i+1\land k-1}(2j)\frac{(2k-1)!!}{(2j-1)!!}\frac{k!}{j!(k-j)!}\frac{(2k-1)!!}{(2(i-j)+1)!!}\frac{(k-1)!}{(i-j+1)!(k-i+j-2)!}\\
		=&-2k\sum_{j=0\lor i+1-k}^{i\land k-1}\frac{(2k-1)!!}{(2j+1)!!}\frac{(k-1)!}{j!(k-j-1)!}\frac{(2k-1)!!}{(2(i-j)-1)!!}\frac{(k-1)!}{(i-j)!(k-i+j-1)!}\\
		&+\sum_{j=0\lor i-k+1}^{i\land k-1}(2j+2)\frac{(2k-1)!!}{(2j+1)!!}\frac{k!}{(j+1)!(k-j-1)!}\frac{(2k-1)!!}{(2(i-j)-1)!!}\frac{(k-1)!}{(i-j)!(k-i+j-1)!}\\
		=&-2\sum_{j=0\lor i+1-k}^{i\land k-1}\frac{(2k-1)!!}{(2j+1)!!}\frac{k!}{j!(k-j-1)!}\frac{(2k-1)!!}{(2(i-j)-1)!!}\frac{(k-1)!}{(i-j)!(k-i+j-1)!}\\
		&+2\sum_{j=0\lor i-k+1}^{i\land k-1}\frac{(2k-1)!!}{(2j+1)!!}\frac{k!}{j!(k-j-1)!}\frac{(2k-1)!!}{(2(i-j)-1)!!}\frac{(k-1)!}{(i-j)!(k-i+j-1)!}\\
		=&0.
	\end{split}
\]

Combine the four parts illustrated above, we have that
\[
	\begin{split}
		&(-1)^i(A_{i}^{k}-2kA_{i+1}^{k}-2kB_{i-1}^{k-1}+(2k)^2 B_{i}^{k-1}+2k(2k-1)^2A_{i+1}^{k-1})\\
		=&\sum_{j=0\lor i-k}^{i\land k}[(2j+1)(2(i-j)+1)+2(k-j)2(k-i+j)\\
		&\quad\quad\quad+2(k-j)(2(i-j)+1)+(2j+1)2(k-i+j)+0]\\
		&\quad\quad\quad\cdot\frac{(2k-1)!!}{(2j+1)!!}\binom{k}{j}\frac{(2k-1)!!}{(2(i-j)+1)!!}\binom{k}{i-j}\\
		=&\sum_{j=0\lor i-k}^{i\land k}(2k+1)^2\frac{(2k-1)!!}{(2j+1)!!}\binom{k}{j}\frac{(2k-1)!!}{(2(i-j)+1)!!}\binom{k}{i-j}\\
		=&\sum_{j=0\lor i-k}^{i\land k}\frac{(2k+1)!!}{(2j+1)!!}\binom{k}{j}\frac{(2k+1)!!}{(2(i-j)+1)!!}\binom{k}{i-j}\\
		=&(-1)^{i}B_{i}^{k}.\\
	\end{split}
\]

Finally, we complete the proof of Eq. (\ref{rec1}).

\paragraph{Proof of Eq. (\ref{rec2}).} Consider the right-hand side of Eq. (\ref{rec2}).

\[
	\begin{split}
		\textrm{RHS}=&(x-(2k-1))x\sum_{i=0}^{2k-2}B_{i}^{k-1}x^{i}-(2k-1)(x-(2k-1))\sum_{i=0}^{2k-2}A_{i}^{k-1}x^{i}+(2k-1)(2k-2)^2x\sum_{i=0}^{2k-4}B_{i}^{k-2}x^{i}\\
		=&\sum_{i=0}^{2k-2}B_{i}^{k-1}x^{i+2}+\sum_{i=0}^{2k-2}(-(2k-1))B_{i}^{k-1}x^{i+1}\\
		&+\sum_{i=0}^{2k-2}(-(2k-1))A_{i}^{k-1}x^{i+1}+\sum_{i=0}^{2k-2}(2k-1)^2A_{i}^{k-1}x^{i}\\
		&+\sum_{i=0}^{2k-4}(2k-1)(2k-2)^2B_{i}^{k-2}x^{i+1}\\
		=&\sum_{i=2}^{2k}B_{i-2}^{k-1}x^{i}+\sum_{i=1}^{2k-1}(-(2k-1))B_{i-1}^{k-1}x^{i}\\
		&+\sum_{i=1}^{2k-1}(-(2k-1))A_{i-1}^{k-1}x^{i}+\sum_{i=0}^{2k-2}(2k-1)^2A_{i}^{k-1}x^{i}\\
		&+\sum_{i=1}^{2k-3}(2k-1)(2k-2)^2B_{i-1}^{k-2}x^{i}\\
		=&\sum_{i=2}^{2k-3}(B_{i-2}^{k-1}-(2k-1)B_{i-1}^{k-1}-(2k-1)A_{i-1}^{k-1}+(2k-1)^2A_{i}^{k-1}+(2k-1)(2k-2)^2B_{i-1}^{k-2})x^{i}\\
		&+(B_{2k-4}^{k-1}-(2k-1)B_{2k-3}^{k-1}-(2k-1)A_{2k-3}^{k-1}+(2k-1)^2A_{2k-2}^{k-1})x^{2k-2}\\
		&+(B_{2k-3}^{k-1}-(2k-1)B_{2k-2}^{k-1}-(2k-1)A_{2k-2}^{k-1})x^{2k-1}+B_{2k-2}^{k-1}x^{2k}\\
		&+(-(2k-1)B_{0}^{k-1}-(2k-1)A_{0}^{k-1}+(2k-1)^2A_{1}^{k-1}+(2k-1)(2k-2)^2B_{0}^{k-2})x\\
		&+(2k-1)^2A_{0}^{k-1}.\\
	\end{split}
\]

It suffices to prove that the coefficients of the above expression are equal to those of $P_k^2(x)=\sum_{i=0}^{2k} A_{i}^{k} x^i$.

For $i=0,1,2k-2,2k-1,2k$, the verifications are trivial. We only need to show that for $2\leq i\leq 2k-3$, it holds that
\[
	B_{i-2}^{k-1}-(2k-1)B_{i-1}^{k-1}-(2k-1)A_{i-1}^{k-1}+(2k-1)^2A_{i}^{k-1}+(2k-1)(2k-2)^2B_{i-1}^{k-2}=A_{i}^{k}.
\]

Consider the right-hand side, we have
\[
	(-1)^{i}A_{i}^{k}=\sum_{j=0\lor i-k}^{i\land k}\frac{(2k-1)!!}{(2j-1)!!}\binom{k}{j}\frac{(2k-1)!!}{(2(i-j)-1)!!}\binom{k}{i-j}.
\]

\newpage
For the left-hand side, we first have
\[
	\begin{split}
		&(-1)^{i}(B_{i-2}^{k-1}-(2k-1)B_{i-1}^{k-1}-(2k-1)A_{i-1}^{k-1}+(2k-1)^2A_{i}^{k-1}+(2k-1)(2k-2)^2B_{i-1}^{k-2})\\
		=&\sum_{j=0\lor i-1-k}^{i-2\land k-1}\frac{(2k-1)!!}{(2j+1)!!}\binom{k-1}{j}\frac{(2k-1)!!}{(2(i-2-j)+1)!!}\binom{k-1}{i-2-j}\\
		&+(2k-1)\sum_{j=0\lor i-k}^{i-1\land k-1}\frac{(2k-1)!!}{(2j+1)!!}\binom{k-1}{j}\frac{(2k-1)!!}{(2(i-1-j)+1)!!}\binom{k-1}{i-1-j}\\
		&+(2k-1)\sum_{j=0\lor i-k}^{i-1\land k-1}\frac{(2k-3)!!}{(2j-1)!!}\binom{k-1}{j}\frac{(2k-3)!!}{(2(i-1-j)-1)!!}\binom{k-1}{i-1-j}\\
		&+(2k-1)^2\sum_{j=0\lor i-k+1}^{i\land k-1}\frac{(2k-3)!!}{(2j-1)!!}\binom{k-1}{j}\frac{(2k-3)!!}{(2(i-j)-1)!!}\binom{k-1}{i-j}\\
		&-(2k-1)(2k-2)^2\sum_{j=0\lor i-k+1}^{i-1\land k-2}\frac{(2k-3)!!}{(2j+1)!!}\binom{k-2}{j}\frac{(2k-3)!!}{(2(i-1-j)+1)!!}\binom{k-2}{i-1-j}\\
		=&\sum_{j=0\lor i-1-k}^{i-2\land k-1}\frac{(2k-1)!!}{(2j+1)!!}\binom{k-1}{j}\frac{(2k-1)!!}{(2(i-j)-3)!!}\binom{k-1}{i-2-j}\\
		&+(2k-1)\sum_{j=0\lor i-k}^{i-1\land k-1}\frac{(2k-1)!!}{(2j+1)!!}\binom{k-1}{j}\frac{(2k-1)!!}{(2(i-j)-1)!!}\binom{k-1}{i-1-j}\\
		&+(2k-1)\sum_{j=0\lor i-k}^{i-1\land k-1}\frac{(2k-3)!!}{(2j-1)!!}\binom{k-1}{j}\frac{(2k-3)!!}{(2(i-j)-3)!!}\binom{k-1}{i-1-j}\\
		&+(2k-1)^2\sum_{j=0\lor i-k+1}^{i\land k-1}\frac{(2k-3)!!}{(2j-1)!!}\binom{k-1}{j}\frac{(2k-3)!!}{(2(i-j)-1)!!}\binom{k-1}{i-j}\\
		&-(2k-1)(2k-2)^2\sum_{j=0\lor i-k+1}^{i-1\land k-2}\frac{(2k-3)!!}{(2j+1)!!}\binom{k-2}{j}\frac{(2k-3)!!}{(2(i-j)-1)!!}\binom{k-2}{i-1-j}\\
		=&\sum_{j=0\lor i-k}^{i-1\land k-1}\frac{(2k-1)!!}{(2j+1)!!}\binom{k-1}{j}\frac{(2k-1)!!}{(2(i-j)-3)!!}\binom{k}{i-1-j}\\
		&-\sum_{j=0\lor i-k}^{i-1\land k-1}\frac{(2k-1)!!}{(2j+1)!!}\binom{k-1}{j}\frac{(2k-1)!!}{(2(i-j)-3)!!}\binom{k-1}{i-1-j}\\
		&+(2k-1)\sum_{j=0\lor i-k}^{i-1\land k-1}\frac{(2k-1)!!}{(2j+1)!!}\binom{k-1}{j}\frac{(2k-1)!!}{(2(i-j)-1)!!}\binom{k-1}{i-1-j}\\
		&+(2k-1)\sum_{j=0\lor i-k}^{i-1\land k-1}\frac{(2k-3)!!}{(2j-1)!!}\binom{k-1}{j}\frac{(2k-3)!!}{(2(i-j)-3)!!}\binom{k-1}{i-1-j}\\
		&+\sum_{j=0\lor i-k+1}^{i\land k-1}\frac{(2k-1)!!}{(2j-1)!!}\binom{k-1}{j}\frac{(2k-1)!!}{(2(i-j)-1)!!}\binom{k-1}{i-j}\\
		&-(2k-1)\sum_{j=0\lor i-k+1}^{i-1\land k-2}2(k-1-j)2(k-i+j)\frac{(2k-3)!!}{(2j+1)!!}\binom{k-1}{j}\frac{(2k-3)!!}{(2(i-j)-1)!!}\binom{k-1}{i-1-j}.\\
	\end{split}
\]

To continue, we have
\[
	\begin{split}
		\textrm{LHS}=&\sum_{j=(0\lor i-k)+1}^{i\land k}\frac{(2k-1)!!}{(2j-1)!!}\binom{k-1}{j-1}\frac{(2k-1)!!}{(2(i-j)-1)!!}\binom{k}{i-j}\\
		&+\sum_{j=0\lor i-k+1}^{i\land k-1}\frac{(2k-1)!!}{(2j-1)!!}\binom{k-1}{j}\frac{(2k-1)!!}{(2(i-j)-1)!!}\binom{k-1}{i-j}\\
		&-\sum_{j=0\lor i-k}^{i-1\land k-1}\frac{(2k-1)!!}{(2j+1)!!}\binom{k-1}{j}\frac{(2k-1)!!}{(2(i-j)-3)!!}\binom{k-1}{i-1-j}\\
		&+(2k-1)\sum_{j=0\lor i-k}^{i-1\land k-1}\frac{(2k-1)!!}{(2j+1)!!}\binom{k-1}{j}\frac{(2k-1)!!}{(2(i-j)-1)!!}\binom{k-1}{i-1-j}\\
		&+(2k-1)\sum_{j=0\lor i-k}^{i-1\land k-1}\frac{(2k-3)!!}{(2j-1)!!}\binom{k-1}{j}\frac{(2k-3)!!}{(2(i-j)-3)!!}\binom{k-1}{i-1-j}\\
		&-(2k-1)\sum_{j=0\lor i-k+1}^{i-1\land k-2}2(k-1-j)2(k-i+j)\frac{(2k-3)!!}{(2j+1)!!}\binom{k-1}{j}\frac{(2k-3)!!}{(2(i-j)-1)!!}\binom{k-1}{i-1-j}\\
		=&\sum_{j=(0\lor i-k)+1}^{i\land k}\frac{(2k-1)!!}{(2j-1)!!}\binom{k-1}{j-1}\frac{(2k-1)!!}{(2(i-j)-1)!!}\binom{k}{i-j}\\
		&+\sum_{j=0\lor i-k+1}^{i\land k-1}\frac{(2k-1)!!}{(2j-1)!!}\binom{k-1}{j}\frac{(2k-1)!!}{(2(i-j)-1)!!}\binom{k-1}{i-j}\\
		&-\sum_{j=0\lor i-k}^{i-1\land k-1}(2k-1)(2(i-j)-1)\frac{(2k-3)!!}{(2j+1)!!}\binom{k-1}{j}\frac{(2k-1)!!}{(2(i-j)-1)!!}\binom{k-1}{i-1-j}\\
		&+\sum_{j=0\lor i-k}^{i-1\land k-1}(2k-1)^2\frac{(2k-3)!!}{(2j+1)!!}\binom{k-1}{j}\frac{(2k-1)!!}{(2(i-j)-1)!!}\binom{k-1}{i-1-j}\\
		&+\sum_{j=0\lor i-k}^{i-1\land k-1}(2j+1)(2(i-j)-1)\frac{(2k-3)!!}{(2j+1)!!}\binom{k-1}{j}\frac{(2k-1)!!}{(2(i-j)-1)!!}\binom{k-1}{i-1-j}\\
		&-\sum_{j=0\lor i-k+1}^{i-1\land k-2}(2k-2(j+1))(2k-2(i-j))\frac{(2k-3)!!}{(2j+1)!!}\binom{k-1}{j}\frac{(2k-1)!!}{(2(i-j)-1)!!}\binom{k-1}{i-1-j}.\\
	\end{split}
\]

We consider combining the last four terms. Because
\[
	\begin{split}
		&-(2k-1)(2(i-j)-1)+(2k-1)^2\\
		&+(2j+1)(2(i-j)-1)-(2k-2(j+1))(2k-2(i-j))\\
		=&-(2k-2j-2)(2(i-j)-1)+(2k-1)^2\\
		&-(2k-2j-2)(2k-2(i-j))\\
		=&(2k-1)(2k-1-2k+2j+2)\\
		=&(2k-1)(2j+1),
	\end{split}
\]
we have
\[
	\begin{split}
		\textrm{LHS}=&\sum_{j=(0\lor i-k)+1}^{i\land k}\frac{(2k-1)!!}{(2j-1)!!}\binom{k-1}{j-1}\frac{(2k-1)!!}{(2(i-j)-1)!!}\binom{k}{i-j}\\
		&+\sum_{j=0\lor i-k+1}^{i\land k-1}\frac{(2k-1)!!}{(2j-1)!!}\binom{k-1}{j}\frac{(2k-1)!!}{(2(i-j)-1)!!}\binom{k-1}{i-j}\\
		&+\sum_{j=0\lor i-k}^{i-1\land k-1}(2k-1)(2j+1)\frac{(2k-3)!!}{(2j+1)!!}\binom{k-1}{j}\frac{(2k-1)!!}{(2(i-j)-1)!!}\binom{k-1}{i-1-j}\\
		=&\sum_{j=(0\lor i-k)+1}^{i\land k}\frac{(2k-1)!!}{(2j-1)!!}\binom{k-1}{j-1}\frac{(2k-1)!!}{(2(i-j)-1)!!}\binom{k}{i-j}\\
		&+\sum_{j=0\lor i-k+1}^{i\land k-1}\frac{(2k-1)!!}{(2j-1)!!}\binom{k-1}{j}\frac{(2k-1)!!}{(2(i-j)-1)!!}\binom{k-1}{i-j}\\
		&+\sum_{j=0\lor i-k}^{i-1\land k-1}\frac{(2k-1)!!}{(2j-1)!!}\binom{k-1}{j}\frac{(2k-1)!!}{(2(i-j)-1)!!}\binom{k-1}{i-1-j}\\
		=&\sum_{j=0\lor i-k+1}^{i\land k-1}\frac{(2k-1)!!}{(2j-1)!!}\binom{k}{j}\frac{(2k-1)!!}{(2(i-j)-1)!!}\binom{k}{i-j}\\
		=&(-1)^{i}A_{i}^{k}.
	\end{split}
\]

Finally, we complete the proof of Eq. (\ref{rec2}).

\end{proof}

\subsection{Proof of Corollary \ref{pro33}}
\label{proofpro33}
\begin{proof}
	The Taylor expansion of $K(r)$ is
	\begin{equation}
		\begin{aligned}
			K(r)=&e^{-p}\frac{h^2}{1+h^2}\sum_{n=0}^{\infty}\frac{R_n(p)}{n!(1+h^2)^n}r^n\\
			=&\sum_{n=0}^{\infty}\frac{e^{-p}R_n(p)}{n!}\cdot\frac{h^2}{(1+h^2)^{n+1}}\langle x, x^\prime\rangle^n\\
			=&\sum_{n=0}^{\infty}\frac{e^{-p}R_n(p)}{n!}\cdot\frac{h^2}{(1+h^2)^{n+1}}\left\langle x^{\otimes n}, {x^\prime}^{\otimes n}\right\rangle\\
			=&\sum_{n=0}^{\infty}\left\langle \frac{he^{-{p\over 2}}R_n^{1\over 2}(p)}{\sqrt{n!(1+h^2)^{n+1}}}x^{\otimes n}, \frac{he^{-{p\over 2}}R_n^{1\over 2}(p)}{\sqrt{n!(1+h^2)^{n+1}}}{x^\prime}^{\otimes n}\right\rangle.
		\end{aligned}
	\end{equation}
	
	Hence the feature mapping with respect to kernel 
	(\ref{kernel}) is 
	\[
		\phi(x)=\left(\frac{he^{-{p\over 2}}R_n^{1\over 2}(p)}{\sqrt{n!(1+h^2)^{n+1}}}x^{\otimes n}\right)_{n=0}^{\infty}.
	\]

	For any target function
	\[
		f(x)=\sum_{n=0}^{\infty} \langle F_n, x^{\otimes n}\rangle,
	\]
	where \(F_n\in\mathbb{R}^{d^n}\), we have
	\[	
		\begin{split}
			f(x)=&\sum_{n=0}^{\infty} \langle F_n, x^{\otimes n}\rangle\\
			=&\left\langle \frac{\sqrt{n!(1+h^2)^{n+1}}}{he^{-{p\over 2}}R_n^{1\over 2}(p)}F_n,(\phi(x))_n\right\rangle.
		\end{split}
	\]
	Hence, we have
	\[
		\begin{split}
			\|f\|_{\mathcal{H}_K}^2\leq& \left\|\left(\frac{\sqrt{n!(1+h^2)^{n+1}}}{he^{-{p\over 2}}R_n^{1\over 2}(p)}F_n\right)_{n=0}^{\infty}\right\|_{\mathcal{H}}^2\\
			=&\sum_{n=0}^{\infty}\frac{n!(1+h^2)^{n+1}}{h^2e^{-p}R_n(p)}\|F_n\|_{\mathrm{Fr}}^2\\
			=&\frac{e^p}{h^2}\sum_{n=0}^{\infty}\frac{n!(1+h^2)^{n+1}}{R_n(p)}\|F_n\|_{\mathrm{Fr}}^2,
		\end{split}
	\]
	where \(\|\cdot\|_{\mathrm{Fr}}\) the the Frobenius norm.

	Let \[D(f):=\frac{e^p}{h^2}\sum_{n=0}^{\infty}\frac{n!(1+h^2)^{n+1}}{R_n(p)}\|F_n\|_{\mathrm{Fr}}^2.\]

	By Proposition (\ref{equalnorm}), we conclude that
	\[
		\|f\|_{\mathcal{F}}=\|f\|_{\mathcal{H}_K}\leq \sqrt{D(f)}.
	\]
	
	Furthermore, there exist $v:\mathcal{W}\rightarrow \mathbb{R}$ such that 
	\[
		f(x)=\mathbb{E}_{w\sim \mathcal{N}(0,I_d)}\left[B(w^\top x)v(w)\right],
	\]
	and
	\[
		\|v\|_{\mathcal{H}_\mathcal{W}}\leq \sqrt{D(f)}.
	\]

\end{proof}

\subsection{Proof of Theorem \ref{pro34}}
\label{proofpro34}

Recall that in (\ref{tgfc}), we assume a mild condition that $v$ is $L_v$-Lipschitz continuous. Because \[|v(w)|^2\leq (|v(\boldsymbol{0})|+L_v\|w\|)^2\leq 2v(\boldsymbol{0})^2+2L_v^2\|w\|^2.\] By setting $R=\sqrt{2L_v^2d+2|v(\boldsymbol{0})|^2}$, we have \[\mathbb{E}_{w\sim \mathcal{N}(0,I_d)}\left[v(w)^2\right]\leq R^2.\]

\begin{proof}
    Let $W=(w_1,w_2,...,w_M)$ and $v_m=v(w_m)$. We already have $\varphi(x):=\mathbb{E}_{w\sim \mathcal{N}(0,I_d)}\left[B(w^\top x)v(w)\right]$.

	To obtain the desired result, we consider the concentration property of the random variable 
	\[
		\mathbb{E}_{x}\left|\hat{\varphi}(x) - \varphi(x)\right| = \mathbb{E}_{x}\left|\frac{1}{M}\sum_{m=1}^{M}B(w_m^\top x)v(w_m)-\varphi(x)\right|,
	\]
	in which the randomness comes from $W$.

	Naturally, we consider
	\begin{equation}
		\label{expMGF}
		\begin{split}
			&\,\mathbb{E}_{W} \exp\left(\lambda^2\left(\mathbb{E}_{x}\left|\frac{1}{M}\sum_{m=1}^{M}B(w_m^\top x)v(w_m)-\varphi(x)\right|\right)^2\right)\\
			\leq&\,\mathbb{E}_{W} \exp\left(\lambda^2\mathbb{E}_{x}\left(\frac{1}{M}\sum_{m=1}^{M}B(w_m^\top x)v(w_m)-\varphi(x)\right)^2\right)\\
			\leq&\,\mathbb{E}_{W} \mathbb{E}_{x}\exp\left(\lambda^2\left(\frac{1}{M}\sum_{m=1}^{M}B(w_m^\top x)v(w_m)-\varphi(x)\right)^2\right)\\
			=&\,\mathbb{E}_{x} \mathbb{E}_{W}\exp\left(\lambda^2\left(\frac{1}{M}\sum_{m=1}^{M}B(w_m^\top x)v(w_m)-\varphi(x)\right)^2\right),\\
		\end{split}
	\end{equation}
	where we used Jensen's inequality twice.

	Next, we prove that $B(w_m^\top x)v(w_m)-\varphi(x)$ are sub-gaussian random variables for every $w_m\sim\mathcal{N}(0,I_d)$ and every $x\in\mathbb{R}$. In addition, they have a uniform sub-gaussian norm.
	
	To start with, for every $x\in\mathbb{R}$, we have the following estimation.
	\[	
		\begin{split}
			&\,(B(w_m^\top x)v(w_m)-\varphi(x))^2\\
			\leq &\, 2 B(w_m^\top x)^2v(w_m)^2+2\varphi(x)^2\\
			\leq &\, 2 B(w_m^\top x)^2(L_v\|w_m-\boldsymbol{0}\|_2+|v(\boldsymbol{0})|)^2+2\left(\mathbb{E}_{w\sim \mathcal{N}(0,I_d)}\left[B(w^\top x)v(w)\right]\right)^2\\
			\leq &\, 2 B(w_m^\top x)^2(2L_v^2\|w_m-\boldsymbol{0}\|_2^2+2|v(\boldsymbol{0})|^2)+2\mathbb{E}_{w\sim \mathcal{N}(0,I_d)}\left[B(w^\top x)^2\right]\mathbb{E}_{w\sim \mathcal{N}(0,I_d)}\left[v(w)^2\right]\\
			\leq &\, 4L_v^2\|w_m\|_2^2+4|v(\boldsymbol{0})|^2+2R^2,\\
		\end{split}
	\]
	where we used the fact that $v$ is $L_v$-Lipschitz and $0\leq B(w^\top x)\leq 1$.

	Therefore, we have
	\begin{equation}
		\label{MGFX1}
		\begin{split}
			&\,\mathbb{E}_{W}\exp\left(\lambda^2(B(w_m^\top x)v(w_m)-\varphi(x))^2\right)\\
			\leq &\, \mathbb{E}_{W}\exp\left(\lambda^2(4L_v^2\|w_m\|_2^2+4|v(\boldsymbol{0})|^2+2R^2)\right)\\
			=&\,\exp\left(\lambda^2(4|v(\boldsymbol{0})|^2+2R^2)\right)\cdot\mathbb{E}_{W}\exp\left(4L_v^2\lambda^2\|w_m\|_2^2\right)\\
			=&\,\exp\left(\lambda^2(4|v(\boldsymbol{0})|^2+2R^2)\right)\cdot\prod_{i=1}^{d}\mathbb{E}_{w_{m,d}\sim \mathcal{N}(0,1)}\exp\left(4L_v^2\lambda^2w_{m,d}^2\right)\\
			=&\,\exp\left(\lambda^2(4|v(\boldsymbol{0})|^2+2R^2)\right)\cdot\prod_{i=1}^{d}\frac{1}{\sqrt{1-8L_v^2\lambda^2}}
		\end{split}
	\end{equation}
	By applying $\frac{1}{1-x}\leq e^{2x}$ over $x\in[0,1/2]$, we have that for $\lambda^2\leq {1\over 16L_v^2}$,
	\begin{equation}
		\label{MGFX2}
		\begin{split}
			&\,\exp\left(\lambda^2(4|v(\boldsymbol{0})|^2+2R^2)\right)\cdot\prod_{i=1}^{d}\frac{1}{\sqrt{1-8L_v^2\lambda^2}}\\
			\leq& \,\exp\left(\lambda^2(4|v(\boldsymbol{0})|^2+2R^2)\right)\exp\left(8dL_v^2\lambda^2\right)\\
			=& \,\exp\left(\lambda^2(8dL_v^2+4|v(\boldsymbol{0})|^2+2R^2)\right)\\
			\leq& \,\exp\left(\lambda^2(16dL_v^2+4|v(\boldsymbol{0})|^2+2R^2)\right)\\
			\leq& \,\exp\left(\lambda^2\cdot10R^2\right).\\
		\end{split}
	\end{equation}
	To summarize, let $Y_m=B(w_m^\top x)v(w_m)-\varphi(x)$, then for $\lambda^2\leq {1/ (10R^2)}$, it holds that
	\[
		\mathbb{E}_{W}\exp\left(\lambda^2Y_m^2\right)\leq\exp\left(\lambda^2\cdot10R^2\right).
	\]

	By Lemma \ref{subg2}, we have that for all $\lambda\in\mathbb{R}$,
	\begin{equation}
		\label{MGFX3}
		\mathbb{E}_{W}\exp\left(\lambda Y_m\right)\leq\exp\left(\lambda^2\cdot10R^2\right).
	\end{equation}
	Note that $Y_1,Y_2,...,Y_M$	are independent. Therefore, we have
	\begin{equation}
		\label{MGFX4}
		\begin{split}
			&\,\mathbb{E}_{W}\exp\left(\lambda\left(\frac{1}{M}\sum_{m=1}^{M}B(w_m^\top x)v(w_m)-\varphi(x)\right)\right)\\
			=&\,\mathbb{E}_{W}\exp\left(\frac{\lambda}{M}\sum_{m=1}^{M}Y_m\right)
			=\prod_{m=1}^{M}\mathbb{E}_{w_m}\exp\left(\frac{\lambda}{M}Y_m\right)\\
			\leq &\,\exp\left(\lambda^2\cdot10R^2/M\right).
		\end{split}
	\end{equation}
	By Lemma \ref{subg2} again, we have that for $\lambda^2\leq M/(160R^2)$,
	\begin{equation}
		\label{MGFX5}
		\begin{split}
			&\,\mathbb{E}_{W}\exp\left(\lambda^2\left(\frac{1}{M}\sum_{m=1}^{M}B(w_m^\top x)v(w_m)-\varphi(x)\right)^2\right)\\
			\leq &\,\exp\left(160R^2\lambda^2/M\right).
		\end{split}
	\end{equation}
	Taking expectation over $x$ on both sides and plugging it back to (\ref{expMGF}), we have that 
	\[
	\mathbb{E}_{W} \exp\left(\lambda^2\left(\mathbb{E}_{x}\left|\frac{1}{M}\sum_{m=1}^{M}B(w_m^\top x)v(w_m)-\varphi(x)\right|\right)^2\right)\leq\exp\left(160R^2\lambda^2/M\right).
	\]
	Because $\sqrt{2}\sqrt{160R^2}\leq 18R$, by Lemma \ref{subg}, we conclude that 
	\[
		\left\|\mathbb{E}_{x}\left|\frac{1}{M}\sum_{m=1}^{M}B(w_m^\top x)v(w_m)-\varphi(x)\right|\right\|_{\psi_2} \leq \frac{18R}{\sqrt{M}}.
	\]

	Consequently, applying Lemma \ref{subg0}, for $\delta>0$, by taking some $\epsilon=\frac{18R\sqrt{\log{(4/\delta)}}}{\sqrt{M}}$, we have that
	\[
		\begin{split}
			&\,P\left(\mathbb{E}_{x}\left|\hat{\varphi}(x)-\varphi(x)\right|\geq \epsilon\right)\\
			=&\,P\left(\mathbb{E}_{x}\left|\frac{1}{M}\sum_{m=1}^{M}B(w_m^\top x)v(w_m)-\varphi(x)\right|\geq \epsilon\right)\\
			\leq &\, 2\exp\left(-\frac{M\epsilon^2}{(18R)^2}\right)\leq \delta/2.
		\end{split}
	\]
	Hence, with probability of at least $1-\delta/2$, it holds that 
	\[
		\mathbb{E}_{x}\left|\hat{\varphi}(x)-\varphi(x)\right|\leq \frac{18R\sqrt{\log{(4/\delta)}}}{\sqrt{M}}.
	\]

	In the remaining part of the proof, we consider the high probability bound of $\sum_{m=1}^{M}v_m^2$. To start with, we show that $v(w)$ is a sub-gaussian random variable in which $w\sim \mathcal{N}(0,I_d)$.
	\[
		\begin{split}
			\mathbb{E}\exp\left(\lambda^2 v(w)^2\right)\leq&\,\mathbb{E}\exp\left(\lambda^2 (2L_v^2\|w_m\|_2^2+2|v(\boldsymbol{0})|^2)\right)\\
			\leq &\,\exp\left((4L_v^2d+2|v(\boldsymbol{0})|^2)\lambda^2 \right),
		\end{split}	
	\]
	for $\lambda$ such that $(4L_v^2d+2|v(\boldsymbol{0})|^2)\lambda^2\leq 1$. By Lemma \ref{subg}, we have $\|v(w)\|_{\psi_2}^2\leq (4L_v^2d+2|v(\boldsymbol{0})|^2)/\log2\leq 4R^2$. Hence, by Lemma \ref{normeq}, we have $\|v(w)^2\|_{\psi_1}=\|v(w)\|_{\psi_2}^2\leq 4R^2$. By triangle inequality, we have $\|v(w)^2-\mathbb{E}[v(w)^2]\|_{\psi_1}\leq\|v(w)^2\|_{\psi_1}+\|\mathbb{E}[v(w)^2]\|_{\psi_1}$. Given that $\mathbb{E}[v(w)^2]$ is a constant with an upper bound $R^2$, by the definition of the sub-exponential norm, we have $\|\mathbb{E}[v(w)^2]\|_{\psi_1}\leq\mathbb{E}[v(w)^2]/\log 2\leq 2R^2$. To conclude, we have that $\|v(w)^2-\mathbb{E}[v(w)^2]\|_{\psi_1}\leq6R^2$.
	
	We apply Lemma \ref{bern} for random variables $X_m=v(w_m)^2-\mathbb{E}[v(w)^2]$ by setting $t=24R^2\left(\sqrt{\frac{\log(2/\delta)}{M}}+\frac{\log(2/\delta)}{M}\right)$. We obtain
	\[
		P\left(\frac{1}{M}\sum_{m=1}^{M}v(w_m)^2-\mathbb{E}[v(w)^2]>t\right)\leq \exp\left(-\min\left\{\frac{Mt^2}{16\|X\|_{\psi_1}^2},\frac{Mt}{4\|X\|_{\psi_1}}\right\}\right)\leq \frac{\delta}{2}.
	\]
	Because $\mathbb{E}[v(w)^2]\leq R^2$, we obtain that
	\[
		P\left(\frac{1}{M}\sum_{m=1}^{M}v(w_m)^2-R^2>t\right)\leq P\left(\frac{1}{M}\sum_{m=1}^{M}v(w_m)^2-\mathbb{E}[v(w)^2]>t\right)\leq \frac{\delta}{2}.
	\]
	Therefore, with probability of at least $1-\delta/2$, we have
	\[
		\frac{1}{M}\sum_{m=1}^{M}v(w_m)^2\leq R^2+24R^2\left(\sqrt{\frac{\log(2/\delta)}{M}}+\frac{\log(2/\delta)}{M}\right)\leq R^2+24R^2\left(\sqrt{{\log(2/\delta)}}+{\log(2/\delta)}\right).
	\]
	Without loss of generality, we assume $\delta<1/2$, then $1<\sqrt{{\log(2/\delta)}}<\log(2/\delta)$ and hence
	\[
		\frac{1}{M}\sum_{m=1}^{M}v(w_m)^2\leq 49R^2{\log(2/\delta)}.
	\]

	% In the meantime, we have
	% \[
	% 	\mathbb{P}\left(\sum_{m=1}^{M}v_m^2\geq \frac{2MR^2}{\delta}\right)\leq \frac{M\mathbb{E}_{w\sim\mathcal{N}(0,1)}[v(w)^2]\delta}{2MR^2}\leq \frac{\delta}{2}.
	% \]

	Combining the two inequalities and taking the union bound of the probabilities, we have that with probability of at least at least $1-\delta$, it holds that
	\[
        \mathbb{E}_{x}\left|\hat{\varphi}(x) - \varphi(x)\right|\leq \frac{18R\sqrt{\log{(4/\delta)}}}{\sqrt{M}},
    \]
	and
	\[
		\frac{1}{M}\sum_{m=1}^{M}v(w_m)^2\leq 49R^2{\log(2/\delta)}.
	\]
	% \[
	% 	\frac{1}{M}\sum_{m=1}^{M}v_m^2\leq \frac{2R^2}{\delta}.
	% \]

\end{proof}

\section{DEFERRED PROOFS IN SECTION 4}
% eferred proof in section 4}
\subsection{Proof of Proposition \ref{pro43}}
\label{proofpro42}
\begin{proof}
	To start with, we define the Gaussian function with parameter $h$ as 
	\[
		\phi_{h}(x):=\frac{1}{\sqrt{2\pi}h}\exp\left(-\frac{x^2}{2h^2}\right).
	\]

	First, we approximate $\sigma$ by $\sigma*\phi_{h}=\int_{\mathbb{R}}\sigma(x-y)\phi_{h}(y)dy$. Because $\sigma$ is $L$-Lipschitz continuous, we have that $|\sigma(x)-\sigma(x-y)|\leq L|y|$. Together with the fact $|\sigma|\leq\|\sigma\|_{\infty}$, we have that
	\begin{equation*}
		\begin{split}
			&\left|\sigma(x)-(\sigma *\phi_{h})(x)\right|\\
			=&\left| \sigma(x) - \int_{\mathbb{R}}\sigma(x-y)\phi_{h}(y)dy\right|\\
			\leq&\int_{\mathbb{R}}\left| \sigma(x) - \sigma(x-y)\right|\phi_{h}(y)dy\\
			=&\int_{[-\delta,\delta]}\left| \sigma(x) - \sigma(x-y)\right|\phi_{h}(y)dy+\int_{\mathbb{R}-[-\delta,\delta]}\left| \sigma(x) - \sigma(x-y)\right|\phi_{h}(y)dy\\
			\leq& \int_{[-\delta,\delta]}L\left| y\right|\phi_{h}(y)dy+\int_{\mathbb{R}-[-\delta,\delta]}2\|\sigma\|_{\infty}\phi_{h}(y)dy\\
			\leq& L\delta + 2\|\sigma\|_{\infty}\cdot P\left(|Z|\geq \frac{\delta}{h}\right),
		\end{split}
	\end{equation*}
	where $Z\sim \mathcal{N}(0,1)$. The tail probability of Gaussian random variable is estimated as
	\[
		\begin{split}
			P\left(|Z|\geq \frac{\delta}{h}\right)
			=&2P\left(Z\geq \frac{\delta}{h}\right)\\
			\overset{\lambda>0}{=}&2P\left(e^{\lambda Z}\geq e^\frac{\lambda\delta}{h}\right)\\
			\leq& 2\inf_{\lambda> 0} \frac{\mathbb{E}e^{\lambda Z}}{e^{\frac{\lambda\delta}{h}}}\\
			=&2\exp\left(-\frac{\delta^2}{2h^2}\right).
		\end{split}
	\]

	By taking \[\delta=\frac{\epsilon}{4L},\quad h\leq \frac{\epsilon}{4\sqrt{2}L\sqrt{\log \frac{16\|\sigma\|_{\infty}}{\epsilon}}},\] we have
	\begin{equation*}
		\begin{split}
			&\left| \sigma(x) - \int_{\mathbb{R}}\sigma(x-y)\phi_{h}(y)dy\right|\\
			\leq& L\delta + 2\|\sigma\|_{\infty}\cdot P\left(|Z|\geq \frac{\delta}{h}\right)\\
			\leq& \frac{\epsilon}{4}+\frac{\epsilon}{4}=\frac{\epsilon}{2}.
		\end{split}
	\end{equation*}

	In the second step, we approximate $\sigma *\phi_{h}$ by the Riemann sum $\sum_{i=1}^{N}f(y_i)\cdot(y_{i}-y_{i-1})\cdot \phi_{h}(x-y_i)$.

	For the convolution part, we have
	\[
		\begin{split}
			(\sigma *\phi_{h})(x)=&\int_{\mathbb{R}}\sigma(x-y)\phi_{h}(y)dy\\
			=&\int_{\mathbb{R}}\sigma(y)\phi_{h}(x-y)dy\\
			=&\int_{\mathcal{K}}\sigma(y)\phi_{h}(x-y)dy\\
			=&\sum_{i=1}^{N}\int_{y_{i-1}}^{y_{i}}\sigma(y)\phi_{h}(x-y)dy.\\
		\end{split}
	\]

	Then we have
	\begin{equation}
		\label{eqb1}
		\begin{split}
			&\left|(\sigma *\phi_{h})(x)-\sum_{i=1}^{N}\sigma(y_i)\cdot(y_{i}-y_{i-1})\cdot \phi_{h}(x-y_i)\right|\\
			=&\left|\sum_{i=1}^{N}\int_{y_{i-1}}^{y_{i}}\sigma(y)\phi_{h}(x-y)dy-\sum_{i=1}^{N}\int_{y_{i-1}}^{y_{i}}\sigma(y_i)\phi_{h}(x-y_i)dy\right|\\
			\leq&\left|\sum_{i=1}^{N}\int_{y_{i-1}}^{y_{i}}\sigma(y)\phi_{h}(x-y)dy-\sum_{i=1}^{N}\int_{y_{i-1}}^{y_{i}}\sigma(y_i)\phi_{h}(x-y)dy\right|\\
			&+\left|\sum_{i=1}^{N}\int_{y_{i-1}}^{y_{i}}\sigma(y_i)\phi_{h}(x-y)dy-\sum_{i=1}^{N}\int_{y_{i-1}}^{y_{i}}\sigma(y_i)\phi_{h}(x-y_i)dy\right|\\
			\leq&\sum_{i=1}^{N}\int_{y_{i-1}}^{y_{i}}\left|\sigma(y)-\sigma(y_i)\right|\phi_{h}(x-y)dy+\sum_{i=1}^{N}\int_{y_{i-1}}^{y_{i}}|\sigma(y_i)|\cdot\left|\phi_{h}(x-y)-\phi_{h}(x-y_i)\right|dy\\
			\leq& \sum_{i=1}^{N}L(y_{i}-y_{i-1})\int_{y_{i-1}}^{y_{i}}\phi_{h}(x-y)dy+\|\sigma\|_{\infty}\sum_{i=1}^{N}\int_{y_{i-1}}^{y_{i}}\left|\phi_{h}(x-y)-\phi_{h}(x-y_i)\right|dy.\\
			% \leq& \frac{L|\mathcal{K}|}{N}+\|\sigma\|_{\infty}N\cdot\frac{1}{\sqrt{2\pi e}h^2} \left(\frac{|\mathcal{K}|}{N}\right)^2\\
			% =&\left(L+\frac{\|\sigma\|_{\infty}|\mathcal{K}|}{\sqrt{2\pi e}h^2}\right)\cdot\frac{|\mathcal{K}|}{N}\\
			% <&\left(L+\frac{\|\sigma\|_{\infty}|\mathcal{K}|}{\sqrt{2\pi e}}\right)\cdot\frac{|\mathcal{K}|}{Nh^2},
		\end{split}
	\end{equation}

	For the first term, if $|\mathcal{K}|/N \leq \epsilon/4L$, then we have
	\begin{equation}
		\label{eqb2}
		\sum_{i=1}^{N}L(y_{i}-y_{i-1})\int_{y_{i-1}}^{y_{i}}\phi_{h}(x-y)dy\leq \frac{L|\mathcal{K}|}{N}\sum_{i=1}^{N}\int_{y_{i-1}}^{y_{i}}\phi_{h}(x-y)dy\leq \frac{L|\mathcal{K}|}{N}\leq \frac{\epsilon}{4}.
	\end{equation}

	For the second term, we first consider the derivative of $\phi_h(x)$.
	\[
		\begin{split}
			|\phi_h^\prime(x)|=&\left|\frac{1}{\sqrt{2\pi}h^2}\cdot\frac{x}{h}\cdot\exp\left(-\frac{1}{2}\left(\frac{x}{h}\right)^2\right)\right|\\
			\leq &\frac{1}{\sqrt{2\pi}h^2}\exp\left(-\frac{1}{4}\left(\frac{x}{h}\right)^2\right)\\
			% \leq& \left|\frac{1}{\sqrt{2\pi}h^2}\sup_{t\in \mathbb{R}} \left\{t\exp\left(-\frac{t^2}{2}\right)\right\}\right|\\
			% =&\frac{1}{\sqrt{2\pi e}h^2},
		\end{split}
	\]
	where we use the inequality $x\leq \exp(x^2/4)$.
	
	Taking $t=\sqrt{4\log\left(\frac{8\|\sigma\|_{\infty}|\mathcal{K}|^2}{\sqrt{2\pi}}\cdot\frac{1}{\epsilon N h^2}\right)}$, if $|x|>th$, then
	\[
		|\phi_h^\prime(x)|\leq \frac{\epsilon N }{8\|\sigma\|_{\infty}|\mathcal{K}|^2}.
	\]
	If $|x|\leq th$, then
	\[
		\begin{split}
			|\phi_h^\prime(x)|=&\left|\frac{1}{\sqrt{2\pi}h^2}\cdot\frac{x}{h}\cdot\exp\left(-\frac{1}{2}\left(\frac{x}{h}\right)^2\right)\right|\\
			\leq& \left|\frac{1}{\sqrt{2\pi}h^2}\sup_{t\in \mathbb{R}} \left\{t\exp\left(-\frac{t^2}{2}\right)\right\}\right|\\
			=&\frac{1}{\sqrt{2\pi e}h^2}.
		\end{split}
	\]
	Consequently, for the second term, it holds that
	\begin{equation}
		\label{eqb3}
		\begin{split}
			&\|\sigma\|_{\infty}\sum_{i=1}^{N}\int_{y_{i-1}}^{y_{i}}\left|\phi_{h}(x-y)-\phi_{h}(x-y_i)\right|dy\\
			= &\|\sigma\|_{\infty}\sum_{i=1}^{N}\int_{y_{i-1}}^{y_{i}}\left|\int_{y}^{y_i}\phi^{\prime}_{h}(x-z)dz\right|dy\\
			\leq &\|\sigma\|_{\infty}\sum_{i=1}^{N}\int_{y_{i-1}}^{y_{i}}\left|\sup_{z\in[x-y_i,x-y_{i-1}]}|\phi^{\prime}_{h}(z)||y-y_i|\right|dy\\
			\leq &\|\sigma\|_{\infty}\sum_{i=1}^{N}\sup_{z\in[x-y_i,x-y_{i-1}]}|\phi^{\prime}_{h}(z)|\left(\frac{|\mathcal{K}|}{N}\right)^2\\
			\leq &\|\sigma\|_{\infty}\frac{2th}{\frac{|\mathcal{K}|}{N}}\cdot \frac{1}{\sqrt{2\pi e}h^2}\cdot\left(\frac{|\mathcal{K}|}{N}\right)^2+ \|\sigma\|_{\infty}N\cdot \frac{\epsilon N }{8\|\sigma\|_{\infty}|\mathcal{K}|^2}\cdot  \left(\frac{|\mathcal{K}|}{N}\right)^2\\
			=&\|\sigma\|_{\infty}\sqrt{\frac{2}{\pi e}}\cdot\frac{t|\mathcal{K}|}{hN}+\frac{\epsilon}{8}\\
			=&\|\sigma\|_{\infty}\sqrt{\frac{2}{\pi e}}\sqrt{4\log\left(\frac{8\|\sigma\|_{\infty}|\mathcal{K}|}{\sqrt{2\pi}}\cdot\frac{|\mathcal{K}|}{\epsilon h^2 N }\right)}\cdot\frac{|\mathcal{K}|}{hN}+\frac{\epsilon}{8}.\\
		\end{split}
	\end{equation}
	The fifth line holds because there are at most $2thN/|\mathcal{K}|$ intervals in which $|\phi_h^{\prime}|>\frac{\epsilon N }{8\|\sigma\|_{\infty}|\mathcal{K}|^2}$.
		
	Let \[\frac{|\mathcal{K}|}{N}\leq\frac{\epsilon h\sqrt{{\pi e}}}{16\sqrt{2}\|\sigma\|_{\infty}\log\left({\frac{8\|\sigma\|_{\infty}|\mathcal{K}|}{\sqrt{2\pi}\epsilon h^2}}\right)}\land \frac{\epsilon}{4L}\ll 1.\]
	Then
	\begin{equation}
		\label{eqb4}
		\begin{split}
			&\|\sigma\|_{\infty}\sqrt{\frac{2}{\pi e}}\sqrt{4\log\left(\frac{8\|\sigma\|_{\infty}|\mathcal{K}|}{\sqrt{2\pi}}\cdot\frac{|\mathcal{K}|}{\epsilon h^2 N }\right)}\cdot\frac{|\mathcal{K}|}{hN}\\
			=&\|\sigma\|_{\infty}\sqrt{\frac{2}{\pi e}}\sqrt{4\log\left(\frac{8\|\sigma\|_{\infty}|\mathcal{K}|}{\sqrt{2\pi}\epsilon h^2 }\right)+4\log\left(\frac{|\mathcal{K}|}{N }\right)}\cdot\frac{|\mathcal{K}|}{hN}\\
			\leq &\|\sigma\|_{\infty}\sqrt{\frac{2}{\pi e}}\sqrt{4\log\left(\frac{8\|\sigma\|_{\infty}|\mathcal{K}|}{\sqrt{2\pi}\epsilon h^2 }\right)}\cdot\frac{|\mathcal{K}|}{hN}\leq \frac{\epsilon}{8}.\\
		\end{split}
	\end{equation}

	Putting (\ref{eqb2}), (\ref{eqb3}) and (\ref{eqb4}) into (\ref{eqb1}), we conclude that
	\[
		\left|(\sigma *\phi_{h})(x)-\sum_{i=1}^{N}\sigma(y_i)\cdot(y_{i}-y_{i-1})\cdot \phi_{h}(x-y_i)\right|\leq \frac{\epsilon}{2}.
	\]

	Hence,
	\[
		\begin{split}
			&\left|\sigma(x)-\sum_{i=1}^{N}\sigma(y_i)\cdot(y_{i}-y_{i-1})\cdot \phi_{h}(x-y_i)\right|\\
			\leq&\left| \sigma(x) - (\sigma *\phi_{h})(x)\right|+\left|(\sigma *\phi_{h})(x)-\sum_{i=1}^{N}\sigma(y_i)\cdot(y_{i}-y_{i-1})\cdot \phi_{h}(x-y_i)\right|\\
			\leq& \frac{\epsilon}{2}+\frac{\epsilon}{2}=\epsilon.
		\end{split}
	\]

	Let
	\[
		B_i(x)=\exp\left(-\frac{(x-y_i)^2}{2h^2}\right),\quad a_i=\frac{|\mathcal{K}|}{\sqrt{2\pi}hN}\cdot\sigma(y_i),
	\]
	then 
	\[
		\sum_{i=1}^{N}\sigma(y_i)\cdot(y_{i}-y_{i-1})\cdot \phi_{h}(x-y_i)=\sum_{i=1}^N a_i B_i(x).
	\]
	
	Hence
	\[
		\left\|\sigma-\sum_{i=1}^N a_i B_i(x)\right\|_{\infty}\leq \epsilon,
	\]
	and
	\[
		\sum_{i=1}^{N}|a_i|\leq \sum_{i=1}^{N}\frac{|\mathcal{K}|}{\sqrt{2\pi}hN}\cdot|\sigma(y_i)|\leq \frac{\|\sigma\|_{\infty}|\mathcal{K}|}{\sqrt{2\pi}h}.
	\]
	In addition,
	\[
		\sum_{i=1}^{N}|a_i|^2\leq \sum_{i=1}^{N}\frac{|\mathcal{K}|^2}{2\pi h^2N^2}\cdot|\sigma(y_i)|^2\leq \frac{\|\sigma\|_{\infty}^2|\mathcal{K}|}{2\pi}\cdot\frac{|\mathcal{K}|}{Nh^2}.
	\]

	To conclude, if one sets
	\begin{equation}
		\label{paraset}
		h_i\equiv h\leq \frac{\epsilon}{4\sqrt{2}L\sqrt{\log \frac{16\|\sigma\|_{\infty}}{\epsilon}}},\quad \frac{|\mathcal{K}|}{N}\leq\frac{\epsilon h\sqrt{{\pi e}}}{16\sqrt{2}\|\sigma\|_{\infty}\log\left({\frac{8\|\sigma\|_{\infty}|\mathcal{K}|}{\sqrt{2\pi}\epsilon h^2}}\right)}\land \frac{\epsilon}{4L},
	\end{equation}
	and $c_i$ be the grid points of $\mathcal{K}$, then there exists $\{a_i\}_{i=1}^N$ such that
	\[
		\left\|\sigma-\sum_{i=1}^N a_i B_i(x)\right\|_{\infty}\leq \epsilon,
	\]
	and
	\[
		\sum_{i=1}^{N}|a_i|\leq \frac{\|\sigma\|_{\infty}|\mathcal{K}|}{\sqrt{2\pi}h},\quad\quad\sum_{i=1}^{N}|a_i|^2\leq \frac{\|\sigma\|_{\infty}^2|\mathcal{K}|}{2\pi}\cdot\frac{|\mathcal{K}|}{Nh^2}.
	\]
	% \[
	% 	\sum_{i=1}^{N}a_i^2\leq \frac{\|\sigma\|_{\infty}^2|\mathcal{K}|}{4\pi\left(L+\frac{\|\sigma\|_{\infty}|\mathcal{K}|}{\sqrt{2\pi e}}\right)} \epsilon.
	% \]
	We remark that the choice of $c_i$ could be arbitrary as long as $c_i\in[y_{i-1},y_i]$. And the $L_2$ bound actually implies the $L_1$ bound because $\sum_{i=1}^{N}|a_i|\leq\sqrt{\sum_{i=1}^{N}|a_i|^2}\sqrt{N}$.
% \end{proof}

% \subsection{Proof of Theorem \ref{pro43}}
% \label{proofpro43}
% \begin{proof}
    % By Proposition \ref{thm1}, for 
	Now, replacing $\epsilon$ with $\epsilon/R$ in (\ref{paraset}), there exists $N>0$ and $\{h_i,c_i,a_i\}_{i=1}^N$ such that
    \[
        \left\|\sigma(x)-\sum_{i=1}^{N}a_i B_i(x)\right\|_{\infty} < \frac{\epsilon}{R}.
    \]

    Thus
    \begin{equation}
        \begin{aligned}
        \left\|f^*(x)-\tilde{f}(x)\right\|_{\infty}&=\left\|\mathbb{E}_{w\sim \mathcal{N}(0,1)}\left[\left(\sigma(w^\top x)-\sum_{i=1}^N a_i B_i(w^\top x)\right)v(w)\right]\right\|_{\infty}\\
        &\leq \mathbb{E}_{w\sim \mathcal{N}(0,1)}\left[\left\|\sigma(w^\top x)-\sum_{i=1}^N a_i B_i(w^\top x)\right\|_{\infty}|v(w)|\right]\\
        &\leq \left\|\sigma(w^\top x)-\sum_{i=1}^N a_i B_i(w^\top x)\right\|_{\infty}\left(\mathbb{E}_{w\sim \mathcal{N}(0,1)}\left[v(w)^2\right]\right)^{1\over 2}\\
        &\leq {\epsilon\over R} \cdot R \leq \epsilon.
        \end{aligned}
    \end{equation}
    
\end{proof}

\subsection{Proof of Theorem \ref{pro44}}
\label{proofpro44}

In the proof, we attempt to approximate $\tilde{f}$ with finite-width random feature model. We clarify the notations here and denote
\[
    \varphi_i(x)=\mathbb{E}_{w\sim \mathcal{N}(0,I_d)}\left[B_i(w^\top x)v(w)\right],\quad
    \hat{\varphi}_i(x)=\frac{1}{M}\sum_{m=1}^{M}B_i(w_m^\top x)v_m.
\]
Then
\[
    \tilde{f}(x) = \sum_{i=1}^{N} a_i \varphi_i(x),\quad
    \hat{f}(x) = \sum_{i=1}^{N} a_i \hat{\varphi}_i(x).
\]

\begin{proof}
	For all $\epsilon>0$, under the parameter settings of Proposition \ref{pro43}, there exists $\{a_i\}_{i=1}^N$ such that 
	\[
		\left\|\tilde{f}(x)-f^*(x)\right\|_{\infty}\leq \epsilon,\quad \left\|\sigma(x)-\sum_{i=1}^{N}a_i B_i(x)\right\|_{\infty} < \epsilon/R,\quad \sum_{i=1}^{N}a_i^2\leq  \frac{\|\sigma\|_{\infty}^2|\mathcal{K}|^2}{2\pi h^2N}.
	\]
	So we first have
	\begin{equation}
		\label{chaxiang}
		\begin{aligned}
			\mathbb{E}_{x}\left|\hat{f}(x)-f^*(x)\right|
			&= \mathbb{E}_{x}\left|\hat{f}(x)-\tilde{f}(x)+\tilde{f}(x)-f^*(x)\right|\\
			&\leq \mathbb{E}_{x}\left|\hat{f}(x)-\tilde{f}(x)\right|+\mathbb{E}_{x}\left|\tilde{f}(x)-f^*(x)\right|\\
			&\leq \mathbb{E}_{x}\left|\sum_{i=1}^{N} a_i \left(\hat{\varphi}_i(x)-\varphi_i(x)\right)\right|+\epsilon.\\
			% &\leq \mathbb{E}_{x}\sum_{i=1}^{N} |a_i| \left|\hat{\varphi}_i(x)-\varphi_i(x)\right|+\epsilon\\
			% &\leq \sum_{i=1}^{N} |a_i|\mathbb{E}_{x} \left|\hat{\varphi}_i(x)-\varphi_i(x)\right|+\epsilon.\\
		\end{aligned}
	\end{equation}
	Next, we aim to derive a high probability bound on $\mathbb{E}_{x}\left|\sum_{i=1}^{N} a_i \left(\hat{\varphi}_i(x)-\varphi_i(x)\right)\right|$. The proof techiniques are similar to those of Theorem \ref{pro34}. First, we have
	\[
		\begin{split}
			\sum_{i=1}^{N} a_i \left(\hat{\varphi}_i(x)-\varphi_i(x)\right)
			=&\,\sum_{i=1}^{N} a_i \left(\frac{1}{M}\sum_{m=1}^{M}B_i(w_m^\top x)v(w_m)-\varphi_i(x)\right)\\
			=&\,\frac{1}{M}\sum_{m=1}^{M}\left(\sum_{i=1}^{N} a_i B_i(w_m^\top x)v(w_m)-\sum_{i=1}^{N} a_i\varphi_i(x)\right)
		\end{split}
	\]
	It boils down to estimating the sub-gaussian norms of the random variables $Z_m=\sum_{i=1}^{N} a_i B_i(w_m^\top x)v(w_m)-\sum_{i=1}^{N} a_i\varphi_i(x)$ where $\{w_m\}_{m\in[M]}\overset{i.i.d.}{\sim}\mathcal{N}(0,I_d)$.

	Consider
	\[
		\begin{split}
			Z_m^2=&\,\left(\sum_{i=1}^{N} a_i B_i(w_m^\top x)v(w_m)-\sum_{i=1}^{N} a_i\varphi_i(x)\right)^2\\
			\leq &\,2\left(\sum_{i=1}^{N} a_i B_i(w_m^\top x)v(w_m)\right)^2+2\left(\sum_{i=1}^{N} a_i\varphi_i(x)\right)^2\\
			= &\,2v(w_m)^2\left(\sum_{i=1}^{N} a_i B_i(w_m^\top x)\right)^2+2\left(\mathbb{E}_{w}\sum_{i=1}^{N} a_i B_i(w^\top x)v(w)\right)^2\\
			\leq &\,2v(w_m)^2\left(\sum_{i=1}^{N} a_i B_i(w_m^\top x)\right)^2+2\mathbb{E}_{w}\left(\sum_{i=1}^{N} a_i B_i(w^\top x)\right)^2 \mathbb{E}_{w}\left(v(w)^2\right).\\
		\end{split}
	\]
	Because $\left\|\sigma(x)-\sum_{i=1}^{N}a_i B_i(x)\right\|_{\infty} < \epsilon/R$, we have $\left|\sum_{i=1}^{N}a_i B_i(x)\right|\leq \|\sigma\|_{\infty}+\epsilon/R$ for all $x$. Hence,
	\[
		\begin{split}
			Z_m^2\leq&\, 2v(w_m)^2\left(\|\sigma\|_{\infty}+\epsilon/R\right)^2+2R^2\left(\|\sigma\|_{\infty}+\epsilon/R\right)^2\\
			\leq &\, 2(L_v\|w_m\|+|v(\boldsymbol{0})|)^2\left(\|\sigma\|_{\infty}+\epsilon/R\right)^2+2R^2\left(\|\sigma\|_{\infty}+\epsilon/R\right)^2\\
			\leq &\, (4L_v^2\|w_m\|^2+4|v(\boldsymbol{0})|^2+2R^2)\left(\|\sigma\|_{\infty}+\epsilon/R\right)^2.
		\end{split}
	\]
	Similar to the estimation in Eq. (\ref{MGFX1}) and Eq. (\ref{MGFX2}), we have that for $\lambda$ such that $10(\|\sigma\|_{\infty}R+\epsilon)^2\lambda^2\leq 1$, it holds that
	\[
		\mathbb{E}_{W}e^{\lambda^2 Z_m^2}\leq e^{10(\|\sigma\|_{\infty}R+\epsilon)^2\lambda^2}.
	\]
	Similar to the estimatio in Eq. (\ref{MGFX3}), Eq. (\ref{MGFX4}) and Eq. (\ref{MGFX5}), we have that for $\lambda$ such that $160(\|\sigma\|_{\infty}R+\epsilon)^2\lambda^2\leq M$, it holds that
	\[
		\mathbb{E}_{W}e^{\lambda^2 \left(\sum_{m=1}^MZ_m/M\right)^2}\leq e^{160(\|\sigma\|_{\infty}R+\epsilon)^2\lambda^2/M}.
	\]
	Hence, similar to Eq. (\ref{expMGF}), we have
	\[
		\begin{split}
			\mathbb{E}_{W}e^{\lambda^2\left(\mathbb{E}_{x}\left|\sum_{i=1}^{N} a_i \left(\hat{\varphi}_i(x)-\varphi_i(x)\right)\right|\right)^2}\leq&\,\mathbb{E}_{x}\mathbb{E}_{W}e^{\lambda^2\left(\left|\sum_{i=1}^{N} a_i \left(\hat{\varphi}_i(x)-\varphi_i(x)\right)\right|\right)^2}\\
			= &\,\mathbb{E}_{x}\mathbb{E}_{W}e^{\lambda^2 \left(\sum_{m=1}^MZ_m/M\right)^2}\\
			\leq&\, e^{160(\|\sigma\|_{\infty}R+\epsilon)^2\lambda^2/M}.
		\end{split}
	\]
	By Lemma \ref{subg}, we obtain $\left\|\mathbb{E}_{x}\left|\sum_{i=1}^{N} a_i \left(\hat{\varphi}_i(x)-\varphi_i(x)\right)\right|\right\|_{\psi_2}\leq 18(\|\sigma\|_{\infty}R+\epsilon)/\sqrt{M}$. By Lemma \ref{subg0}, we have that 
	\[
		P\left(\mathbb{E}_{x}\left|\sum_{i=1}^{N} a_i \left(\hat{\varphi}_i(x)-\varphi_i(x)\right)\right|\geq \frac{18(\|\sigma\|_{\infty}R+\epsilon)\sqrt{\log(4/\delta)}}{\sqrt{M}}\right)\leq \frac{\delta}{2}.
	\]
	Namely, with probability of at least $1-\delta/2$, we have
	\[
		\mathbb{E}_{x}\left|\sum_{i=1}^{N} a_i \left(\hat{\varphi}_i(x)-\varphi_i(x)\right)\right|\leq \frac{18(\|\sigma\|_{\infty}R+\epsilon)\sqrt{\log(4/\delta)}}{\sqrt{M}}.
	\]
	Therefore, putting it back to (\ref{chaxiang}), with probability of at least $1-\delta/2$, we have
	\[
		\mathbb{E}_{x}\left|\hat{f}(x)-f^*(x)\right|\leq \frac{18(\|\sigma\|_{\infty}R+\epsilon)\sqrt{\log(4/\delta)}}{\sqrt{M}}+\epsilon.
	\]
	Further, by the proof of Theorem \ref{pro34}, the event
	\[
		\frac{1}{M}\sum_{m=1}^{M}v_m^2\leq 49R^2{\log(2/\delta)}
	\]
	happens with probability of at least $1-\delta/2$.

	Taking the union bounds of the probability, we conclude that with probability of at least $1-\delta$, the inequalities hold:
	\begin{equation*}
		\mathbb{E}_{x}\left|\hat{f}(x)-f^*(x)\right|\leq \frac{18(\|\sigma\|_{\infty}R+\epsilon)\sqrt{\log(4/\delta)}}{\sqrt{M}}+\epsilon,
	\end{equation*}
	and\[
		\frac{1}{M}\sum_{m=1}^{M}v_m^2\leq 49R^2{\log(2/\delta)}.
	\]
\end{proof}

\section{DEFERRED PROOFS IN SECTION 5}
% eferred proof in section 5}
% \subsection{Proof of Theorem \ref{pro51}}
\label{proofpro51}

We use Rademacher complexity to obtain the result in Theorem \ref{pro51}. We first recall the definition of Rademacher complexity. Suppose we are given samples $S=\{z_i=(x_i,y_i)\}_{i=1}^{n}$. Let
\[
	\ell \circ f_{\mathcal{V}}:= \{(x,y)\mapsto \ell(f(x),y):f\in f_{\mathcal{V}}\}
\]
be the function class.
Let 
\[
	f_{\mathcal{V}}\circ S:= \{(f(x_1),...,f(x_n)):f\in f_{\mathcal{V}}\},
\]
\[
	\ell \circ f_{\mathcal{V}}\circ S:= \{(\ell(f(x_1),y_1),...,\ell(f(x_n),y_n)):f\in f_{\mathcal{V}}\}
\]
be vector sets. The Rademacher complexity of a function class $\mathcal{H}$ with respect to $S$ is defined as
\[
	\mathcal{R}(\mathcal{H}\circ S):=\frac{1}{n}\mathbb{E}_{\boldsymbol{\xi}}\sup_{h\in \mathcal{H}}\sum_{i=1}^{n}\xi_ih(z_i),
\]
where $\boldsymbol{\xi}=(\xi_1,...,\xi_n)$ and $\{\xi_i\}_{i\in[n]}$ are independent symmetric Bernoulli random variables.

Next, we introduce three lemmas for proving Theorem \ref{pro51}. The first one is a technical tool.
\begin{lemma}[Talagrand's contraction principle (e.g., Exercise 6.7.7 in \citep{vershynin2018high})]
	\label{tala}
	Consider a bounded subset $T\subset \mathbb{R}^n$, and let $\{\xi_i\}_{i\in[n]}$ be independent symmetric Bernoulli random variables. If $\phi_i:\mathbb{R}\rightarrow \mathbb{R}$ are $\rho$-Lipschitz functions, then
	\begin{equation*}
		\mathbb{E}_{\boldsymbol{\xi}}\sup_{t\in T}\sum_{i=1}^{n}\xi_i \phi_i(t_i)\leq \rho\,\mathbb{E}_{\boldsymbol{\xi}}\sup_{t\in T}\sum_{i=1}^{n}\xi_i t_i.
	\end{equation*}
\end{lemma}

Then, through Lemma \ref{tala}, we can obtain the following result describing the Rademacher complexity of the function class of interests.
\begin{lemma}
	\label{proC2}
	All $f\in f_{\mathcal{V}}$ are bounded: \[\|f\|_{\infty}\leq \frac{7\|\sigma\|_{\infty}|\mathcal{K}|R\sqrt{\log(2/\delta)}}{h\sqrt{2\pi}}.\]
	Furthermore, the Rademacher complexity of $\ell \circ f_{\mathcal{V}}$ with respect to samples $S$ is bounded as
	\[
		\mathcal{R}(\ell \circ f_{\mathcal{V}}\circ S)\leq \frac{7\rho\|\sigma\|_{\infty}|\mathcal{K}|R\sqrt{\log(2/\delta)/2\pi}}{h\sqrt{n}}.
	\]
\end{lemma}
For the coherence of the statements, we give the proof of Lemma \ref{proC2} at the end of this section. Finally, we derive the excess risk from the Rademacher complexity using the well known result in supervised learning illustrated below.
\begin{lemma}[e.g., Theorem 26.5 in \citep{shalev2014understanding}]
	\label{proC3}
	Assume that for all $z=(x,y)\sim \mathtt{P}$ and $f\in f_{\mathcal{V}}$ we have that $|\ell(f(x),y)|\leq c$. Then for any $\hat{f}\in f_{\mathcal{V}}$, with probability of at least $1-\delta$ over $\{(x_i,y_i)\}_{i\in[n]}\overset{i.i.d}{\sim}\mathtt{P}$, it holds that
	\[
		L_{D}(f_S)-L_{D}(\hat{f})  \leq 2\mathcal{R}(\ell \circ f_{\mathcal{V}}\circ S)+5c\sqrt{\frac{2\log(8/\delta)}{n}}.
	\]
\end{lemma}

\paragraph{Formal Proof of Theorem \ref{pro51}.}
	Under the conditions andparameter settings of $h,N,\{c_i\}_{i=1}^{N}$ in Theorem \ref{pro44}, with probability of at least $1-\delta$ over $W=(w_1,...,w_M)$, there exists $\hat{f}\in f_\mathcal{V}$ such that
    \[
        \mathbb{E}_{x}\left|\hat{f}(x)-f^*(x)\right|\leq \frac{18(\|\sigma\|_{\infty}R+\epsilon)\sqrt{\log(4/\delta)}}{\sqrt{M}}+\epsilon.
    \]
	% Therefore, by using the relation $\sqrt{a+b}\leq \sqrt{a}+\sqrt{b}$, we also obtain
	% \[
	% 	\begin{split}
	% 		\left\|\hat{f}-f^*\right\|_{L^2}=&\sqrt{\mathbb{E}_{x}\left|\hat{f}(x)-f^*(x)\right|^2}\\
	% 		\leq & 2\sqrt{\frac{C_1N^2\epsilon R}{\delta M}} + 2\epsilon.
	% 	\end{split}
	% \]
	On the other hand, for all $f\in f_{\mathcal{V}}$ and $(x,y)$, we have that 
	\[
		|\ell(f(x),y)|\leq |\ell(0,y)|+\rho |f(x)-0|\leq \rho\left(1+\frac{7\|\sigma\|_{\infty}|\mathcal{K}|R\sqrt{\log(2/\delta)}}{h\sqrt{2\pi}}\right)=:c,
	\]
	where we use the first part of Lemma \ref{proC2}, the Lipschitz property of $\ell$ and the relation $|\ell(0,y)|\leq \rho$.

	Apply the second part of Lemma \ref{proC2} and \ref{proC3} for $\hat{f}$. Then with probability of at least $1-\delta$ over $\{(x_i,y_i)\}_{i\in[n]}\overset{i.i.d}{\sim}\mathtt{P}$, we have that
	\[
		\begin{split}
			L_{D}(f_S)-L_{D}(\hat{f})  \leq&\, 2\mathcal{R}(\ell \circ f_{\mathcal{V}}\circ S)+5c\sqrt{\frac{2\log(8/\delta)}{n}}\\
			\leq &\,2\rho\frac{7\|\sigma\|_{\infty}|\mathcal{K}|R\sqrt{\log(2/\delta)/2\pi}}{h}\sqrt{{1\over n}}+5\rho\left(1+\frac{7\|\sigma\|_{\infty}|\mathcal{K}|R\sqrt{\log(2/\delta)/2\pi}}{h}\right)\sqrt{\frac{2\log(8/\delta)}{n}}\\
			\leq&\,7\rho\left(1+\frac{7\|\sigma\|_{\infty}|\mathcal{K}|R\sqrt{\log(2/\delta)}}{h}\right)\sqrt{\frac{2\log(8/\delta)}{n}}.
		\end{split}
	\]
	Next, we notice that with probability of at least $1-\delta$ over $W=(w_1,...,w_M)$,
	\[
		\begin{split}
			L_D(\hat{f})-L_D(f^*)=&\,\mathbb{E}_{x,y\sim \mathtt{P}}[\ell(\hat{f}(x),y)-\ell(f^*(x),y)]\\
			\leq&\, \rho\,\mathbb{E}_{x}\left|\hat{f}(x)-f^*(x)\right|\\
			\leq&\,\rho \frac{18(\|\sigma\|_{\infty}R+\epsilon)\sqrt{\log(4/\delta)}}{\sqrt{M}}+\rho\epsilon.
		\end{split}
	\]
	Combining the two inequalities and taking the union bounds of the probabilities, we conclude that with probability of at least $1-2\delta$ over $W$ and $S$, it holds that
	\[
		L_{D}(f_S)-L_D(f^*)\leq 7\rho\left(1+\frac{7\|\sigma\|_{\infty}|\mathcal{K}|R\sqrt{\log(2/\delta)}}{h}\right)\sqrt{\frac{2\log(8/\delta)}{n}}+\rho \frac{18(\|\sigma\|_{\infty}R+\epsilon)\sqrt{\log(4/\delta)}}{\sqrt{M}}+\rho\epsilon.
	\]
	Without loss of generality, assume $h\leq 1$ and $\delta\leq 1/2$, then $1\leq \sqrt{\log(2/\delta)}/h$, $\sqrt{\log(2/\delta)}\leq \sqrt{2\log(8/\delta)}$. Consequently,
	\[
		L_{D}(f_S)-L_D(f^*)\leq \rho\frac{14\left(1+7\|\sigma\|_{\infty}|\mathcal{K}|R\right){\log(8/\delta)}}{h\sqrt{n}}+\rho \frac{18(\|\sigma\|_{\infty}R+\epsilon)\sqrt{\log(4/\delta)}}{\sqrt{M}}+\rho\epsilon.
	\]
	Replacing $\delta$ with $\delta/2$, with probability of at least $1-\delta$, we have 
	\[
		L_{D}(f_S)-L_D(f^*)\leq \rho\frac{14\left(1+7\|\sigma\|_{\infty}|\mathcal{K}|R\right){\log(16/\delta)}}{h\sqrt{n}}+\rho \frac{18(\|\sigma\|_{\infty}R+\epsilon)\sqrt{\log(8/\delta)}}{\sqrt{M}}+\rho\epsilon.
	\]
	Let $C=\max\{14\left(1+7\|\sigma\|_{\infty}|\mathcal{K}|R\right), 18(\|\sigma\|_{\infty}R+\epsilon)\}$, we obtain that
	\[
		L_{D}(f_S)-L_D(f^*)\leq \frac{\rho C{\log(16/\delta)}}{h\sqrt{n}}+\frac{\rho C\sqrt{\log(8/\delta)}}{\sqrt{M}}+\rho\epsilon.
	\]

\hfill $\square$

At the end of the proof, we supplement the proof of the second lemma. The proof of Lemma \ref{tala} and \ref{proC3} can be found readily in the literature and are hence omitted.
\paragraph{Proof of Lemma \ref{proC2}.}
	Let $\phi_i(t)=\ell(t,y_i)$ and $t_i=f(x_i)$. Then $\phi_i(t)$ is $\rho$-Lipschitz continuous with respect to $t$. For the boundedness of $T=\{(f(x_1),...,f(x_n)):f\in f_\mathcal{V}\}$, we can see that for all $f\in f_\mathcal{V}$, it holds that
	\begin{equation}
		\label{BD}
		\begin{split}
			|f|=&\left|\frac{1}{M}\sum_{k=1}^{N}a_k\sum_{m=1}^{M}B_k(w_m^\top x)v_m\right|\\
			\leq&\frac{1}{M}\sqrt{\sum_{k=1}^{N}a_k^2}\cdot \sqrt{\sum_{k=1}^{N}\left(\sum_{m=1}^{M}B_k(w_m^\top x)v_m\right)^2}\\
			\leq&\frac{1}{M}\sqrt{\sum_{k=1}^{N}a_k^2}\cdot \sqrt{\sum_{k=1}^{N}\left(\sum_{m=1}^{M}B_k^2(w_m^\top x)\sum_{m=1}^{M}v_m^2\right)}\\
			\leq&\frac{1}{M}\|\boldsymbol{a}\|_2\cdot\sqrt{NM}\|\boldsymbol{v}\|_2\\
			\leq&\sqrt{\frac{N}{M}}\frac{\|\sigma\|_{\infty}|\mathcal{K}|}{h\sqrt{2\pi N}}\cdot\sqrt{49MR^2{\log\left({2\over\delta}\right)}}\\
			=&\frac{7\|\sigma\|_{\infty}|\mathcal{K}|R\sqrt{\log(2/\delta)}}{h\sqrt{2\pi}}.
		\end{split}
	\end{equation}
	Hence, 
	\[
		\|f\|_{\infty}\leq \frac{7\|\sigma\|_{\infty}|\mathcal{K}|R\sqrt{\log(2/\delta)}}{h\sqrt{2\pi}},
	\]
	and for all $t\in T$, $t= (f(x_1),...,f(x_n))$ and $\|t\|\leq \sqrt{n}\|f\|_{\infty}\leq\frac{7\|\sigma\|_{\infty}|\mathcal{K}|R\sqrt{n\log(2/\delta)}}{h\sqrt{2\pi}}$.

	By applying Lemma \ref{tala}, we have
	\[
		\mathbb{E}_{\boldsymbol{\xi}}\sup_{f\in f_\mathcal{V}}\sum_{i=1}^{n}\xi_i \ell(f(x_i),y_i)\leq \rho\,\mathbb{E}_{\boldsymbol{\xi}}\sup_{f\in f_\mathcal{V}}\sum_{i=1}^{n}\xi_i f(x_i).
	\]
	To continue, let $K_1=\frac{\|\sigma\|_{\infty}|\mathcal{K}|}{h\sqrt{2\pi N}}$, $K_2=\sqrt{49MR^2{\log\left({2\over\delta}\right)}}$, $\mathbf{B}_i\in \mathbb{R}^{N\times M}$ with $(\mathbf{B}_i)_{k,m}={B}_k(w_m^\top x_i)$, then we have
	\[
		\begin{split}
			&\mathbb{E}_{\boldsymbol{\xi}}\sup_{f\in f_\mathcal{V}}\sum_{i=1}^{n}\xi_i f(x_i)\\
			=&\mathbb{E}_{\boldsymbol{\xi}}\sup_{\substack{\|\boldsymbol{a}\|_2\leq K_1\\\|\boldsymbol{v}\|_2\leq K_2}}\sum_{i=1}^{n}\xi_i \frac{1}{M}\sum_{k=1}^{N}a_k\sum_{m=1}^{M}B_k(w_m^\top x_i)v_m\\
			=&\frac{1}{M}\mathbb{E}_{\boldsymbol{\xi}}\sup_{\substack{\|\boldsymbol{a}\|_2\leq K_1\\\|\boldsymbol{v}\|_2\leq K_2}}\sum_{k=1}^{N}a_k\sum_{m=1}^{M}\left(\sum_{i=1}^{n}\xi_i B_k(w_m^\top x_i)\right)v_m\\
			=&\frac{1}{M}\mathbb{E}_{\boldsymbol{\xi}}\sup_{\substack{\|\boldsymbol{a}\|_2\leq K_1\\\|\boldsymbol{v}\|_2\leq K_2}}\boldsymbol{a}^{\top}\left(\sum_{i=1}^{n}\xi_i \mathbf{B}_i\right)\boldsymbol{v},\\
		\end{split}
	\]

	Let $\|\cdot\|$ be the operator norm of a matrix, namely the largest singular value of a matrix. Then by the equivalent definition of the operator norm, we have that
	\[
		\begin{split}
			\frac{1}{M}\mathbb{E}_{\boldsymbol{\xi}}\sup_{\substack{\|\boldsymbol{a}\|_2\leq K_1\\\|\boldsymbol{v}\|_2\leq K_2}}\boldsymbol{a}^{\top}\left(\sum_{i=1}^{n}\xi_i \mathbf{B}_i\right)\boldsymbol{v}
			=\frac{K_1K_2}{M}\mathbb{E}_{\boldsymbol{\xi}}\left\|\sum_{i=1}^{n}\xi_i \mathbf{B}_i\right\|.
		\end{split}
	\]
	Furthermore, we have that for any matrix $\mathbf{A}$, it holds that
	\[
		\|\mathbf{A}\|\leq \|\mathbf{A}\|_{\mathrm{Fr}} =\sqrt{\mathrm{Tr}\left(\mathbf{A}\mathbf{A}^\top\right)}.
	\]
	Plugging it into the former expression, we have
	\[
		\begin{split}
			\frac{K_1K_2}{M}\mathbb{E}_{\boldsymbol{\xi}}\left\|\sum_{i=1}^{n}\xi_i \mathbf{B}_i\right\|
			\leq& \frac{K_1K_2}{M}\mathbb{E}_{\boldsymbol{\xi}}\left\|\sum_{i=1}^{n}\xi_i \mathbf{B}_i\right\|_{\mathrm{Fr}}\\
			\leq& \frac{K_1K_2}{M}\sqrt{\mathbb{E}_{\boldsymbol{\xi}}\left\|\sum_{i=1}^{n}\xi_i \mathbf{B}_i\right\|_{\mathrm{Fr}}^2}\\
			=&\frac{K_1K_2}{M}\sqrt{\mathbb{E}_{\boldsymbol{\xi}}\mathrm{Tr}\left(\sum_{i=1}^{n}\xi_i \mathbf{B}_i\right)\left(\sum_{i=1}^{n}\xi_i \mathbf{B}_i\right)^{\top}}\\
			=&\frac{K_1K_2}{M}\sqrt{\mathrm{Tr}\mathbb{E}_{\boldsymbol{\xi}}\left(\sum_{i=1}^{n}\xi_i^2 \mathbf{B}_i\mathbf{B}_i^{\top}+\sum_{\substack{i\neq j\\i,j\in[n]}}\xi_i\xi_j\mathbf{B}_i\mathbf{B}_j^{\top}\right)}\\
			=&\frac{K_1K_2}{M}\sqrt{\mathrm{Tr}\left(\sum_{i=1}^{n} \mathbf{B}_i\mathbf{B}_i^{\top}\right)}\\
			=&\frac{K_1K_2}{M}\sqrt{\sum_{i=1}^{n}\left\| \mathbf{B}_i\right\|_{\mathrm{Fr}}^2}\\
			\leq & \frac{K_1K_2}{M}\sqrt{nNM}=\frac{7\|\sigma\|_{\infty}|\mathcal{K}|R\sqrt{n\log(2/\delta)}}{h\sqrt{2\pi}}.
		\end{split}
	\]
	Finally, we conclude that
	\[
		\begin{split}
			\mathcal{R}(\ell \circ f_{\mathcal{V}}\circ S)=&\frac{1}{n}\mathbb{E}_{\boldsymbol{\xi}}\sup_{t\in T}\sum_{i=1}^{n}\xi_i \ell(f(x_i),y_i)\\
			\leq& \frac{\rho}{n}\,\mathbb{E}_{\boldsymbol{\xi}}\sup_{t\in T}\sum_{i=1}^{n}\xi_i f(x_i)\\
			\leq& \frac{7\rho\|\sigma\|_{\infty}|\mathcal{K}|R\sqrt{\log(2/\delta)/2\pi}}{h\sqrt{ n}}.
		\end{split}
	\]

\hfill $\square$

\newpage

\section{FURTHER DETAILS ON EXPERIMENTS}
% urther details on experiments}
\label{experiments}

\subsection{Datasets}
\label{E1}

\paragraph{Benchmark Datasets.} MNIST and CIFAR-10 are loaded using \texttt{torchvision.datasets} in Python. The UCI datasets are downloaded from the urls listed in Table \ref{urltable}.

\begin{table}[htbp]
	\caption{Urls for downloading the UCI datasets.}
	\label{urltable}
	\vskip 0.15in
	\begin{center}
	\begin{small}
	\begin{sc}
	\begin{tabular}{l|p{10cm}}
	\toprule
	Data set & urls \\
	\midrule
	adult    &  \url{https://archive.ics.uci.edu/ml/machine-learning-databases/adult/adult.data}\\
	protein    & \url{https://archive.ics.uci.edu/ml/machine-learning-databases/00265/CASP.csv} \\
	ct   &  \url{https://archive.ics.uci.edu/ml/machine-learning-databases/00206/slice\_localization\_data.zip}\\
	workloads      &  \url{https://archive.ics.uci.edu/ml/machine-learning-databases/00493/datasets.zip}\\
	millionsongs      & \url{https://archive.ics.uci.edu/ml/machine-learning-databases/00203/YearPredictionMSD.txt.zip} \\
	\bottomrule
	\end{tabular}
	\end{sc}
	\end{small}
	\end{center}
	\vskip -0.1in
\end{table}

\paragraph{Synthetic Datasets.} We choose target functions to be of the form \[f(x)=\mathbb{E}_{w\sim \mathcal{N}(0,I_d)}\left[\sigma(w^\top x)v(w)\right]\] where $x\in\mathbb{R}^d$ and $d=2$. We set $f_1,f_2,f_3$ with the corresponding $\sigma_1, \sigma_2, \sigma_3$ as
\[
   \begin{split}
        \sigma_1(x)=&\sin(\pi x)\bm{1}_{[-1,1]},
        \quad\quad\quad
        \sigma_2(x)=\sin(\pi x)\bm{1}_{[0,1]},\\
        \sigma_3(x)=&-\sin(\pi(x+0.5))\bm{1}_{[-1.5,-0.5]} +  \sin(\pi (x-0.5))\bm{1}_{[0.5,1.5]},
    \end{split}
\]
and $v_i(w)=c_i\max\{b_1^\top w,b_2^\top w\},i\in[3]$, where $b_1, b_2$ are two fixed vectors, and $c_i$ are set as to ensure that $\mathbb{E}_{x}|f_i(x)|\approx 1$. To create the synthetic datasets, we sampled $10^5$ values of $w$ and using the empirical average $\sum_{m=1}^{10^5}\sigma_i(w_m^\top x)v_i(w_m)/10^5$ to approximate $f_i(x)$, so that the approximation error is around $C*10^{-3}$. We sampled $\{x_i\}_{i\in[n]}\overset{i.i.d.}{\sim}\mathcal{N}(0,I_d)$ for sample size $n=15000$ and $d=2$.

\begin{figure}[htbp]
    \centering
    \includegraphics[width=\linewidth]{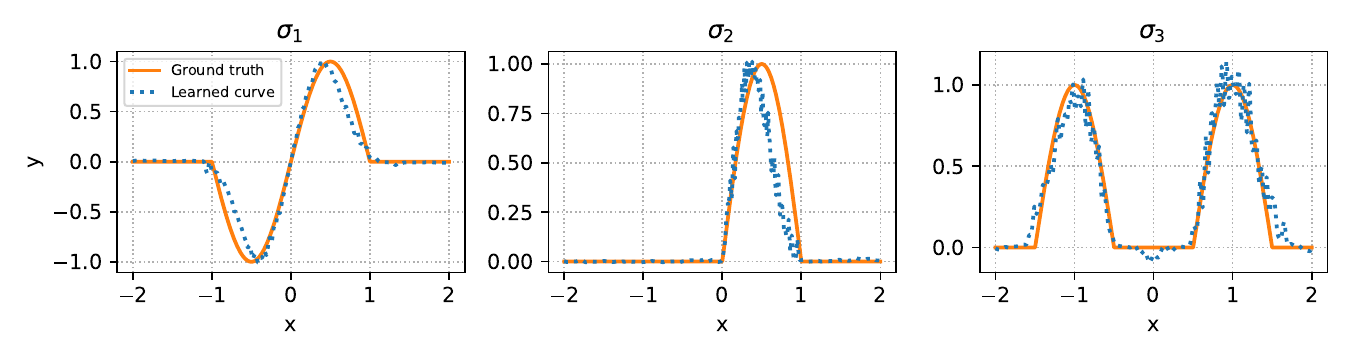}
    \caption{The Activation Functions Learned in RFLAF.}
    \label{synactfall}
\end{figure}

We show all the learned activation functions in RFLAF in Figure \ref{synactfall}. We can see that the learned activation function is close to the ground-truth function with a level of approximation error.

% The codes are available at \url{https://github.com/3b6bf22/repo}.

\subsection{More Experimental Results}
\label{E3}

In Table \ref{acccomp}, we show the test accuracies on the three classification tasks as supplementary results for Table \ref{losscomp}. RFLAFs show consistently advantages over RFMLPs.

\begin{table}[htbp]
  \renewcommand{\arraystretch}{1.3}
  \caption{\centering{Test Accuracies (\%) of Random Feature Models. $N = 16$ for RFLAFs.\\Results are reported as mean $\pm$ std. Best in bold. Second best in italics.}}
	\label{acccomp}
  \begin{center}
	\begin{small}
	\begin{sc}
	\begin{tabular}{l|ccc|cccc}
	\noalign{\vskip 0.5pt}      
  \noalign{\hrule height 0.75pt}
  \noalign{\vskip 0.5pt}
  \multirow{2}{*}{Data set}& \multicolumn{3}{c|}{rflaf} & \multicolumn{4}{c}{rfmlp} \\ \cline{2-8}
    & rbf & bs & pl & relu & cos & tanh & sigmoid \\
  \hline
MNIST       	& $\it{96.10_{\pm 0.30}}$ & $94.95_{\pm 0.63}$ & $\bf{96.26_{\pm 0.13}}$ & $95.40_{\pm 0.09}$ & $65.98_{\pm 4.43}$ & $92.30_{\pm 0.06}$ & $86.86_{\pm 0.56}$ \\
CIFAR-10    	& $48.95_{\pm 0.36}$ & $44.29_{\pm 0.58}$ & $\bf{49.20_{\pm 0.42}}$ & $\it{48.99_{\pm 0.23}}$ & $18.44_{\pm 1.52}$ & $39.62_{\pm 0.23}$ & $34.12_{\pm 1.60}$ \\
adult       	& $\bf{85.66_{\pm 0.15}}$ & $85.05_{\pm 0.18}$ & - & $\it{85.63_{\pm 0.11}}$ & $83.85_{\pm 0.74}$ & $85.05_{\pm 0.15}$ & $84.90_{\pm 0.31}$ \\
  \noalign{\vskip 0.5pt}      
  \noalign{\hrule height 0.75pt}
  \noalign{\vskip 0.5pt}
  \end{tabular}
  \end{sc}
  \end{small}
  \end{center}
\end{table}

In Table \ref{acccomp4}, we show the test accuracies on the three classification tasks as supplementary results for Table \ref{losscomp4}. LANs show better performances in the first two tasks and similar performance in the third task compared to MLP and KAN.

\begin{table}[htbp]
  \renewcommand{\arraystretch}{1.3}
  \caption{Test Accuracies (\%) of Regular Two-layer Networks. $N = 16$ for LAN and KAN.}
	\label{acccomp4}
	\begin{center}
	\begin{small}
	\begin{sc}
	\begin{tabular}{l|ccc|c|c}
  \noalign{\vskip 0.5pt}      
  \noalign{\hrule height 0.75pt}
  \noalign{\vskip 0.5pt}
  \multirow{3}{*}{Data set} & \multicolumn{5}{c}{loss} \\
  \cline{2-6}
    & \multicolumn{3}{c|}{lan} &  {mlp} & \multirow{2}{*}{kan}  \\ \cline{2-5}
    & rbf & bs & pl & relu &  \\
	\hline
MNIST       	& $\it{96.40_{\pm 1.15}}$ & $\bf{96.85_{\pm 0.15}}$ & $95.10_{\pm 0.16}$ & $95.99_{\pm 0.07}$ & $95.19_{\pm 0.14}$ \\
CIFAR-10    	& $\it{52.74_{\pm 0.36}}$ & $\bf{53.18_{\pm 0.61}}$ & $48.13_{\pm 3.91}$ & $49.61_{\pm 0.17}$ & $47.90_{\pm 0.18}$ \\
adult       	& $85.77_{\pm 0.08}$ & $85.76_{\pm 0.30}$ & $83.15_{\pm 0.50}$ & $\bf{85.85_{\pm 0.08}}$ & $\it{85.80_{\pm 0.07}}$ \\
  \noalign{\vskip 0.5pt}      
  \noalign{\hrule height 0.75pt}
  \noalign{\vskip 0.5pt}
  \end{tabular}
  \end{sc}
  \end{small}
  \end{center}
\end{table}

In the following, we additionally summarize the results on RFLAFs of $N=8,16,32,64,128$ and low-degree polynomials, including the test losses (Table \ref{fulllosscomp}), training time (Table \ref{fulltraincomp}) and testing time (Table \ref{fulltestcomp}).

\begin{table}[htbp]
  \renewcommand{\arraystretch}{1.35}
  \caption{Test Losses for All $N$. (Best in Bold.)}
	\label{fulllosscomp}
	\begin{center}
	\begin{small}
	\begin{sc}
	\begin{tabular}{l|ccccccc}
	\noalign{\vskip 0.5pt}      
  \noalign{\hrule height 0.75pt}
  \noalign{\vskip 0.5pt}
  Model-N& MNIST	& CIFAR-10 & adult & protein & ct & workloads & millionsongs \\
  \hline
relu	& $0.159$	& $1.466$	& $0.311$	& $0.241$	& $0.356$	& $2.771$	& $0.951$	\\
cos	& $1.390$	& $2.641$	& $0.363$	& $0.371$	& $0.589$	& $2.634$	& $0.280$	\\
tanh	& $0.277$	& $1.769$	& $0.324$	& $0.650$	& $1.241$	& $24.997$	& $8.434$	\\
sigmoid	& $0.498$	& $1.930$	& $0.327$	& $0.280$	& $0.692$	& $1.707$	& $0.118$	\\
\hline
RBF-8	& $0.204$	& $\bf{1.440}$	& $\bf{0.307}$	& $0.230$	& $0.202$	& $0.736$	& $0.103$	\\
BS-8	& $0.146$	& $1.490$	& $0.312$	& $0.198$	& $0.259$	& $0.476$	& $0.129$	\\
PL-8	& $0.129$	& $1.470$	& $0.322$	& \sc{inf}	& \sc{inf}	& $0.827$	& \sc{inf}	\\
\hline
RBF-16	& $0.126$	& $1.450$	& $0.309$	& $0.204$	& $0.212$	& $0.465$	& $0.102$	\\
BS-16	& $0.165$	& $1.609$	& $0.324$	& $0.194$	& $0.302$	& $0.546$	& $0.120$	\\
PL-16	& $\bf{0.124}$	& $1.482$	& \sc{inf}	& \sc{inf}	& \sc{inf}	& \sc{inf}	& \sc{inf}	\\
\hline
RBF-32	& $0.137$	& $1.462$	& $0.314$	& $0.184$	& $0.201$	& $0.336$	& $0.103$	\\
BS-32	& $0.152$	& $1.493$	& $0.331$	& $0.196$	& $0.283$	& $0.548$	& $0.135$	\\
PL-32	& $0.132$	& $3.493$	& \sc{inf}	& \sc{inf}	& \sc{inf}	& \sc{inf}	& \sc{inf}	\\
\hline
RBF-64	& $0.142$	& $1.480$	& $0.316$	& $\bf{0.180}$	& $0.200$	& $0.302$	& $0.102$	\\
BS-64	& $0.153$	& $1.498$	& $0.349$	& $0.225$	& $0.356$	& $0.616$	& $0.215$	\\
PL-64	& $10.599$	& \sc{inf}	& \sc{inf}	& \sc{inf}	& \sc{inf}	& \sc{inf}	& \sc{inf}	\\
\hline
RBF-128	& $0.145$	& $1.471$	& $0.315$	& $0.183$	& $\bf{0.185}$	& $\bf{0.295}$	& $\bf{0.100}$	\\
BS-128	& $0.161$	& $1.537$	& $0.392$	& $0.286$	& $0.419$	& $0.791$	& $0.377$	\\
PL-128	& \sc{inf}	& \sc{inf}	& \sc{inf}	& \sc{inf}	& \sc{inf}	& \sc{inf}	& \sc{inf}	\\
\hline
PL-2	& $0.189$	& $1.484$	& $0.313$	& $0.266$	& $4.007$	& $1.292$	& $0.119$	\\
PL-4	& $0.137$	& $1.462$	& $0.313$	& $0.330$	& \sc{inf}	& $0.857$	& $0.405$	\\
PL-6	& $0.144$	& $1.470$	& $0.312$	& $26.614$	& \sc{inf}	& $0.810$	& \sc{inf}	\\
	\noalign{\vskip 0.5pt}      
  \noalign{\hrule height 0.75pt}
  \noalign{\vskip 0.5pt}
  \end{tabular}
  \end{sc}
  \end{small}
  \end{center}
\end{table}

\begin{table}[htbp]
  \renewcommand{\arraystretch}{1.35}
  \caption{Train Time for All $N$.}
	\label{fulltraincomp}
	\begin{center}
	\begin{small}
	\begin{sc}
	\begin{tabular}{l|ccccccc}
	\noalign{\vskip 0.5pt}      
  \noalign{\hrule height 0.75pt}
  \noalign{\vskip 0.5pt}
  Model-N& MNIST	& CIFAR-10 & adult & protein & ct & workloads & millionsongs \\
  \hline
relu	& $1.000$	& $1.000$	& $1.000$	& $1.000$	& $1.000$	& $1.000$	& $1.000$	\\
cos	& $0.999$	& $1.000$	& $0.991$	& $1.046$	& $1.049$	& $1.017$	& $1.135$	\\
tanh	& $0.994$	& $1.001$	& $0.986$	& $1.040$	& $1.018$	& $1.028$	& $1.147$	\\
sigmoid	& $0.993$	& $0.996$	& $0.943$	& $0.974$	& $0.993$	& $0.995$	& $1.166$	\\
\hline
RBF-8	& $1.065$	& $1.041$	& $1.525$	& $1.499$	& $1.399$	& $1.065$	& $1.691$	\\
BS-8	& $1.267$	& $1.226$	& $2.886$	& $3.119$	& $2.647$	& $2.280$	& $3.426$	\\
PL-8	& $1.059$	& $1.043$	& $1.388$	& $1.415$	& $1.259$	& $1.007$	& $1.589$	\\
\hline
RBF-16	& $1.099$	& $1.063$	& $1.683$	& $1.756$	& $1.618$	& $1.269$	& $1.974$	\\
BS-16	& $1.452$	& $1.367$	& $4.236$	& $4.732$	& $3.864$	& $3.372$	& $5.139$	\\
PL-16	& $1.078$	& $1.057$	& $1.431$	& $1.410$	& $1.319$	& $1.005$	& $1.669$	\\
\hline
RBF-32	& $1.153$	& $1.126$	& $2.380$	& $2.496$	& $2.197$	& $1.782$	& $2.746$	\\
BS-32	& $1.812$	& $1.699$	& $7.134$	& $8.122$	& $6.471$	& $5.773$	& $8.650$	\\
PL-32	& $1.090$	& $1.082$	& $1.819$	& $1.796$	& $1.578$	& $1.263$	& $1.993$	\\
\hline
RBF-64	& $1.261$	& $1.341$	& $4.303$	& $4.709$	& $3.948$	& $3.366$	& $5.122$	\\
BS-64	& $2.352$	& $2.310$	& $12.927$	& $14.683$	& $11.867$	& $10.549$	& $15.883$	\\
PL-64	& $1.046$	& $1.173$	& $2.797$	& $2.989$	& $2.578$	& $2.148$	& $3.312$	\\
\hline
RBF-128	& $1.520$	& $1.738$	& $7.878$	& $8.716$	& $7.152$	& $6.325$	& $9.563$	\\
BS-128	& $3.742$	& $3.686$	& $25.557$	& $28.601$	& $22.978$	& $20.695$	& $31.085$	\\
PL-128	& $1.151$	& $1.412$	& $5.008$	& $5.467$	& $4.538$	& $3.955$	& $5.997$	\\
\hline
PL-2	& $1.085$	& $1.040$	& $1.336$	& $1.395$	& $1.236$	& $0.963$	& $1.523$	\\
PL-4	& $1.077$	& $1.049$	& $1.355$	& $1.355$	& $1.276$	& $0.977$	& $1.530$	\\
PL-6	& $1.064$	& $1.044$	& $1.366$	& $1.373$	& $1.278$	& $0.995$	& $1.527$	\\
	\noalign{\vskip 0.5pt}      
  \noalign{\hrule height 0.75pt}
  \noalign{\vskip 0.5pt}
  \end{tabular}
  \end{sc}
  \end{small}
  \end{center}
\end{table}

\begin{table}[htbp]
  \renewcommand{\arraystretch}{1.35}
  \caption{Test Time for All $N$.}
	\label{fulltestcomp}
	\begin{center}
	\begin{small}
	\begin{sc}
	\begin{tabular}{l|ccccccc}
	\noalign{\vskip 0.5pt}      
  \noalign{\hrule height 0.75pt}
  \noalign{\vskip 0.5pt}
  Model-N& MNIST	& CIFAR-10 & adult & protein & ct & workloads & millionsongs \\
  \hline
relu	& $1.000$	& $1.000$	& $1.000$	& $1.000$	& $1.000$	& $1.000$	& $1.000$	\\
cos	& $1.002$	& $1.000$	& $1.029$	& $1.022$	& $1.053$	& $0.996$	& $1.179$	\\
tanh	& $0.998$	& $1.001$	& $0.996$	& $1.028$	& $1.006$	& $1.026$	& $1.218$	\\
sigmoid	& $0.992$	& $0.997$	& $0.931$	& $0.903$	& $0.965$	& $0.988$	& $1.241$	\\
\hline
RBF-8	& $1.022$	& $0.997$	& $1.474$	& $1.337$	& $1.214$	& $0.924$	& $1.630$	\\
BS-8	& $1.242$	& $1.193$	& $3.730$	& $3.707$	& $2.730$	& $2.683$	& $4.330$	\\
PL-8	& $1.018$	& $1.010$	& $1.312$	& $1.191$	& $1.020$	& $0.814$	& $1.396$	\\
\hline
RBF-16	& $1.060$	& $1.042$	& $2.103$	& $1.971$	& $1.613$	& $1.381$	& $2.387$	\\
BS-16	& $1.438$	& $1.360$	& $6.197$	& $6.292$	& $4.400$	& $4.425$	& $7.324$	\\
PL-16	& $1.043$	& $1.037$	& $1.677$	& $1.486$	& $1.277$	& $1.041$	& $1.874$	\\
\hline
RBF-32	& $1.137$	& $1.154$	& $3.664$	& $3.523$	& $2.683$	& $2.503$	& $4.182$	\\
BS-32	& $1.843$	& $1.750$	& $11.612$	& $11.918$	& $8.112$	& $8.439$	& $13.763$	\\
PL-32	& $1.080$	& $1.108$	& $2.722$	& $2.508$	& $1.888$	& $1.732$	& $2.924$	\\
\hline
RBF-64	& $1.290$	& $1.375$	& $6.785$	& $6.736$	& $4.845$	& $4.802$	& $7.969$	\\
BS-64	& $2.483$	& $2.390$	& $21.305$	& $21.835$	& $15.033$	& $15.594$	& $25.539$	\\
PL-64	& $1.091$	& $1.206$	& $4.415$	& $4.285$	& $3.173$	& $3.075$	& $5.124$	\\
\hline
RBF-128	& $1.584$	& $1.776$	& $12.730$	& $12.679$	& $8.916$	& $9.170$	& $15.144$	\\
BS-128	& $3.983$	& $3.818$	& $42.399$	& $42.576$	& $29.200$	& $30.676$	& $50.142$	\\
PL-128	& $1.200$	& $1.444$	& $8.028$	& $7.884$	& $5.604$	& $5.691$	& $9.398$	\\
\hline
PL-2	& $1.041$	& $0.994$	& $1.068$	& $0.991$	& $0.920$	& $0.673$	& $1.216$	\\
PL-4	& $1.035$	& $1.007$	& $1.161$	& $1.035$	& $0.994$	& $0.713$	& $1.212$	\\
PL-6	& $1.032$	& $1.001$	& $1.220$	& $1.122$	& $0.981$	& $0.750$	& $1.273$	\\
	\noalign{\vskip 0.5pt}      
  \noalign{\hrule height 0.75pt}
  \noalign{\vskip 0.5pt}
  \end{tabular}
  \end{sc}
  \end{small}
  \end{center}
\end{table}

\newpage

\subsection{Discussions over the Regular Two-layer Networks}
\label{E4}

There are a few points that we would like to supplement.

(1) Implementation details of KAN.

We do not adopt the \texttt{pykan} package developed by the author of KAN due to its inefficiency. KAN was initially proposed to address problems typical of science-related tasks, which are generally smaller in scale than typical machine learning tasks. The author states that they will continue developing the repository primarily for scientific discovery and computing, but without significant updates for efficiency\footnote{See \url{https://github.com/KindXiaoming/pykan?tab=readme-ov-file}}. Furthermore, we tested the model on the baseline MNIST dataset and found that the original KAN struggles to scale as the width $M$ increases due to its inefficiency.

To address this issue, we adopt the \texttt{efficient-KAN} package\footnote{See \url{https://github.com/Blealtan/efficient-kan}}, which enables the model to be sufficiently fast and scalable for comparisons. We use the default KAN configuration in this package, employing B-splines of degree three. To ensure consistency, all models are trained using the Adam optimizer with identical parameter settings.

(2) Why do LAN perform better than MLP and KAN?

The fact that LAN performs better than MLP is natural because the structure of learnable activation functions enables the two-layer networks to represent a broader class of functions. The improvement over MLP is not free though, but in the cost of additional computation for combining the basis functions. However, as shown in Table \ref{losscomp4}, LAN of $N=16$ runs within two times of the running time of MLP, which is acceptable in practice. And we highlight that this is probably not the speed limit of LAN as more techniques over codes may be applied to speed up the training and computation process of LAN (e.g., parallel computing of the $N$ basis functions), just as how KAN has evolves. We do not contribute to the code improvements of LAN, since our study does not focus on LAN but on RFLAF, but we believe that this is probably another interesting future direction.

For the other method, KAN fails to consistently outperform even MLP in the results. This is probably due to the convergence problem of KAN. To control the variables, we apply Adam for all models as the optimizer and train all models with the same epoch number. The author of KAN proposed to use LBFGS to boost convergence, but we do not know how sensitive the model is to the use of the optimizer. But for LAN, the common optimizer Adam works well. The difficulty of optimization in KAN probably comes from the high degree of freedom induced by the extensive number of learnable activation functions. While MLP and LAN (with \texttt{RBF} or \texttt{BS}) successfully achieve very low test errors, KAN seems to converge very slowly and be underfitting in some tasks. In contrast, the two-layer LAN only contains one learnable activation function of $N$ extra learnable parameters compared to MLP. Hence, training LAN is almost as easy as training MLP (at least for the case of two layer), which indicates that LAN is probably more capable of scaling up in typical machine learning problems than KAN.

\end{document}